%% file: acl_latex.tex
\documentclass[11pt]{article}

\usepackage[final]{acl}

\usepackage{times}
\usepackage{latexsym}

\usepackage[T1]{fontenc}

\usepackage[utf8]{inputenc}

\usepackage[varqu]{inconsolata}
\usepackage{graphicx}

\usepackage[utf8]{inputenc} 
\usepackage[T1]{fontenc}    
\usepackage{hyperref}       
\usepackage{url}            
\usepackage{booktabs}       
\usepackage{amsfonts}       
\usepackage{nicefrac}       
\usepackage[nopatch=footnote]{microtype}      
\usepackage{xcolor}         
\usepackage{graphicx}
\usepackage{enumitem}
\usepackage{multirow}
\usepackage{colortbl}
\usepackage{amsmath}
\usepackage{wrapfig}
\usepackage{subfigure}
\usepackage{makecell}
\usepackage{graphicx}
\usepackage{subcaption}
\usepackage{tikzducks}
\usepackage{fvextra}
\usepackage{adjustbox}
\usepackage{listings}
\newcommand{\lmeb}{LMEB}
\usepackage{pdflscape}
\usepackage{tikz}
\usepackage[most]{tcolorbox}
\usepackage{fontawesome5}
\usepackage{simpleicons}
\usepackage{textcomp}
\usepackage{xspace}

\usepackage{setspace}

\usepackage{tabularx}

\definecolor{trophygold}{HTML}{F4B400}
\definecolor{hfYellow}{HTML}{FFD21E}
\definecolor{collectionpurple}{HTML}{7C3AED}
\definecolor{collectionblue}{HTML}{3B82F6}
\definecolor{leaderboardgreen}{HTML}{2E7D32}
\tcbset{
  promptstyle/.style={
    colback=gray!15,   
    colframe=gray!60, 
    boxrule=0.4pt,   
    arc=2pt,         
    left=4pt,right=4pt,top=4pt,bottom=4pt,  
    fonttitle=\bfseries,
  }
}
\newtcolorbox{promptblock}[1][]{promptstyle,#1}

\usepackage{fvextra}
\RecustomVerbatimEnvironment{verbatim}{Verbatim}{breaklines=true, breakanywhere=true}

\newcommand{\smallsection}[1]{{\vspace{0.07in} \noindent \bf {#1.\hspace{5pt}}}}

\newcommand{\smallsectionnodot}[1]{{\vspace{0.07in} \noindent \bf {#1\hspace{5pt}}}}

\title{\href{https://huggingface.co/papers/2603.12572}{LMEB}: Long-horizon Memory Embedding Benchmark}

\author{
    \textbf{Xinping Zhao$^{1,2}$, Xinshuo Hu, Jiaxin Xu$^{1}$, Danyu Tang$^{1}$, Xin Zhang$^{1}$, Mengjia Zhou$^{1}$,} \\
    \textbf{Yan Zhong$^{3}$, Yao Zhou, Zifei Shan,  Meishan Zhang$^{1}$, Baotian Hu$^{1,2}$\thanks{Corresponding Author}~~, Min Zhang$^{1,2}$} \\
    $^{1}$Harbin Institute of Technology (Shenzhen);
    $^{2}$Shenzhen Loop Area Institute (SLAI);
    $^{3}$Peking University \\
    \texttt{zhaoxinping@stu.hit.edu.cn, mason.zms@gmail.com} \\
    \texttt{hubaotian@hit.edu.cn, zhangmin2021@hit.edu.cn}  \\
}

\begin{document}
\maketitle
\begin{abstract}
Embedding models are increasingly used as retrievers in memory-augmented systems, but are typically evaluated on traditional passage retrieval benchmarks.
It remains unclear whether models that excel at passage retrieval can retrieve fragmented, context-dependent, and temporally distant memories.
To address this gap, we introduce the \textbf{L}ong-horizon \textbf{M}emory \textbf{E}mbedding \textbf{B}enchmark (\textbf{LMEB}), comprising 22 English datasets and 193 zero-shot retrieval tasks across four memory types: episodic, dialogue, semantic, and procedural\footnote{\url{https://huggingface.co/datasets/KaLM-Embedding/LMEB}}.
We benchmark 30 embedding models ranging from 33M to 12B parameters.
We find that no single model dominates across memory types and that evaluation scores on LMEB correlate only weakly with those on the MTEB (English, v2) retrieval subset.
These findings underscore the need for benchmarks tailored to long-horizon memory retrieval.
LMEB comes with a versatile, open-source evaluation framework\footnote{\url{https://github.com/embeddings-benchmark/mteb}} and a public leaderboard\footnote{\url{https://mteb-leaderboard.hf.space/benchmark/LMEB}} to guide model selection and advance research on memory embeddings.
\end{abstract}


\input{chapters/intro}
\input{chapters/lmeb}

\input{chapters/setup}
\input{chapters/results}
\input{chapters/related}
\input{chapters/conclusion}
\section*{Limitations}
LMEB broadens embedding evaluation to long-horizon memory retrieval across episodic, dialogue, semantic, and procedural memory.
However, it currently focuses on text-only retrieval and does not cover multimodal memories spanning text, images, audio, video, or other modalities.
In multimodal settings, identifying relevant memories may require combining evidence across modalities.
Extending LMEB to multimodal long-horizon memory retrieval is an important direction for future work.
This extension would need to support both unimodal and multimodal queries and memory items, with annotations indicating which memories are relevant to each query.
It should also extend LMEB's contextualization and delimitation mechanisms to multimodal scenarios.
This would better reflect the retrieval needs of multimodal memory-augmented systems.
In addition, LMEB currently focuses on English datasets.
Extending its language coverage would enable evaluation of multilingual and cross-lingual memory retrieval.

\bibliography{reference}

\appendix
\input{chapters/appendix}
\end{document}

%% file: chapters/intro.tex
\section{Introduction}
\label{sec:intro}
\begin{figure}[t]
    \centering
    \includegraphics[width=1.0\linewidth]{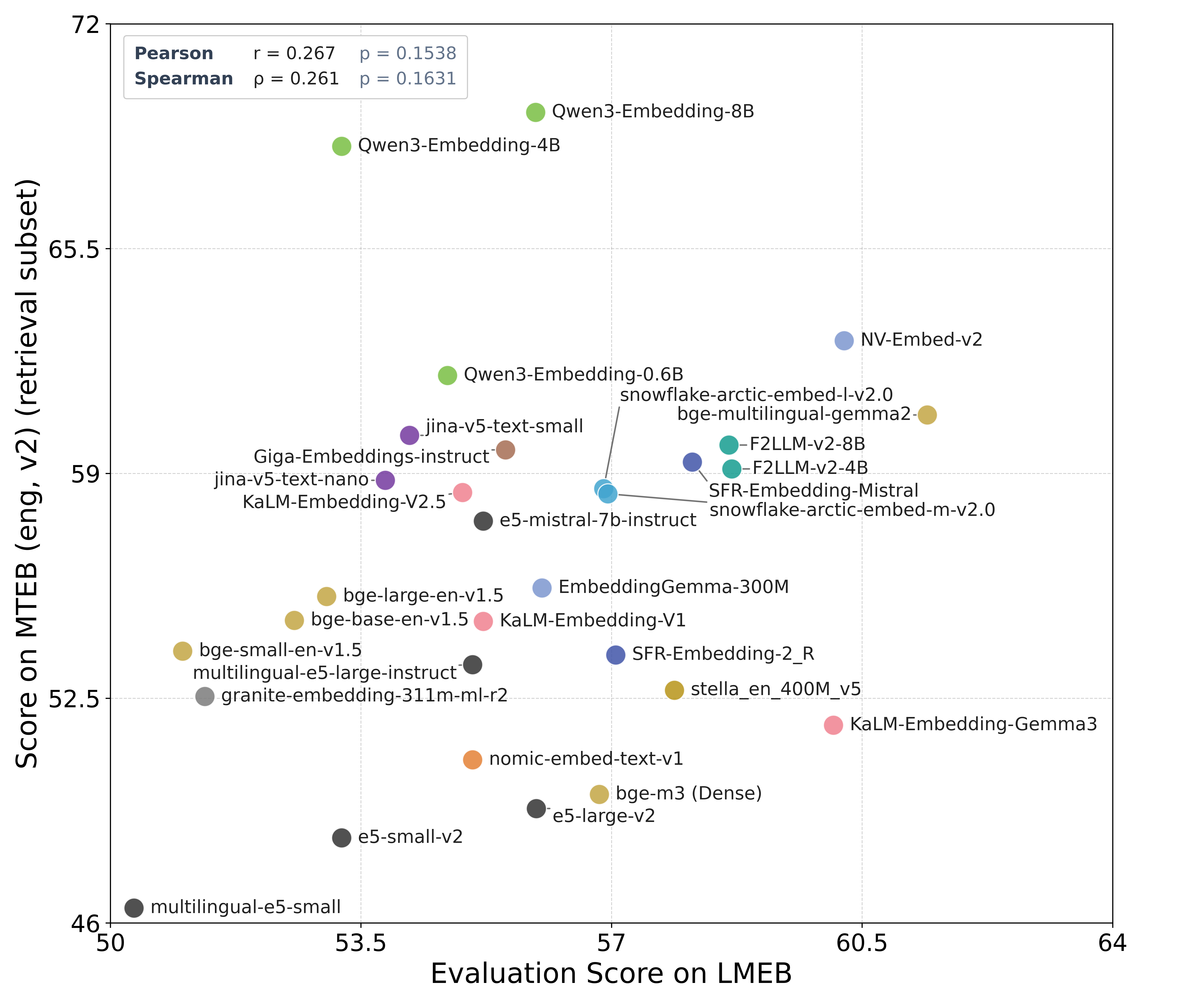}
    \caption{Correlation between the evaluation scores on LMEB and MTEB (eng, v2) (retrieval subset)~\cite{DBLP:conf/iclr/EnevoldsenCKKMS25}. Both  use nDCG@10 as the evaluation metric. Marker colors indicate model families.} 
    \label{fig:corr_main}
    \vspace{-0.1cm}
\end{figure}

Embedding models are increasingly used as retrievers in memory-augmented systems, including agents that retain past interactions~\citep{openclaw2026,DBLP:journals/corr/abs-2508-06433,DBLP:conf/emnlp/SongXZZWWLPL24} and reuse task experiences~\citep{cao2025remember,DBLP:journals/corr/abs-2509-25140,DBLP:journals/corr/abs-2507-23361,DBLP:journals/corr/abs-2504-07079}.
These systems store, retrieve, update, and reason over memories, with retrieval central to their effectiveness, enabling access to relevant long-horizon memories beyond their current context~\citep{du2025rethinkingmemoryllmbased,roediger2022double}.
Long-horizon memory retrieval may require identifying when an event occurred, which interaction a fact belongs to, or whether a past experience is useful for a new task.
This is challenging because relevant memories may be fragmented, context-dependent, and temporally distant~\citep{DBLP:conf/iclr/WuWYZCY25,DBLP:conf/iclr/HuetB025,DBLP:journals/corr/abs-2511-21730}.
Benchmarks such as BEIR, MTEB, and C-MTEB have advanced the standardized evaluation of embedding models~\citep{DBLP:journals/corr/abs-2104-08663,DBLP:conf/eacl/MuennighoffTMR23,DBLP:conf/sigir/XiaoLZMLN24,DBLP:conf/iclr/EnevoldsenCKKMS25}.
However, it remains unclear whether models that perform well on passage retrieval can also retrieve past events, details from earlier conversations, experiences relevant to a new task, and other memories.
Memories differ in their granularity, scope, and organization, making it difficult to compare retrieval performance across settings.
A comprehensive evaluation protocol is therefore needed to assess embedding models' capabilities across long-horizon memory settings.
To address this gap, we introduce the \textbf{L}ong-horizon \textbf{M}emory \textbf{E}mbedding \textbf{B}enchmark (\textbf{LMEB}), comprising \textbf{22 English datasets} and \textbf{193 zero-shot retrieval tasks}.
LMEB spans four memory types~\citep{du2025rethinkingmemoryllmbased}, which differ in what is stored, how memories are organized, and how strongly retrieval depends on temporal context:
\begin{itemize}[leftmargin=1em]
    \item \textbf{Episodic Memory} involves retrieving past events associated with entities, times, and locations~\citep{DBLP:journals/corr/abs-2407-09450,DBLP:journals/corr/abs-2502-06975}. It tests the recovery of specific events through temporal and contextual cues~\citep{DBLP:conf/acl/Miao0ZQ24}.
    \item \textbf{Dialogue Memory} focuses on retrieving relevant turns, rounds, or sessions from conversation histories~\citep{DBLP:conf/iclr/WuWYZCY25,DBLP:conf/acl/MaharanaLTBBF24}. It evaluates the ability to recall previously shared facts and user preferences across multi-turn interactions~\citep{DBLP:journals/corr/abs-2503-07018,DBLP:journals/corr/abs-2402-16288}.
    \item \textbf{Semantic Memory} involves retrieving general knowledge from sentences and paragraphs, independently of particular events~\citep{tulving1972episodic}. It supports reasoning over factual information across domains~\citep{DBLP:journals/corr/abs-2506-15841}.
    \item \textbf{Procedural Memory} focuses on retrieving tools, skills, and task experiences~\citep{DBLP:journals/corr/abs-2508-06433,DBLP:journals/corr/abs-2509-25140}. It tests the recovery of relevant know-how for action selection and problem-solving~\citep{DBLP:journals/corr/abs-2507-21428,DBLP:journals/corr/abs-2507-02259}.
\end{itemize}
LMEB is designed to characterize embedding model performance across diverse long-horizon memory retrieval settings and to support the development of more capable memory embeddings.
Built on MTEB~\citep{DBLP:conf/eacl/MuennighoffTMR23}, LMEB provides a scalable, extensible, reproducible evaluation framework and a public leaderboard to track progress in long-horizon memory retrieval.
Using LMEB, we benchmark \textbf{30 embedding models} ranging from 33M to 12B parameters and \textbf{13 rerankers} ranging from 270M to 8B parameters.
Among the embedding models, the best model achieves a Mean (Dataset) nDCG@10 score of 61.41, leaving substantial room for improvement.
Our evaluation reveals four main findings:
\textbf{(1) Model strengths vary across memory types.} No single embedding model performs best across all four types.
\textbf{(2) Larger models do not consistently perform better.} Model size alone is insufficient to predict memory retrieval performance.
\textbf{(3) LMEB and MTEB are only weakly correlated.} Evaluation scores on LMEB correlate weakly with those on the MTEB (eng, v2) retrieval subset (Figure~\ref{fig:corr_main}).
\textbf{(4) Reranking plays a key role in memory retrieval.} Rerankers substantially improve overall performance over the first-stage retriever.
These findings underscore the need for a long-horizon memory embedding benchmark.
We hope our work helps select embedding models and advance future embedding research.

%% file: chapters/lmeb.tex
\begin{figure*}[t]
    \centering
    \includegraphics[width=1.\linewidth]{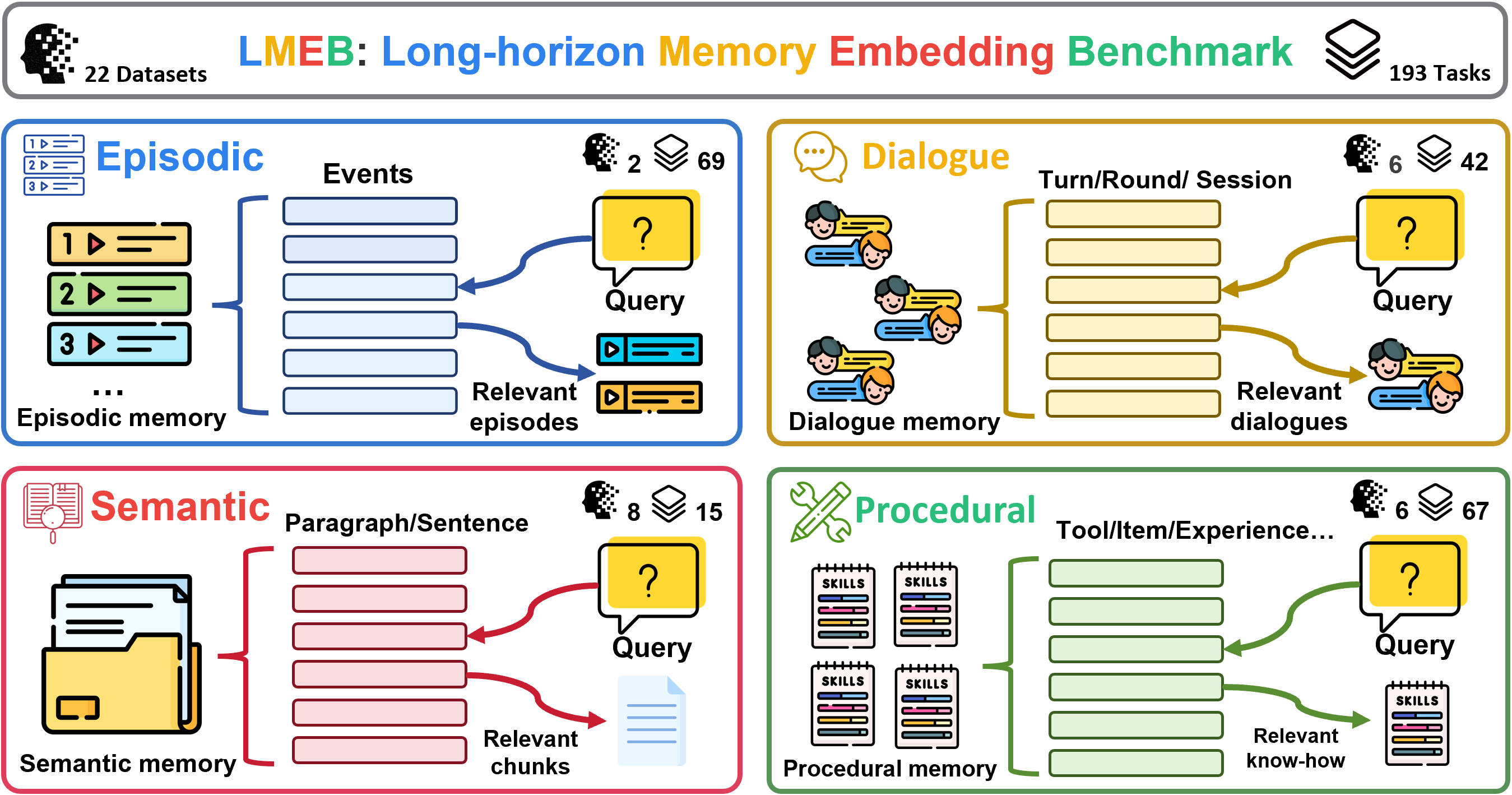}
    \caption{Overview of LMEB. Tables~\ref{tab:episodic_example}, \ref{tab:dialogue_example}, \ref{tab:semantic_example}, and \ref{tab:procedural_example} provide query--document examples for each dataset. Tables~\ref{tab:episodic_tasks}, \ref{tab:dialogue_tasks}, \ref{tab:semantic_tasks}, and \ref{tab:procedural_tasks} summarize task types and example abilities assessed. Dataset descriptions and links, task details, task instructions, and dataset licenses are provided in \S\ref{app:datasets}, \S\ref{app:tasks}, \S\ref{app:instrcutions}, and \S\ref{app:license}, respectively.} 
    \label{fig:framework}
\end{figure*}

\section{The LMEB Benchmark}
\label{sec:lmeb}

\subsection{Desiderata}
\label{sec:lmeb_con_tax}
The \textbf{L}ong-horizon \textbf{M}emory \textbf{E}mbedding \textbf{B}enchmark (\textbf{LMEB}) supports scalable, extensible, and reproducible evaluation of embedding models on long-horizon memory retrieval tasks involving fragmented, context-dependent, and temporally distant memories.
Its design is guided by six desiderata:
\textbf{(1) Diversity.}  LMEB spans diverse memory types, retrieval tasks, and granularities, with queries drawn from diverse sources, memory content spanning multiple domains, and a broad range of query and document lengths.
\textbf{(2) Generalization.} LMEB emphasizes zero-shot evaluation, assessing how embedding models generalize to long-horizon memory retrieval tasks without task-specific fine-tuning.
\textbf{(3) Fidelity.} LMEB preserves available contextual cues, such as timestamps and session or turn identifiers, and restricts retrieval to the memory scope defined for each task.
\textbf{(4) Difficulty.} LMEB includes tasks with varying levels of difficulty, such as retrieving events from specific times, finding information spread across multiple memories, and retrieving past experiences useful for new tasks.
\textbf{(5) Extensibility.} LMEB supports new models through shared interfaces and new datasets and memory retrieval tasks through modular task definitions.
\textbf{(6) Reproducibility.} LMEB provides open-source code and documented task specifications, data processing, and evaluation settings to support reproducible model comparisons.
\begin{figure}[t]
    \centering
    \includegraphics[width=.9\linewidth]{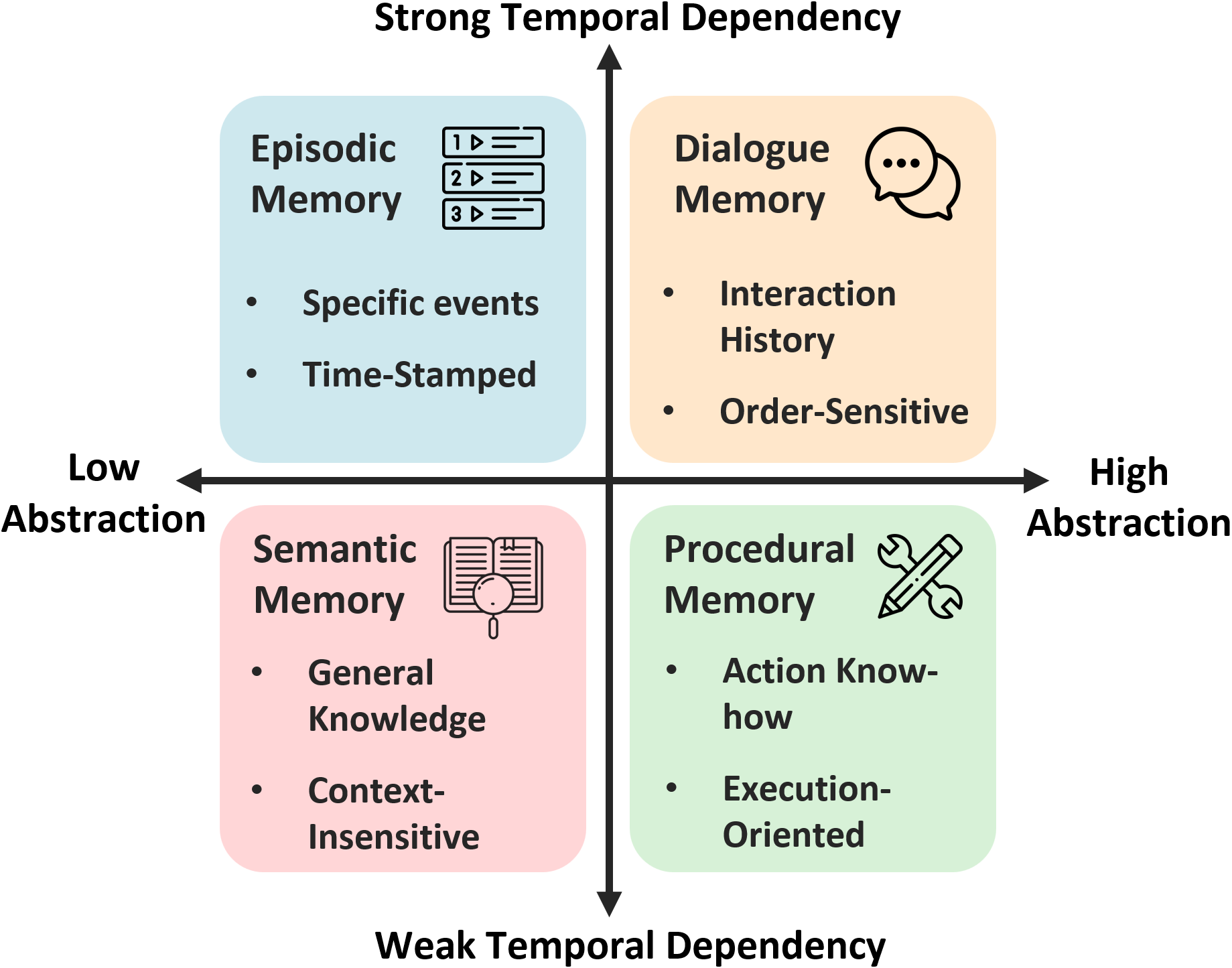}
    \caption{Memory taxonomy of \lmeb.} 
    \label{fig:taxo}
\end{figure}

\subsection{Memory Taxonomy and Tasks}
LMEB comprises 22 English evaluation datasets and 193 zero-shot retrieval tasks spanning four memory types: episodic, dialogue, semantic, and procedural memory~\citep{du2025rethinkingmemoryllmbased}.
Figure~\ref{fig:framework} outlines the four memory types as well as retrieval units, and Table~\ref{tab:dataset_stats} summarizes the dataset statistics.
Figure~\ref{fig:taxo} characterizes these memory types along two conceptual dimensions: \textbf{(\romannumeral1) Level of Abstraction}, which reflects the degree of abstraction in memory representations, and \textbf{(\romannumeral2) Temporal Dependency}, which reflects how strongly retrieval depends on temporal context.
The four types differ in their retrieval units and requirements:
\textbf{(1)~Episodic memory} retrieves concrete events associated with entities, times, and locations. Queries may depend on when events occurred or the order in which they happened.
\textbf{(2) Dialogue memory} retrieves turns, rounds, or sessions from conversation histories. Queries may target specific interactions or information accumulated across them, making conversational and temporal context important for retrieval.
\textbf{(3) Semantic memory} retrieves sentences or paragraphs that provide factual evidence for questions and claims across scientific papers, reports, books, and other knowledge sources.
It focuses on general knowledge across domains, usually with limited temporal dependency.
\textbf{(4) Procedural memory} retrieves tool descriptions, task experiences, trajectories, or items for tool selection, problem-solving, and planning. It emphasizes reusable know-how and experiences applicable to new tasks, typically with limited temporal dependency.
\smallsection{Task Coverage}
Within these memory types, the \textbf{193 tasks} in LMEB cover diverse task requirements, including retrieving the latest information after updates and locating evidence across multiple sessions or passages.
Tasks are instantiated from valid subset--task-type combinations. 
For example, EPBench contains nine subsets with six task types each, yielding 54 tasks.
The benchmark includes queries with \textbf{one or more relevant memories}; for example, TMD has an average of 35.75 relevant memories per query.
The datasets include human-authored and AI-generated content and vary in corpus size and query and document lengths (see Table~\ref{tab:dataset_stats}).
This broad coverage helps identify \textbf{model strengths and weaknesses} across long-horizon memory retrieval tasks and helps guide model selection and future research on memory embeddings.

\input{tables/overall_stats}

\begin{figure}[!t]
    \centering  
    \includegraphics[width=.95\linewidth]{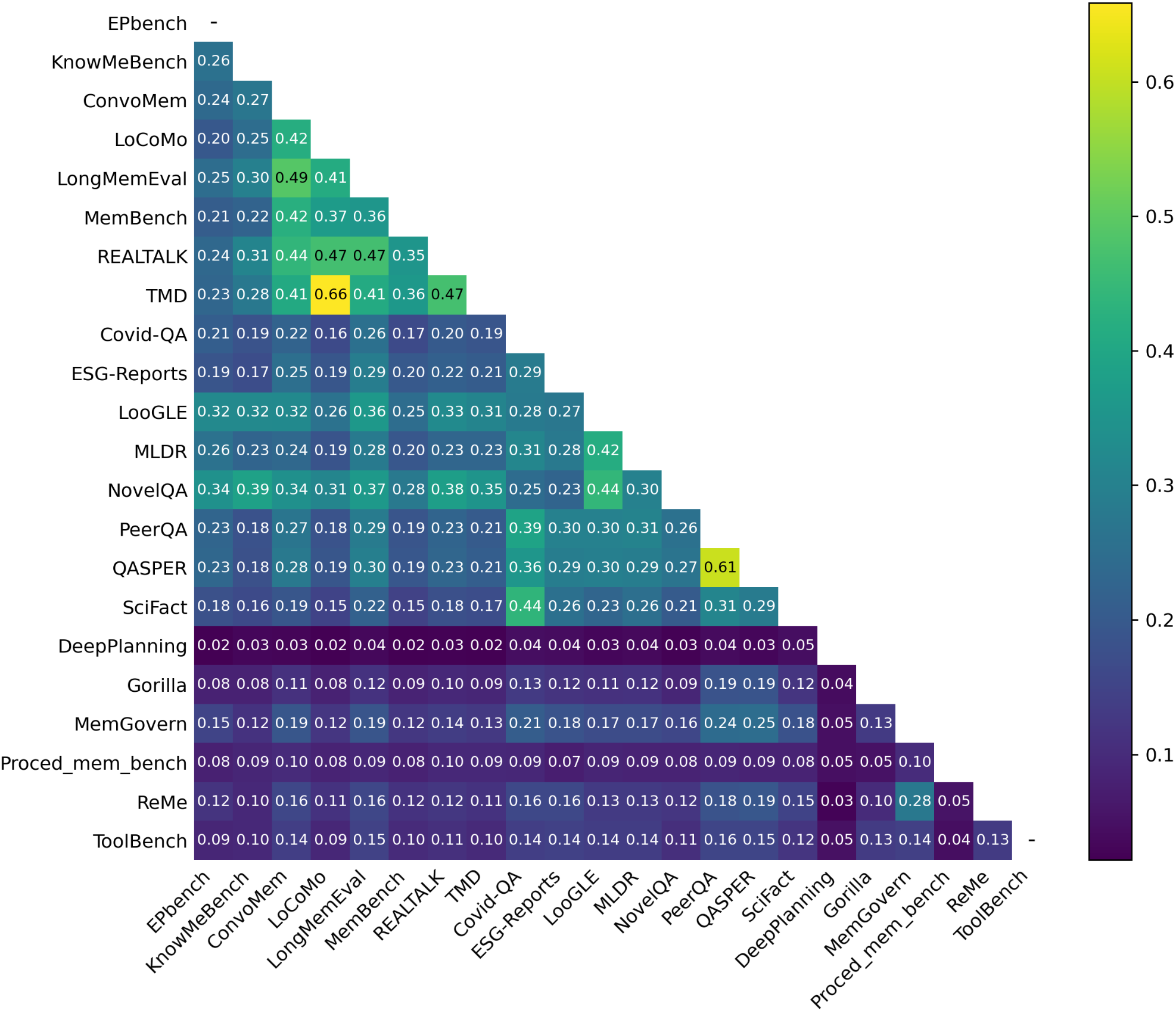}
    \caption{Pairwise lexical similarity among LMEB datasets, measured using weighted Jaccard similarity (JS) between the unigram distributions of their corpora.}
    \label{fig:diversity}
    \label{fig:heatmap}
\end{figure}

\smallsection{Diversity Analysis} 
Following BEIR~\citep{DBLP:journals/corr/abs-2104-08663}, we characterize lexical diversity using pairwise weighted Jaccard similarity (JS)~\citep{DBLP:conf/icdm/Ioffe10} over the unigram word distributions of dataset corpora.
Figure~\ref{fig:diversity} visualizes these similarities, with the formulation provided in \S\ref{app:jaccard}.
The results reveal two main findings:
\noindent\textbf{(1) Lexical variation within memory types.}
Dialogue corpora show relatively similar word distributions, whereas procedural corpora vary across coding, planning, and tool use.
This variation underscores the need to evaluate each memory type across multiple datasets, as any single domain provides only partial coverage.
\textbf{(2) Similar corpora can pose distinct retrieval demands.}
TMD incorporates the LoCoMo corpus but emphasizes queries involving dates, time windows, and specific conversation sessions.
These queries require temporal and interaction context to identify relevant memories, a requirement that lexical similarity between corpora does not capture.
Overall, LMEB covers diverse memory domains and retrieval demands, enabling a comprehensive evaluation of embedding models.
\begin{table*}[t]
    \centering
    \footnotesize
    \setlength{\tabcolsep}{6pt}
    \renewcommand{\arraystretch}{0.95}
    \resizebox{\textwidth}{!}{%
    \begin{tabular}{p{0.96\textwidth}}
        \toprule
        \multicolumn{1}{c}{\textbf{Standardized Data Format in LMEB}} \\
        \midrule
        \rowcolor{gray!10}
        \textbf{Queries (\texttt{queries.jsonl})} \\
        \addlinespace[1pt]
        \texttt{\{"id": "query\_id", "text": "query text"\}} \\
        \midrule
        \rowcolor{gray!10}
        \textbf{Corpus (\texttt{corpus.jsonl})} \\
        \addlinespace[1pt]
        \texttt{\{"id": "corpus\_id", "text": "corpus text", "title": "corpus title"\}} \\
        \midrule
        \rowcolor{gray!10}
        \textbf{Relevance Judgments (\texttt{qrels.tsv})} \\
        \addlinespace[1pt]
        \texttt{query\_id \qquad corpus\_id \qquad 1} \\
        \midrule
        \rowcolor{gray!10}
        \textbf{Candidate Pool (\texttt{candidates.jsonl}, optional)} \\
        \addlinespace[1pt]
        \texttt{\{"scene\_id": "scene\_id", "candidate\_doc\_ids": ["corpus\_id", \ldots]\}} \\
        \bottomrule
    \end{tabular}%
    }
    \caption{Data format in LMEB. The format includes queries, a corpus, and relevance judgments (qrels), with an optional candidate file. The \texttt{query\_id} or its scene prefix is matched to \texttt{scene\_id} to identify the candidate pool.}
    \label{tab:unified_data_format}
\end{table*}
\subsection{Benchmark Construction}
\noindent\textbf{Data Curation.}
We adapt source datasets to memory retrieval tasks through dataset-specific filtering, retrieval unit selection, and relevance mapping.
For PeerQA, we exclude questions whose corresponding paper text is unavailable; for SciFact, we remove claims without annotated evidence.
We split NovelQA and LooGLE documents into passages without breaking up evidence spans and label passages containing the evidence as relevant to the corresponding queries.
For Proced\_mem\_bench, only trajectories with source relevance scores of at least 7.0 are treated as relevant.
We further assess the source relevance annotations using GPT-5.6-sol with high reasoning effort as an \textbf{LLM judge} and conduct \textbf{manual spot checks} on up to 100 annotated query--memory pairs per dataset.
Both assessments indicate that the reviewed annotations are reliable.
Each relevant query--memory pair is then assigned a label of $1$.
The common format contains \texttt{queries.jsonl}, \texttt{corpus.jsonl}, and \texttt{qrels.tsv}, together with an optional \texttt{candidates.jsonl} file (Table~\ref{tab:unified_data_format}).
Dataset-specific filtering and annotation procedures are documented in~\S\ref{app:datasets} to ensure that \textbf{the construction is transparent and reproducible}.
\smallsection{Contextualization}
LMEB incorporates temporal and interaction information to contextualize memories and queries.
\textbf{At the memory level}, we retain available timestamps and applicable session, round, and turn identifiers in the \texttt{title} or \texttt{text} fields.
\textbf{At the query level}, we add a reference time, when available, to clarify relative temporal expressions such as \emph{two days ago}.
Together, these contextual cues clarify temporal and conversational references in queries, helping models identify memories relevant to the specified time or interaction.
\smallsection{Delimitation}
LMEB enforces memory scope constraints where required by the task, such as limiting retrieval to the conversation history associated with a query.
The optional \texttt{candidates.jsonl} file specifies the identifiers of the admissible memory items for each query or scenario (Table~~\ref{tab:unified_data_format}).
When a scope constraint applies, only memories in the specified candidate pool are eligible for retrieval; otherwise, retrieval uses the full task corpus.
\subsection{Evaluation Protocol}
\label{sec:evaluation_protocol}
Models are evaluated on LMEB without task-specific fine-tuning.
For each query, an embedding model ranks memories in the admissible corpus by vector similarity.
Rerankers rescore and reorder the memories retrieved by the first-stage embedding model, using the same relevance judgments for evaluation.
LMEB supports a range of retrieval metrics, including nDCG@$k$, MRR@$k$, MAP@$k$, precision@$k$, and recall@$k$ at arbitrary top-$k$ cutoffs.
Following MTEB~\citep{DBLP:conf/eacl/MuennighoffTMR23}, we use nDCG@10 as the primary metric.
It accounts for both relevance and ranking positions.
\smallsection{Scoring}
Each dataset score is computed by averaging scores across its retrieval tasks.
\textbf{Mean (Dataset)} is the arithmetic mean of all 22 LMEB dataset scores.
For each memory type, we average the scores of its constituent datasets; \textbf{Mean (Type)} is then the arithmetic mean of these four type scores.
Evaluation scores are reported on a 0--100 scale and rounded only after aggregation.
\smallsection{Evaluation Framework}
LMEB is {officially supported} by the open-source MTEB framework~\citep{DBLP:conf/eacl/MuennighoffTMR23}, enabling scalable, extensible, and reproducible evaluation.
\textbf{As shown in the Python code example}, evaluation can be launched with only a few lines of code.
The framework supports model implementations based on Transformers, Sentence-Transformers, and vLLM.
Modular interfaces allow new models and datasets to be integrated with minimal additional code.
Results can be submitted to the public leaderboard to track progress in long-horizon memory retrieval.
\begin{promptblock}[
    colback=black!88.2,
    colframe=black!76.5,
    boxrule=0.4pt,
    arc=2pt,
    boxsep=0pt,
    left=8pt,
    right=8pt,
    top=7pt,
    bottom=7pt,
    before skip=8pt,
    after skip=8pt
]
\begin{lstlisting}[
    language=Python,
    basicstyle={
        \fontencoding{T1}\ttfamily\footnotesize
        \linespread{1.12}\selectfont
        \color[HTML]{D4D4D4}
    },
    upquote=true,
    backgroundcolor={\color[HTML]{1E1E1E}},
    keywordstyle={\color[HTML]{C586C0}},
    stringstyle={\color[HTML]{CE9178}},
    commentstyle={\color[HTML]{6A9955}},
    morekeywords={[2]get_model,get_benchmark,evaluate},
    keywordstyle={[2]\color[HTML]{DCDCAA}},
    morekeywords={[3]mteb,model,model_name_or_path,benchmark,results},
    keywordstyle={[3]\color[HTML]{9CDCFE}},
    columns=fullflexible,
    keepspaces=true,
    showstringspaces=false,
    breaklines=true,
    numbers=none,
    frame=none,
    aboveskip=0pt,
    belowskip=0pt
]
import mteb

model = mteb.get_model("BAAI/bge-m3")
benchmark = mteb.get_benchmark("LMEB")
results = mteb.evaluate(model, benchmark)
\end{lstlisting}
\end{promptblock}

%% file: tables/overall_stats.tex
\begin{table*}[t!]
    \small
    \renewcommand\arraystretch{1.2}
    \resizebox{\textwidth}{!}{\begin{tabular}{ l | l | c | c | c  | c | c  c  c c | c c }
        \toprule
         \multicolumn{1}{l}{\textbf{Setting} ($\rightarrow$)} &
         \multicolumn{5}{c}{\bf Dataset Information} &
         \multicolumn{4}{c}{\textbf{Dataset statistics}}    &
         \multicolumn{2}{c}{\textbf{Avg.~Length (words)}} \\
         \cmidrule(lr){1-1}
         \cmidrule(lr){2-6}
         \cmidrule(lr){7-10}
         \cmidrule(lr){11-12}
           \textbf{Memory Type ($\downarrow$)} &\textbf{Dataset ($\downarrow$)} & \textbf{Granularity ($\downarrow$)} & \textbf{\# Tasks} & \textbf{$\text{Query}_{src}$} & \textbf{$\text{Corpus}_{src}$} & \textbf{\#Query} & \textbf{\#Corpus} & \textbf{\#Qrels}  & \textbf{Avg. D~/~Q} & \textbf{Query} & \textbf{Document} \\
         \midrule
     \multirow{3}{*}{\textbf{Episodic Memory}}  & EPBench~\citep{DBLP:conf/iclr/HuetB025} & Event & 54 & AI & AI  &  3,644  &   2,838  & 11,458 & 3.144 &  24.07 & 410.69 \\ 
       & KnowMeBench~\citep{wu2026knowme} & Event &  15 & AI &  Human & 2,162 &  27,062   &     4,840     & 2.239 &  31.80  & 58.68 \\ 
        \rowcolor{gray!18}
    \cellcolor{white!20}  & Total & Event &  69 & AI & Hybrid  & 5,806  &  29,900   &   16,298       &  2.807 &  - & -  \\ 
    \hline 
    \multirow{7}{*}{\textbf{Dialogue Memory}}  & LoCoMo~\citep{DBLP:conf/acl/MaharanaLTBBF24} & Turn & 5 & AI & AI~\&~Human  & 1,976 &  5,882   &    2,801      & 1.418 & 10.36  & 38.73 \\ 
       & LongMemEval~\citep{DBLP:conf/iclr/WuWYZCY25} & Session & 6 & AI~\&~Human & AI &  500 &   237,655  &     948     & 1.896 &  15.89 & 242.97 \\ 
       & REALTALK~\citep{DBLP:journals/corr/abs-2502-13270} & Turn & 3 & Human & Human  & 679 &  8,944  &   1,553       & 2.287 & 8.990  & 34.20 \\ 
       & TMD~\citep{DBLP:journals/corr/abs-2406-00057} & Turn & 12 & AI~\&~Human & AI~\&~Human  & 2,134 & 7,463    &  76,287  & 35.748 & 15.66  & 45.85 \\ 
       & MemBench~\citep{DBLP:conf/acl/Tan000DD25} & Round & 10 & AI & AI  &  10,000 &   929,115  &     29,592     & 2.959 &  10.41 & 42.92  \\ 
       & ConvoMem~\citep{DBLP:journals/corr/abs-2511-10523} & Turn & 6 & AI & AI  & 5,867 &   500,221  & 13,779 & 2.349 & 23.19  & 27.33 \\ 
       \rowcolor{gray!18}
    \cellcolor{white!20} & Total & Multi-granularity &   42 & Hybrid &  Hybrid & 21,156 &  1,689,280   &    124,960      & 5.907 &  - & - \\ 
    \hline 
    \multirow{9}{*}{\textbf{Semantic Memory}}  & QASPER~\citep{DBLP:conf/naacl/DasigiLBCSG21} & Paragraph & 1 & Human & Human  & 1,335 &   65,300  & 2,319 & 1.737 & 7.934  & 77.45 \\ 
    & NovelQA~\citep{DBLP:conf/iclr/WangNPWGDBH0WZ25} & Paragraph & 7 & Human & Human  & 1,541 & 79,286 & 2,506 & 1.626 & 19.79  & 139.51 \\ 
    & PeerQA~\citep{DBLP:conf/naacl/BaumgartnerBG25} & Sentence & 1 & Human & Human  & 136 & 18,593 & 389 & 2.860 & 15.65 & 24.67 \\ 
    & Covid-QA~\citep{moller-etal-2020-covid} & Paragraph & 1 & Human & Human & 1,111 & 3,351 & 1,111 & 1.0 & 9.541 & 110.91 \\ 
    & ESG-Reports~\citep{DBLP:journals/corr/abs-2505-17166} & Paragraph & 1 & Human & Human  & 36 & 2,407 & 75 & 2.083 & 9.389 & 129.35 \\
    & MLDR~\citep{DBLP:journals/corr/abs-2402-03216} & Paragraph & 1 & AI & Human  & 100 & 1,536 & 100 & 1.0 & 11.34  & 112.94 \\ 
    & LooGLE~\citep{DBLP:conf/acl/LiWZZ24} & Paragraph & 2 & AI~\&~Human & Human  & 3,052 & 28,190 & 5,176 & 1.696 & 13.72 & 164.82 \\ 
    & SciFact~\citep{DBLP:conf/emnlp/WaddenLLWZCH20} & Sentence & 1 & Human & Human & 188 & 1,748 & 366 & 1.947 & 12.89  & 39.93 \\
    \rowcolor{gray!18}
    \cellcolor{white!20}  & Total & Multi-granularity &  15 & Hybrid &  Human & 7,499 & 200,411 & 12,042 &  1.606 &  - & - \\ 
    \hline 
     \cellcolor{white!20} \multirow{7}{*}{\textbf{Procedural Memory}} 
     & Gorilla~\citep{DBLP:conf/nips/PatilZ0G24} & Tool & 3 & AI & Human & 598 & 1,005 & 598 & 1.0 & 22.41 & 146.83 \\ 
     & ToolBench~\citep{DBLP:conf/iclr/QinLYZYLLCTQZHT24} & Tool & 1 & AI & Human & 1,100 & 13,862 & 2,629 & 2.39 & 46.06 & 87.93 \\ 
     & ReMe~\citep{cao2025remember} & Experience & 9 & AI & AI  & 1,217 & 914 & 1,217 & 1.0 & 13.51 & 47.36 \\ 
     & Proced\_mem\_bench~\citeyear{DBLP:journals/corr/abs-2511-21730} & Trajectory & 3 & Human  & AI & 40 & 336 & 529 & 13.225 & 8.175 & 362.94 \\ 
     & MemGovern~\citep{wang2026memgovern} & Experience & 48 &  AI & AI & 121,475 & 121,475 & 121,475 & 1.0 & 18.59 & 104.01 \\ 
     & DeepPlanning~\citep{zhang2026deepplanning} & Item & 3 &  AI~\&~Human & AI & 120 & 19,839 & 515 & 4.292 & 161.55 & 127.43 \\ 
    \rowcolor{gray!18}
    \cellcolor{white!20} & Total & Multi-granularity & 67 & Hybrid & Hybrid & 124,550 & 157,431 & 126,963 & 1.019 & - & - \\ 
    \bottomrule
    \end{tabular}}
    \caption{Dataset statistics for LMEB. $\text{Query}_{src}$ and $\text{Corpus}_{src}$ denote the sources of queries and corpus documents, respectively; ``AI'' and ``Human'' indicate AI-generated and human-authored content, respectively. Qrels stands for \underline{q}uery \underline{rel}evance judgment\underline{s}, and Avg. D~/~Q denotes the average number of relevant documents per query.}
    \label{tab:dataset_stats}
\end{table*}

%% file: chapters/setup.tex
\section{Experimental Setup}
Experimental settings, including the benchmarked models and implementation details, are provided in \S\ref{app:setup}.
Task-specific instructions are provided in \S\ref{app:instrcutions}.

%% file: chapters/results.tex
\input{tables/main_ret_best}

\input{tables/main_ret_rerank}

\section{Evaluation}

\subsection{Main Results}
\label{sec:main_rets}
Table~\ref{table:main_best_inst} reports nDCG@10 results for 30 embedding models using the instruction settings
specified in Table~\ref{tab:model_links}.
The results reveal three main findings:
\textbf{(1) No single model leads across all memory types.}
bge-multilingual-gemma2 achieves the highest Mean (DS) score (61.41) and leads dialogue retrieval.
KaLM-Embedding-Gemma3 leads episodic and procedural retrieval, while NV-Embed-v2 leads semantic retrieval.
Model selection should therefore consider the target memory type.
\textbf{(2)  Larger models do not consistently perform better.}
stella\_en\_400M\_v5 (435M) achieves a higher Mean (DS) score than Qwen3-Embedding-8B (57.88 vs.\ 55.94).
Within the Qwen3-Embedding series, the 0.6B model also exceeds the 4B model (54.71 vs.\ 53.23).
\textbf{(3) Dialogue memory retrieval is a common weakness.}
Dialogue is the lowest-scoring memory type for 24 of the 30 models, with a best score of 59.60.
This finding highlights dialogue memory retrieval as a priority for further improvement.

\subsection{Reranking Performance}
\label{sec:reranking_results}
Table~\ref{tab:main_lmeb_rerank} compares the first-stage retriever, KaLM-Embedding-V2.5, with 13 second-stage rerankers across the four memory types.
For each query, all rerankers rescore and reorder the same set of up to 100 candidate memories retrieved by the
first-stage embedding model.
\textbf{(1) Reranking yields substantial overall gains.}
All 13 rerankers improve Mean (DS) over the first-stage score of 54.84.
Lychee-Reranker-Large achieves the highest Mean (DS) score of 68.65, improving on the first-stage retriever by 13.81 points.
\textbf{(2) Compact rerankers remain competitive.}
Lychee-Reranker-Nano achieves a Mean (DS) score of 63.22 with 0.27B active decoder parameters, close to the 63.68 achieved by the 2.5B bge-reranker-v2-gemma.
Strong reranking performance is therefore attainable with fewer than 0.5B active parameters.
\textbf{(3) Reranking substantially improves dialogue memory retrieval.}
Qwen3-Reranker-8B increases the dialogue memory retrieval score from 50.80 to 66.42, a substantial improvement of 15.62 points.

\input{tables/lmeb_mteb_correlation}

\subsection{In-depth Analysis}
\label{sec:in_depth_analysis}
\begin{figure*}[t]
    \centering
    \includegraphics[width=1.0\textwidth]{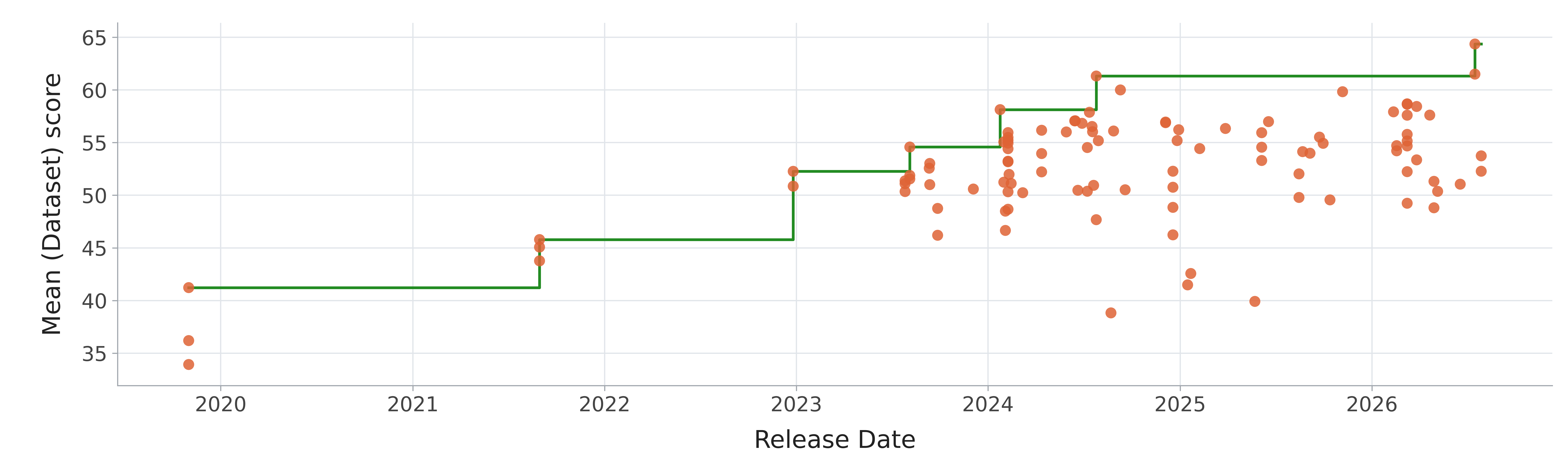}
    \vspace{-0.3cm}
    \caption{Retrieval model performance on LMEB by release date, measured by nDCG@10. Each marker is a model at its release date; the step line traces the running best Mean (Dataset) score achieved up to each date.}
    \label{fig:lmeb_perf_release_date}
\end{figure*}

\begin{figure}[t]
    \centering      
     \subfigure{
        \includegraphics[width=.45\linewidth]{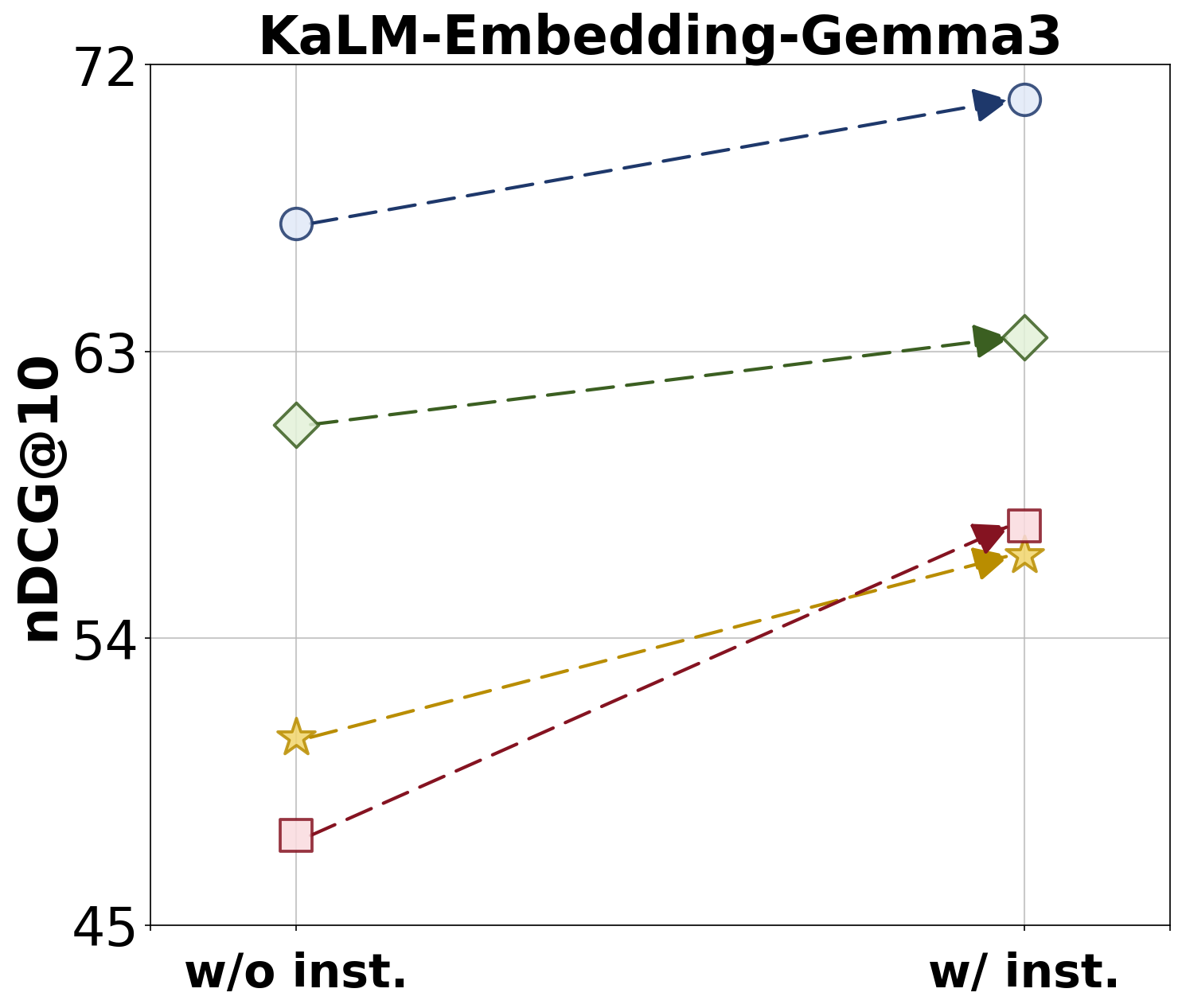}\label{fig:KaLM-Embedding-Gemma3_trend}}
    \subfigure{
        \includegraphics[width=.45\linewidth]{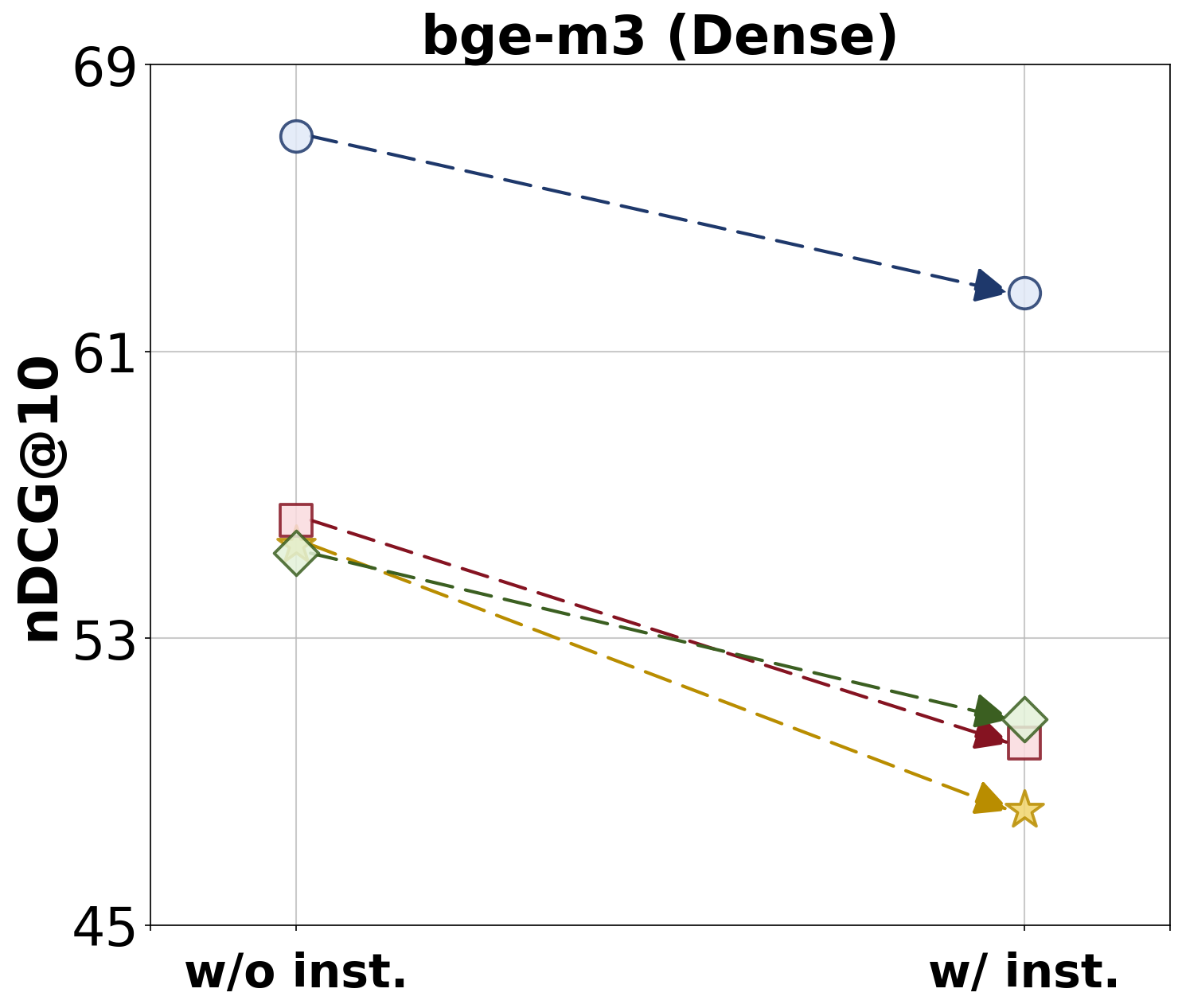}\label{fig:bge-m3_trend}}
    \vspace{-0.2cm}
    \subfigure{
        \includegraphics[width=.45\linewidth]{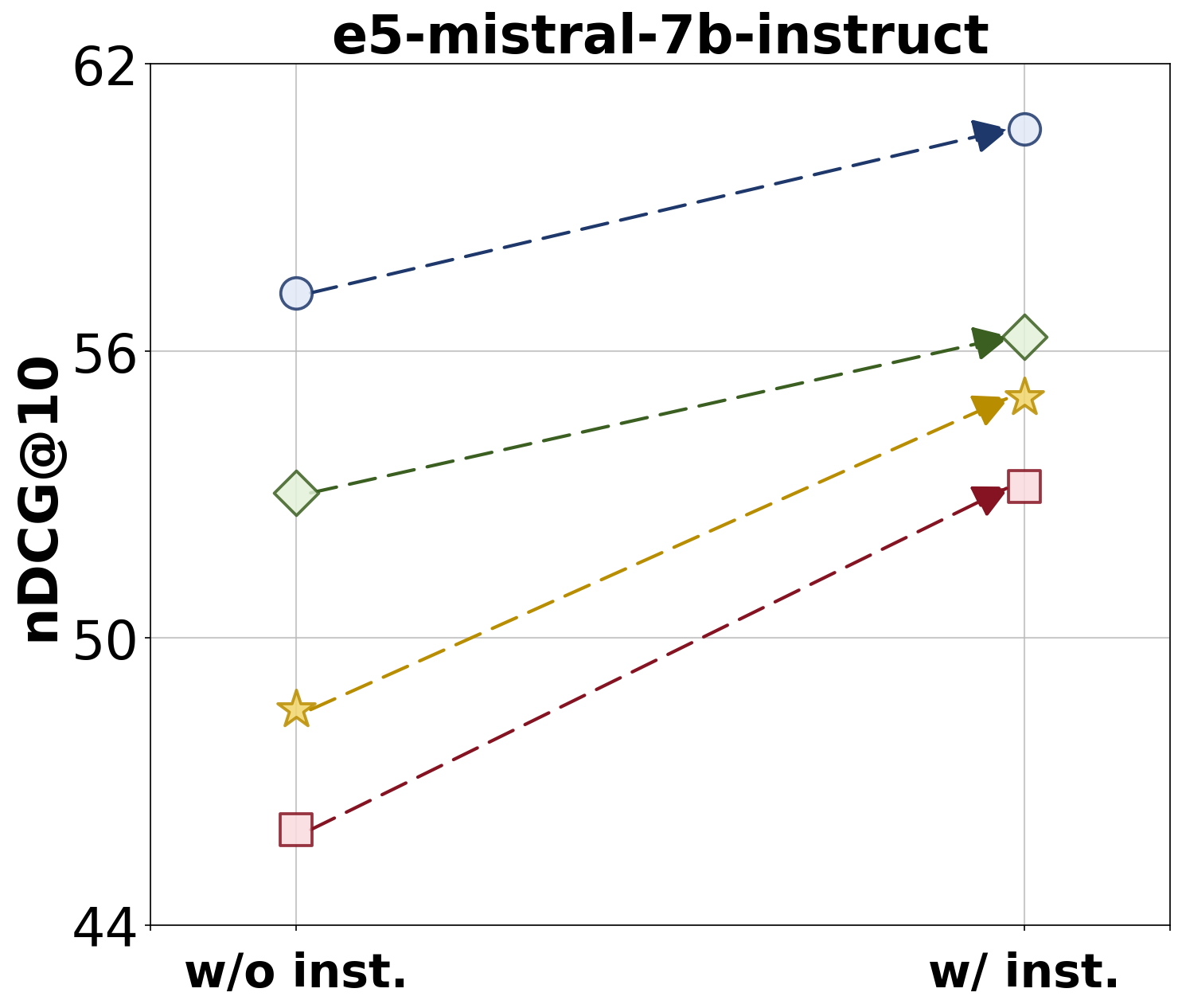}\label{fig:e5-mistral-7b-instruct_trend}}
    \subfigure{
        \includegraphics[width=.45\linewidth]{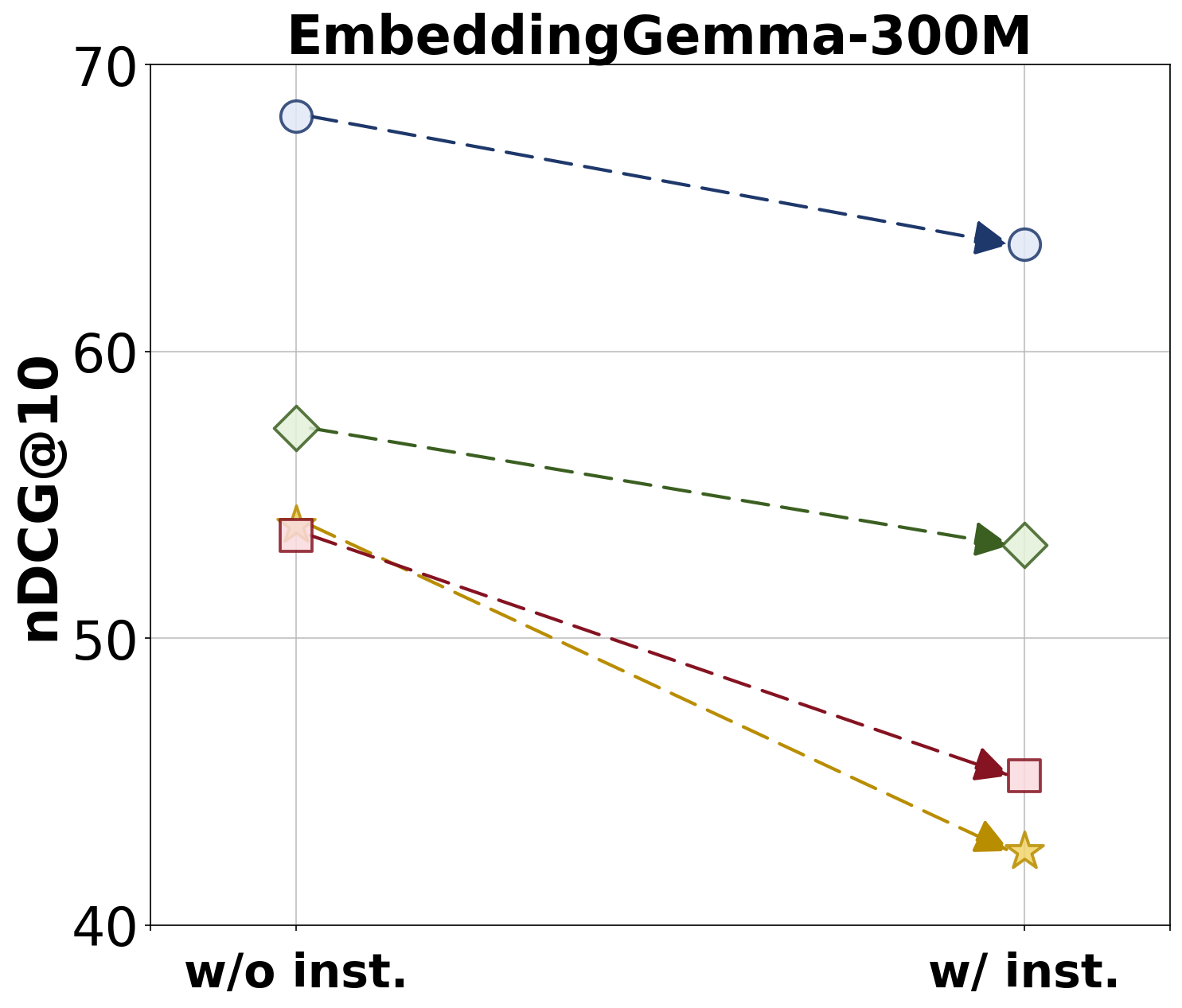}\label{fig:EmbeddingGemma-300M_trend}}
    \vspace{-0.2cm}
    \subfigure{
        \includegraphics[width=.45\linewidth]{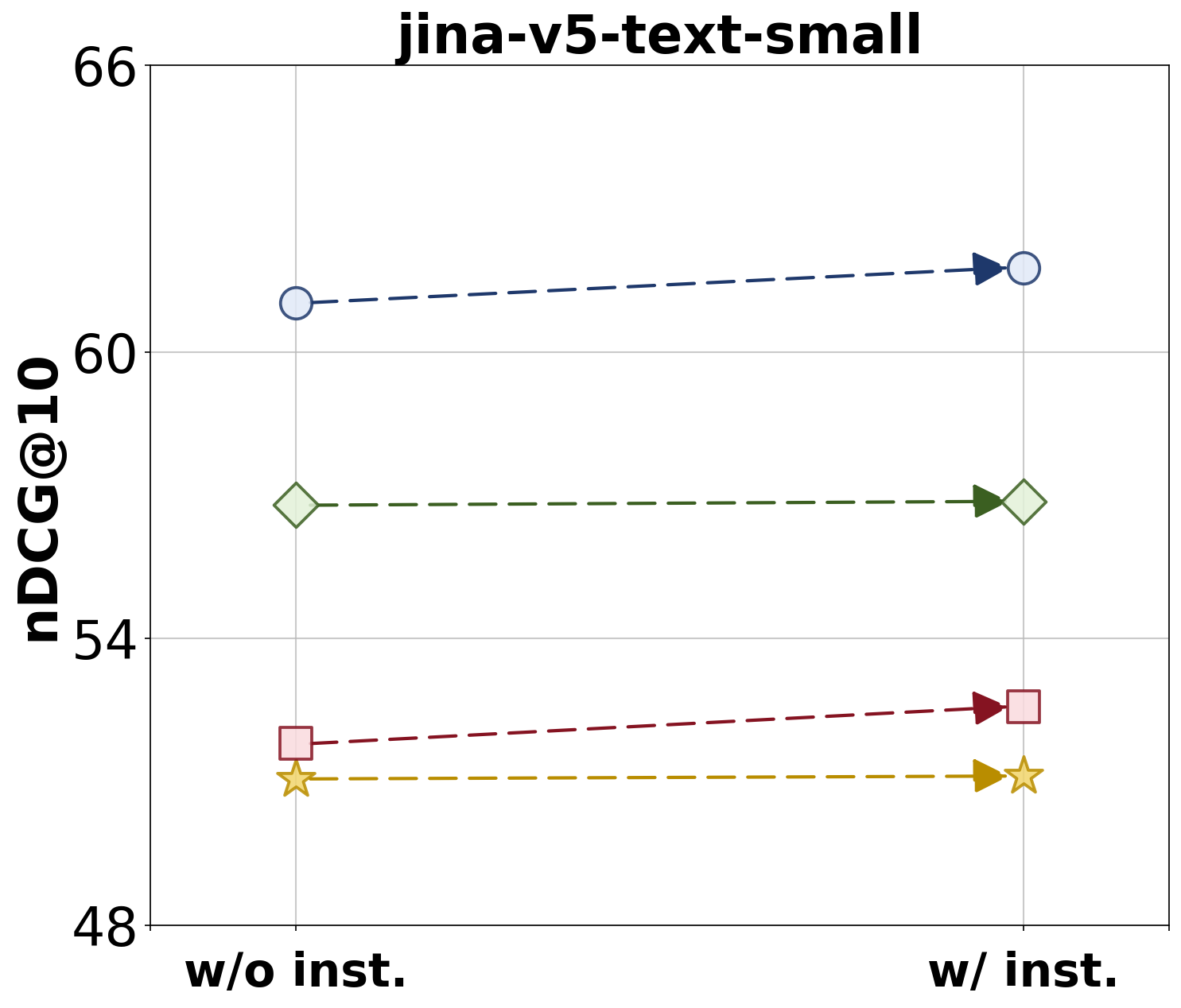}\label{fig:jina-v5-text-small_trend}}
    \subfigure{
        \includegraphics[width=.45\linewidth]{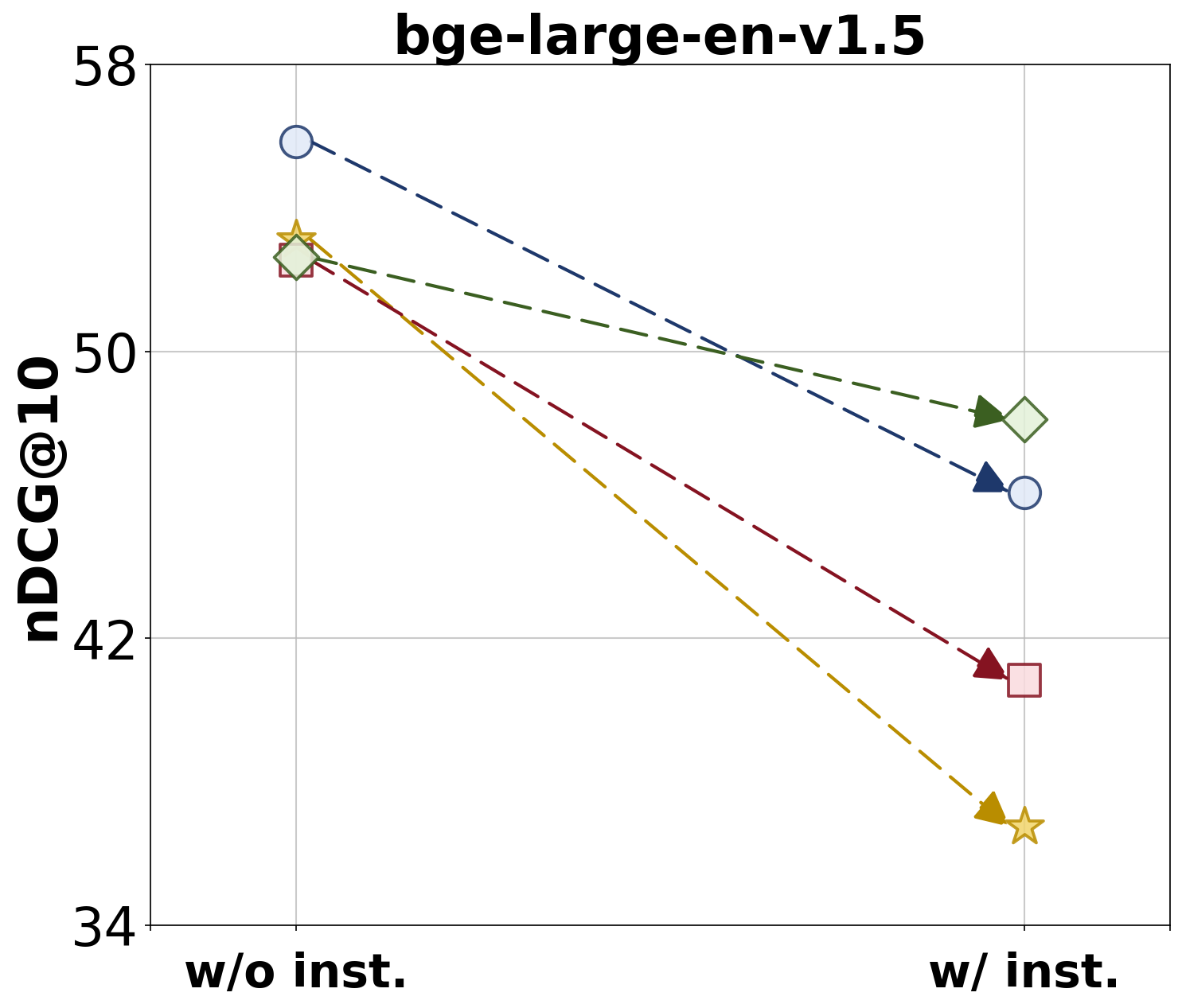}\label{fig:bge-large-en-v1.5_trend}}
    \vspace{-0.2cm}
    \subfigure{
        \includegraphics[width=0.9\linewidth]{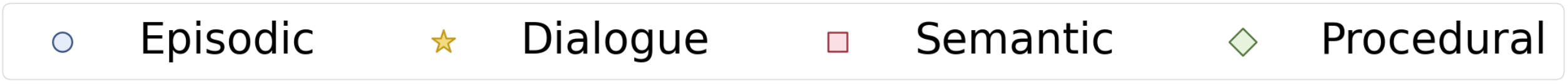}\label{fig:legend_trend}}
    \caption{Performance comparison between the \textbf{{w/o inst.}} and \textbf{{w/ inst.}} settings. In the \textbf{{w/o inst.}} setting, the model encodes only the query; in the \textbf{{w/ inst.}} setting, it encodes the instruction concatenated with the query.}
    \label{fig:inst_trend}
\end{figure}

We further examine correlations with MTEB, performance trends across model releases, and the effects of task instructions.
\textbf{(1) LMEB and MTEB are only weakly correlated.}
Table~\ref{tab:lmeb_mteb_correlation} and Figure~\ref{fig:corr_main} show that, among the evaluated models, overall LMEB scores correlate weakly with scores on the MTEB (eng, v2) retrieval subset ($r=0.267$, $\rho=0.261$).
Only procedural memory shows statistically significant correlations, as shown in Figure~\ref{fig:corr_by_memory_type}.
These results suggest that LMEB and MTEB emphasize different capabilities, highlighting LMEB's value in evaluating long-horizon memory retrieval.
\noindent\textbf{(2) Memory retrieval is improving but remains challenging.}
Figure~\ref{fig:lmeb_perf_release_date} tracks the performance of more than 100 retrieval models on LMEB by release date.
The best score increases over time, but even the strongest models leave substantial room for improvement.
Recent models vary widely in performance, suggesting that newer releases are not always better.
\textbf{(3) Instructions can help or hinder retrieval.}
Figure~\ref{fig:inst_trend} shows gains for KaLM-Embedding-Gemma3 and e5-mistral-7b-instruct, declines for bge-m3 (Dense), EmbeddingGemma-300M, and bge-large-en-v1.5, and little change for jina-v5-text-small.
Task instruction sensitivity varies across models, so instruction use should be assessed for each model.

%% file: tables/main_ret_best.tex
\begin{table*}[!t]
\centering
\LARGE
\renewcommand\arraystretch{1.2}
\setlength\tabcolsep{6pt}
\begin{adjustbox}{width=\textwidth,center}
\begin{tabular}{l cc| cccc | cc}
\toprule
\textbf{Model} &
\textbf{Size} &
\textbf{Dim} &
\textbf{Episodic} &
\textbf{Dialogue} &
\textbf{Semantic} &
\textbf{Procedural} &
\textbf{Mean (DS)} &
\textbf{Mean (TY)} \\
\hline
\multicolumn{3}{l|}{Num. Datasets ($\rightarrow$)} &
2 & 6 & 8 & 6 & 22 & 22 \\
\midrule
\multicolumn{9}{c}{\textbf{Embedding Models $>1$B parameters}} \\
\hline
KaLM-Embedding-Gemma3 & 12B & 3840 & \textbf{70.89} & \underline{56.59} & 57.53 & \textbf{63.43} & 60.10 & \underline{62.11} \\
bge-multilingual-gemma2 & 9B & 3584 & \underline{70.88} & \textbf{59.60} & \underline{60.41} & \underline{61.40} & \textbf{61.41} & \textbf{63.07} \\
Qwen3-Embedding-8B & 8B & 4096 & 60.85 & 51.69 & 55.51 & 59.12 & 55.94 & 56.79 \\
F2LLM-v2-8B & 8B & 4096 & 66.11 & 55.26 & 57.38 & 61.22 & 58.64 & 59.99 \\
NV-Embed-v2 & 7B & 4096 & 68.45 & 56.42 & \textbf{62.18} & 58.77 & \underline{60.25} & 61.46 \\
e5-mistral-7b-instruct & 7B & 4096 & 60.64 & 55.03 & 53.16 & 56.30 & 55.21 & 56.28 \\
SFR-Embedding-Mistral & 7B & 4096 & 61.50 & 56.38 & 58.66 & 58.04 & 58.13 & 58.64 \\
SFR-Embedding-2\_R & 7B & 4096 & 63.12 & 55.10 & 56.53 & 57.72 & 57.06 & 58.12 \\
Qwen3-Embedding-4B & 4B & 2560 & 57.24 & 43.58 & 54.52 & 59.81 & 53.23 & 53.79 \\
F2LLM-v2-4B & 4B & 2560 & 66.45 & 55.55 & 57.85 & 60.34 & 58.68 & 60.05 \\
Giga-Embeddings-instruct & 3B & 2048 & 68.05 & 51.26 & 54.03 & 57.59 & 55.52 & 57.73 \\
\hline
\multicolumn{9}{c}{\textbf{Embedding Models $<1$B parameters}} \\
\hline
jina-v5-text-small & 596M & 1024 & 61.76 & 51.12 & 52.57 & 56.87 & 54.18 & 55.58 \\
Qwen3-Embedding-0.6B & 596M & 1024 & 59.54 & 50.97 & 52.67 & \textbf{59.56} & 54.71 & 55.69 \\
snowflake-arctic-embed-l-v2.0 & 568M & 1024 & 65.50 & 53.52 & \underline{57.27} & 56.88 & 56.89 & 58.29 \\
multilingual-e5-large-instruct & 560M & 1024 & 59.77 & 54.62 & 53.86 & 55.54 & 55.06 & 55.95 \\
bge-m3 (Dense) & 560M & 1024 & 67.00 & \underline{55.61} & 56.29 & 55.37 & 56.83 & \underline{58.57} \\
KaLM-Embedding-V2.5 & 494M & 896 & 62.70 & 50.78 & 54.72 & 56.72 & 54.92 & 56.23 \\
KaLM-Embedding-V1 & 494M & 896 & 60.21 & 51.60 & 54.81 & 57.70 & 55.21 & 56.08 \\
stella\_en\_400M\_v5 & 435M & 1024 & \underline{67.42} & \textbf{55.84} & 56.58 & \underline{58.46} & \textbf{57.88} & \textbf{59.57} \\
bge-large-en-v1.5 & 335M & 1024 & 55.85 & 53.12 & 52.55 & 52.62 & 53.02 & 53.54 \\
e5-large-v2 & 335M & 1024 & 63.43 & 54.76 & 56.05 & 54.52 & 55.95 & 57.19 \\
granite-embedding-311m-ml-r2 & 311M & 768 & 58.17 & 45.56 & 51.37 & 54.75 & 51.32 & 52.46 \\
EmbeddingGemma-300M & 308M & 768 & \textbf{68.19} & 53.94 & 53.58 & 57.32 & 56.03 & 58.26 \\
snowflake-arctic-embed-m-v2.0 & 305M & 768 & 64.86 & 54.31 & \textbf{57.39} & 56.38 & \underline{56.95} & 58.23 \\
jina-v5-text-nano & 212M & 768 & 58.00 & 50.50 & 54.99 & 54.27 & 53.84 & 54.44 \\
nomic-embed-text-v1 & 137M & 768 & 61.37 & 49.26 & 56.80 & 56.44 & 55.06 & 55.97 \\
multilingual-e5-small & 118M & 384 & 57.70 & 41.18 & 54.13 & 51.98 & 50.33 & 51.24 \\
bge-base-en-v1.5 & 109M & 768 & 52.08 & 50.67 & 52.59 & 54.62 & 52.57 & 52.49 \\
bge-small-en-v1.5 & 33M & 384 & 55.09 & 47.81 & 51.49 & 52.23 & 51.01 & 51.65 \\
e5-small-v2 & 33M & 384 & 60.63 & 50.41 & 54.60 & 51.77 & 53.23 & 54.35 \\
\bottomrule
\end{tabular}
\end{adjustbox}
\caption{Performance of embedding models on LMEB. Per-dataset results are reported in Table~\ref{tab:lmeb_embedding_dataset_results}. Mean (DS) denotes the average score across all 22 datasets, while Mean (TY) denotes the average score across the four memory types. Within each parameter group, the best results are \textbf{boldfaced}, and the second-best results are \underline{underlined}.}
\label{table:main_best_inst}
\end{table*}

%% file: tables/main_ret_rerank.tex
\begin{table*}[t]
    \centering
    \renewcommand\arraystretch{0.96}
    \setlength{\tabcolsep}{4pt}
    \resizebox{\textwidth}{!}{%
    \begin{tabular}{lccc|cccc}
        \toprule
        \textbf{Models} &
        \textbf{Size} &
        \shortstack{\textbf{Mean} \textbf{(DS)}} &
        \shortstack{\textbf{Mean} \textbf{(TY)}} &
        \textbf{Episodic} &
        \textbf{Dialogue} &
        \textbf{Semantic} &
        \textbf{Procedural} \\
        \midrule
        \multicolumn{8}{c}{\textbf{First-stage Retriever}} \\
        \midrule
        KaLM-Embedding-V2.5 &494M &
        54.84 & 55.99 & 61.75 & 50.80 & 54.82 & 56.59 \\
        \midrule
        \multicolumn{8}{c}{\textbf{Second-stage Reranker}} \\
        \midrule
        \rowcolor{gray!10}
        \multicolumn{8}{l}{\emph{Models with more than 1B parameters}} \\
        Qwen3-Reranker-8B & 8B &
        \underline{68.51} & \underline{70.19} & \underline{78.72} & \textbf{66.42} & \underline{68.78} & \underline{66.84} \\
        Qwen3-Reranker-4B & 4B &
        67.06 & 68.58 &
        76.79 & 62.86 &
        68.25 & 66.42 \\
        Lychee-Reranker-Large$^{\dagger}$ & 4B &
        \textbf{68.65} & \textbf{70.39} &
        \textbf{79.35} & 65.71 &
        \textbf{69.09} & \textbf{67.43} \\
        bge-reranker-v2-gemma & 2.5B &
        63.68 & 65.26 & 73.72 & 63.50 & 64.79 & 59.01 \\
        mxbai-rerank-large-v2 & 1.5B &
        66.06 & 67.79 & 76.36 & \underline{66.08} & 65.90 & 62.82 \\
        \midrule
        \rowcolor{gray!10}
        \multicolumn{8}{l}{\emph{Models with 0.5B--1B parameters}} \\
        Lychee-Reranker-Small$^{\dagger}$ & 1B &
        \textbf{66.35} & \textbf{68.08} &
        \textbf{76.80} & \underline{63.65} &
        \textbf{66.43} & \textbf{65.45} \\
        bge-reranker-large & 0.6B &
        59.14 & 61.26 & 71.53 & 60.64 & 58.41 & 54.47 \\
        bge-reranker-v2-m3 & 0.6B &
        63.76 & 65.58 & \underline{75.27} & 62.38 & 64.94 & 59.72 \\
        Qwen3-Reranker-0.6B & 0.6B &
        \underline{65.59} & \underline{67.09} &
        74.39 & \textbf{63.69} &
        \underline{65.27} & \underline{65.00} \\
        mxbai-rerank-base-v2 & 0.5B &
        63.38 & 65.28 & 74.83 & 62.47 & 63.49 & 60.34 \\
        \midrule
        \rowcolor{gray!10}
        \multicolumn{8}{l}{\emph{Models with fewer than 0.5B parameters}} \\
        gte-multilingual-reranker-base & 0.3B &
        60.69 & \underline{62.40} &
        \underline{71.90} & 56.08 &
        62.60 & \underline{59.01} \\
        jina-reranker-v2-base-multilingual & 0.3B &
        \underline{61.09} & 62.13 &
        68.46 & \underline{59.60} &
        \underline{63.29} & 57.18 \\
        Lychee-Reranker-Nano$^{\dagger}$ & 0.27B &
        \textbf{63.22} & \textbf{64.78} &
        \textbf{72.91} & \textbf{61.07} &
        \textbf{64.00} & \textbf{61.12} \\
        \bottomrule
    \end{tabular}%
    }
    \caption{Reranking performance on LMEB. 
    Detailed per-dataset results are reported in Table~\ref{tab:lmeb_reranking_dataset_results}. Minor discrepancies between the KaLM-Embedding-V2.5 scores reported here and those in Table~\ref{table:main_best_inst} arise from differences in hardware and experimental environments and are negligible. $^{\dagger}$For the Lychee-Reranker series, the reported parameter counts refer to active parameters (\textit{i.e.,} decoder parameters). }
    \label{tab:main_lmeb_rerank}
\end{table*}

%% file: tables/lmeb_mteb_correlation.tex
\begin{table}[t]
    \centering
    \small
    \renewcommand{\arraystretch}{0.95}
    \setlength{\tabcolsep}{4pt}
    \begin{tabular*}{\columnwidth}{@{\extracolsep{\fill}}lcccc@{}}
        \toprule
        \multirow{2}{*}{\textbf{LMEB}}
        & \multicolumn{2}{c}{\textbf{Pearson}}
        & \multicolumn{2}{c}{\textbf{Spearman}} \\
        \cmidrule(lr){2-3}
        \cmidrule(lr){4-5}
        & $r$ & $p$ & $\rho$ & $p$ \\
        \midrule
        Overall    & 0.267 & 0.1538 & 0.261 & 0.1631 \\
        Episodic   & 0.064 & 0.7353 & 0.129 & 0.4960 \\
        Dialogue   & 0.092 & 0.6283 & 0.140 & 0.4608 \\
        Semantic   & 0.187 & 0.3212 & 0.160 & 0.3985 \\
        Procedural & \textbf{0.537} & \textbf{0.0022}
                   & \textbf{0.602} & \textbf{0.0004} \\
        \bottomrule
    \end{tabular*}
    \caption{Pearson and Spearman correlations between the evaluation scores on LMEB and MTEB (eng, v2) (retrieval subset). Overall denotes the average score across all 22 LMEB datasets. All $p$-values are two-sided; correlations with $p<0.05$ are \textbf{boldfaced}.}
    \label{tab:lmeb_mteb_correlation}
\end{table}

%% file: chapters/related.tex
\section{Related Work}

\noindent\textbf{Embedding Benchmarks.}
Early embedding benchmarks, such as SentEval~\citep{DBLP:conf/lrec/ConneauK18} and SemEval~\citep{DBLP:conf/semeval/AgirreCDG12,DBLP:conf/starsem/AgirreCDGG13,DBLP:conf/semeval/AgirreBCCDGGMRW14,DBLP:conf/semeval/AgirreBCCDGGLMM15,DBLP:conf/semeval/AfzalWL16,agirre2016semeval}, focused on sentence representations and semantic textual similarity (STS).
BEIR~\citep{DBLP:journals/corr/abs-2104-08663} broadened retrieval evaluation across domains, while MIRACL~\citep{DBLP:journals/tacl/0018TOKAL0RL23} and AIR-Bench~\citep{DBLP:conf/acl/ChenWLWXXLLL25} cover multiple languages and retrieval settings.
MTEB, C-MTEB, and MMTEB~\citep{DBLP:conf/eacl/MuennighoffTMR23,DBLP:conf/sigir/XiaoLZMLN24,DBLP:conf/iclr/EnevoldsenCKKMS25} standardize evaluation across retrieval, classification, clustering, reranking, and STS, with public leaderboards enabling researchers to track progress in embedding model development.
MIEB~\citep{DBLP:journals/corr/abs-2504-10471} and MMEB~\citep{DBLP:conf/iclr/JiangMYYZC25} extend evaluation to image and multimodal embeddings.
Despite their broad coverage, long-horizon memory retrieval involving fragmented, context-dependent, and temporally distant memories remains underexplored, leaving model capabilities in memory-intensive scenarios unclear.
\smallsection{Embedding Models}
Text embedding models map text to vectors for similarity search~\citep{babakhin2026nemotron,DBLP:conf/naacl/ZhaoZSHLLHZ25,DBLP:journals/corr/abs-2507-20783,DBLP:conf/emnlp/ZhaoLZHCHZ24}.
Embedding research has progressed from static word embeddings such as GloVe~\citep{DBLP:conf/emnlp/PenningtonSM14} to contextual representations with BERT~\citep{DBLP:conf/naacl/DevlinCLT19} and sentence-pair fine-tuning with Sentence-BERT~\citep{DBLP:conf/emnlp/ReimersG19}.
Recent embedding model families, including GTE~\citep{DBLP:journals/corr/abs-2308-03281,DBLP:conf/emnlp/ZhangZLXDTLYXHZ24}, Qwen3-Embedding~\citep{DBLP:journals/corr/abs-2506-05176}, BGE~\citep{DBLP:journals/corr/abs-2402-03216}, Jina~\citep{DBLP:journals/corr/abs-2506-18902,akram2026jinaembeddingsv5texttasktargetedembeddingdistillation}, NV-Embed~\citep{DBLP:conf/iclr/Lee0XRSCP25}, and KaLM~\citep{DBLP:journals/corr/abs-2501-01028,DBLP:journals/corr/abs-2506-20923}, span encoder-based and LLM-based architectures.
Their training commonly uses multi-task contrastive learning~\citep{DBLP:journals/corr/abs-2402-17016} and instruction tuning~\citep{DBLP:conf/acl/AsaiSL0I0HY23,DBLP:conf/acl/SuSKWHOYSZ023}.
LLM-based models often generalize well but involve trade-offs in efficiency, latency, and deployment cost. Their gains in passage retrieval may not transfer to long-horizon memory scenarios.

%% file: chapters/conclusion.tex
\section{Conclusion}
\label{sec:conclu}
We introduce the Long-horizon Memory Embedding Benchmark (LMEB) to assess long-horizon memory retrieval capabilities that remain underexplored in existing benchmarks.
LMEB comprises 22 datasets and 193 zero-shot retrieval tasks across four memory types, with a focus on fragmented, context-dependent, and temporally distant memories.
Integrated with MTEB, LMEB provides an open-source framework and a public leaderboard for scalable, extensible, and reproducible evaluation, enabling new models and datasets to be added with minimal code.
We hope LMEB will help guide model selection, support future research, and track progress in long-horizon memory retrieval.

%% file: chapters/appendix.tex
\input{tables/datasets_links}

\section{Datasets}
\label{app:datasets}
LMEB comprises 22 evaluation datasets spanning four memory types.
Each dataset provides a corpus $\mathcal{C}$ and a set of test queries $\mathcal{Q}$.
The following descriptions summarize the source datasets and their adaptation to LMEB; 
task types are detailed in~\S\ref{app:tasks}.
Table~\ref{tab:dataset_links} lists the corresponding dataset links.

\input{tables/episodic_examples}

\subsection{Episodic Memory Retrieval}
Episodic memory retrieval recalls past events using temporal cues, entities, event content, and spatial context~\citep{DBLP:journals/corr/abs-2407-09450,DBLP:journals/corr/abs-2502-06975}.
In LMEB, episodic queries are used to retrieve relevant event memories.
Examples are presented in Table~\ref{tab:episodic_example}.

\smallsectionnodot{EPBench}~\citep{DBLP:conf/iclr/HuetB025} is a synthetic episodic memory benchmark for evaluating event recall and episodic reasoning in LLMs.
It represents events through structured fields, including temporal and spatial context, the entities involved, and detailed event descriptions.
We use its event set as the corpus $\mathcal{C}$ and the provided task queries as queries $\mathcal{Q}$.
\smallsectionnodot{KnowMeBench}~\citep{wu2026knowme} is a benchmark for long-horizon person understanding built from long-form autobiographical narratives.
It converts each narrative into a time-anchored, flashback-aware memory stream, with evidence-linked queries assessing factual recall, subjective state attribution, and principle-level reasoning.
We use the reconstructed narrative stream as the corpus $\mathcal{C}$ and the benchmark questions as queries $\mathcal{Q}$.
We exclude queries without annotated relevant memories while retaining \texttt{adversarial\_abstention} queries with counter-evidence annotated as relevant.

\subsection{Dialogue Memory Retrieval}
Dialogue memory retrieval maintains context across multi-turn interactions by recalling relevant conversation history and user preferences~\citep{DBLP:conf/iclr/WuWYZCY25,DBLP:conf/acl/MaharanaLTBBF24}.
Such recall supports coherent conversations and personalization over time~\citep{DBLP:journals/corr/abs-2402-16288,DBLP:journals/corr/abs-2503-07018}.
In LMEB, dialogue queries are used to retrieve relevant turns, rounds, or sessions.
Examples are presented in Table~\ref{tab:dialogue_example}.
\smallsectionnodot{LoCoMo}~\citep{DBLP:conf/acl/MaharanaLTBBF24} is a dataset of very long-term, persona-grounded dialogues with explicit temporal event structure.
It contains conversations of up to 32 sessions ($\sim$ 600 turns; $\sim$ 16K tokens on average) and evaluates long-range dialogue memory through tasks such as question answering.
We use the dialogue history as the corpus $\mathcal{C}$ and the benchmark questions as queries $\mathcal{Q}$.
\smallsectionnodot{LongMemEval}~\citep{DBLP:conf/iclr/WuWYZCY25} evaluates long-term memory in chat assistants through sustained, multi-session interactions.
It includes six question types covering user and assistant information, user preferences, multi-session reasoning, knowledge updates, and temporal reasoning.
We use the cleaned LongMemEval-M release (\texttt{longmemeval\_m\_cleaned.json}), treating the chat history as the corpus $\mathcal{C}$ and the curated questions as queries $\mathcal{Q}$.
\smallsectionnodot{REALTALK}~\citep{DBLP:journals/corr/abs-2502-13270} consists of a 21-day corpus of authentic dialogues, capturing long-term, open-domain interactions with richer emotional expressions and greater variation in persona consistency than synthetic data.
Its tasks include persona simulation and memory probing, in which models answer questions based on long-term dialogue history.
LMEB focuses on memory probing, using the dialogue history as the corpus $\mathcal{C}$ and the memory-probing questions as queries $\mathcal{Q}$.
\smallsectionnodot{TMD}~\citep{DBLP:journals/corr/abs-2406-00057} evaluates retrieval-augmented generation models on long-term conversational memory.
It addresses two challenges: time/event-based queries, which require retrieving information from past conversations using temporal cues or event order, and ambiguous queries, which require surrounding conversational context for interpretation.
Its corpus incorporates LoCoMo conversations generated by
GPT-4 and subsequently edited by humans.
Time-based queries are automatically instantiated from
manually written templates, while \emph{Time+Content} queries
are mainly generated by GPT-4 and manually revised.
In LMEB, we use the conversation histories as the corpus
$\mathcal{C}$ and these questions as queries $\mathcal{Q}$.
\smallsectionnodot{MemBench}~\citep{DBLP:conf/acl/Tan000DD25} evaluates the memory capabilities of LLM-based agents across multiple memory levels, including factual and reflective memory, and interactive scenarios involving participation and observation.
We use the dialogues as the corpus $\mathcal{C}$ and the associated queries as queries $\mathcal{Q}$.
\smallsectionnodot{ConvoMem}~\citep{DBLP:journals/corr/abs-2511-10523} is a conversational memory benchmark containing 75,336 question-answer pairs across categories such as user facts, assistant recall, preferences, and temporal changes.
In LMEB, we randomly select up to five questions from each source file using a fixed random seed of 55.
We retain sampled questions only if all annotated evidence texts can be matched to turns in the corresponding conversation histories, yielding 5,867 queries.
We use the conversation turns as the corpus $\mathcal{C}$ and the retained questions as queries $\mathcal{Q}$.
\subsection{Semantic Memory Retrieval}
Semantic memory retrieval focuses on general knowledge and concepts, usually with limited temporal dependency~\citep{tulving1972episodic}.
This knowledge is relatively stable and generalizable, supporting knowledge-augmented reasoning~\citep{DBLP:journals/corr/abs-2506-15841}.
In LMEB, questions or claims are used to retrieve relevant passages or sentences.
Examples are presented in Table~\ref{tab:semantic_example}.
For LooGLE and NovelQA, we segment documents using \texttt{semchunk}\footnote{\url{https://github.com/isaacus-dev/semchunk}} with a target chunk size of 256 tokens.
We check each evidence span and adjust passage boundaries as needed to keep it intact within a single passage.
Passages containing annotated evidence are labeled as relevant to the corresponding queries.
\smallsectionnodot{QASPER}~\citep{DBLP:conf/naacl/DasigiLBCSG21} is a question-answering (QA) dataset grounded in full research papers, with information-seeking queries that require reasoning across multiple document sections.
It contains 5,049 questions over 1,585 NLP papers.
Questions are written based on the title and abstract, while answers are grounded in evidence from the full text.
In LMEB, we use the test split and retain questions whose first answerable annotation contains evidence, yielding 1,335 queries.
Relevance judgments are constructed solely from the evidence in this selected annotation.
We use the paper content, segmented into passages, as the corpus $\mathcal{C}$ and the retained QASPER questions as queries $\mathcal{Q}$.
\smallsectionnodot{NovelQA}~\citep{DBLP:conf/iclr/WangNPWGDBH0WZ25} evaluates long-context comprehension of complex narratives from English novels.
It provides manually annotated questions targeting detailed and multi-hop understanding, with average context lengths exceeding 200K tokens.
We use the novel text, segmented into passages, as the corpus $\mathcal{C}$ and the benchmark questions as queries $\mathcal{Q}$.
\smallsectionnodot{PeerQA}~\citep{DBLP:conf/naacl/BaumgartnerBG25} is a document-level scientific QA dataset built from questions raised in peer reviews and answers provided by the original authors.
It contains 579 QA pairs over 208 academic papers, mostly from ML and NLP, and supports evidence retrieval.
In LMEB, we exclude questions whose corresponding paper text is unavailable and retain 136 questions.
We use the paper text, segmented into sentences, as the corpus $\mathcal{C}$ and the retained PeerQA questions as queries $\mathcal{Q}$.
\smallsectionnodot{Covid-QA}~\citep{moller-etal-2020-covid} is a biomedical QA dataset based on COVID-19 scientific articles in CORD-19.
It contains expert-written question-answer pairs, created by marking answer spans and formulating corresponding questions.
In LMEB, we use the processed version released by~\citet{DBLP:conf/emnlp/ContiFVBHC25}, treating the selected CORD-19 articles, segmented into passages, as the corpus $\mathcal{C}$ and the annotated questions as queries $\mathcal{Q}$.
In the version processed by ConTEB, GPT-4o annotates the passage in the source document that best provides the information needed to answer each question.
\smallsectionnodot{ESG-Reports}~\citep{DBLP:journals/corr/abs-2505-17166} consists of lengthy ESG reports from the fast-food industry.
It is derived from ViDoRe Benchmark v2~\citep{DBLP:journals/corr/abs-2505-17166}, which provides query--page relevance annotations for visual document retrieval.
In LMEB, we use the processed version released by \citet{DBLP:conf/emnlp/ContiFVBHC25}.
This version converts documents to text and segments them into passages, filters queries that rely solely on visual evidence such as tables or graphs, and manually re-annotates the remaining query--passage pairs.
We use the resulting passages as the corpus $\mathcal{C}$ and the retained queries as $\mathcal{Q}$.
\smallsectionnodot{MLDR}~\citep{DBLP:journals/corr/abs-2402-03216} is a multilingual long-document retrieval dataset constructed from Wikipedia, Wudao, and mC4 across 13 languages.
Questions are generated from sampled paragraphs and paired with their source documents.
In LMEB, we use only the English subset, sourced from Wikipedia, and adopt the processed version released by~\citet{DBLP:conf/emnlp/ContiFVBHC25}.
In this version, documents are segmented into passages, and GPT-4o identifies the passage within each source document that best provides the information needed to answer the corresponding query.
We use these passages as the corpus $\mathcal{C}$ and the generated questions as $\mathcal{Q}$.
\smallsectionnodot{LooGLE}~\citep{DBLP:conf/acl/LiWZZ24} is a long-context understanding benchmark built from recent documents spanning diverse domains, with an average document length of approximately 24K tokens.
It includes manually authored and cross-validated long-dependency QA pairs, alongside short-dependency QA pairs generated by GPT-3.5-turbo-16k and reviewed by human annotators.
We use the segmented documents as the corpus $\mathcal{C}$ and the corresponding questions as queries $\mathcal{Q}$.
\smallsectionnodot{SciFact}~\citep{DBLP:conf/emnlp/WaddenLLWZCH20} is a scientific claim verification dataset that requires retrieving evidence abstracts from the research literature to support or refute a claim.
It contains approximately 1.4K expert-written claims paired with annotated evidence abstracts, labels, and rationales.
In LMEB, we use the development split of 300 claims because the test split is unlabeled.
After excluding claims without annotated evidence, we retain 188 queries.
We use all 1,748 sentences from the 182 evidence abstracts as the corpus $\mathcal{C}$ and the retained claims as queries $\mathcal{Q}$.
For each claim, we map the sentence indices in all annotated rationales to corpus items and assign relevance $1$ to both supporting and refuting evidence.
\subsection{Procedural Memory Retrieval}
Procedural memory retrieval focuses on learned skills, action patterns, and structured procedures that guide task execution and multi-step reasoning~\citep{DBLP:journals/corr/abs-2508-06433,DBLP:journals/corr/abs-2509-25140}.
Such retrieval supports complex behaviors, tool use, and adaptive decision-making in reinforcement learning and tool-augmented systems~\citep{DBLP:journals/corr/abs-2507-21428,DBLP:journals/corr/abs-2507-02259}.
In LMEB, task requests or problem descriptions are used to retrieve relevant procedural memories, such as tool documentation, task experiences, trajectories, and items.
Examples are presented in Table~\ref{tab:procedural_example}.
\smallsectionnodot{Gorilla}~\citep{DBLP:conf/nips/PatilZ0G24} introduces APIBench, a benchmark covering APIs from HuggingFace, TorchHub, and TensorHub to evaluate tool use through API call generation.
Gorilla integrates a document retriever to access up-to-date API documentation and adapt to version changes at test time.
In LMEB, we use the processed version released by \citet{DBLP:conf/acl/ShiWYRWYR25}, treating API documentation as the corpus $\mathcal{C}$ and natural language tool-use instructions as queries $\mathcal{Q}$.
\smallsectionnodot{ToolBench}~\citep{DBLP:conf/iclr/QinLYZYLLCTQZHT24} is a large-scale tool-use benchmark constructed from 16,464 real-world RESTful APIs collected from RapidAPI Hub.
It provides automatically generated instructions for both single-tool and multi-tool scenarios, along with annotated solution paths.
In LMEB, we use the processed version released by~\citet{DBLP:conf/acl/ShiWYRWYR25} to evaluate retrieval in both scenarios, treating API documentation as the corpus $\mathcal{C}$ and tool-use instructions as queries $\mathcal{Q}$.
\smallsectionnodot{ReMe}~\citep{cao2025remember} extracts fine-grained procedural experiences through multi-faceted distillation, supports context-adaptive retrieval through scenario-aware indexing, and maintains a compact memory pool through utility-based refinement.
It emphasizes retrieval as a means of adapting past solution trajectories to new tasks.
In LMEB, we use the ready-to-use memories released for BFCL~\citep{DBLP:conf/icml/PatilMYJSSG25} and AppWorld~\citep{DBLP:conf/acl/TrivediKHMDLGSB24}, treating distilled experiences as the corpus $\mathcal{C}$ and task queries and generalized queries as queries $\mathcal{Q}$.
\smallsectionnodot{Proced\_mem\_bench}~\citep{DBLP:journals/corr/abs-2511-21730} isolates procedural memory retrieval from task execution, evaluating whether agents can retrieve functionally equivalent procedures under novel object instantiations.
Built on ALFWorld~\citep{DBLP:conf/iclr/ShridharYCBTH21}, it provides separate corpora of expert and LLM-generated trajectories, along with systematically stratified queries.
In LMEB, we use the LLM-generated trajectories as the corpus $\mathcal{C}$ and procedural task queries as queries $\mathcal{Q}$.
For LMEB, trajectories with a provided relevance score of at least 7.0 are treated as relevant memories.
\smallsectionnodot{MemGovern}~\citep{wang2026memgovern} introduces a procedural memory framework that transforms
large-scale, unstructured GitHub issue data into agent-usable experiential memory.
Through experience governance, it uses an LLM to convert historical issue--PR--patch records into structured experience cards and employs an agentic search strategy to retrieve relevant past solutions.
In LMEB, we use the fix experiences from these cards as the corpus $\mathcal{C}$ and the problem summaries as
queries $\mathcal{Q}$.
\smallsectionnodot{DeepPlanning}~\citep{zhang2026deepplanning} is a long-horizon planning benchmark that evaluates procedural reasoning under local and global constraints.
It includes multi-day travel and multi-product shopping tasks, requiring models to retrieve and integrate information to construct plans that satisfy time, budget, and resource constraints.
In LMEB, we adapt the shopping tasks to item-level retrieval, using shopping requests as queries $\mathcal{Q}$ and product records as the corpus $\mathcal{C}$.
We evaluate product rankings using nDCG@10 based on item-level relevance annotations.
The product records are synthetic, and the LLM-generated queries are reviewed and revised by human experts.
\section{Tasks}
\label{app:tasks}
LMEB comprises 193 zero-shot retrieval tasks from 22 English datasets across four memory types.
These tasks cover temporal constraints, information updates, evidence distributed across interactions or long documents, and past procedures applicable to new tasks.
Tables~\ref{tab:episodic_tasks}, \ref{tab:dialogue_tasks}, \ref{tab:semantic_tasks}, and \ref{tab:procedural_tasks} summarize the task types and example abilities assessed for each dataset, together with the numbers of subsets and tasks.
Here, \textbf{Task Type} refers to a specific category of retrieval tasks, indicating what the model is expected to retrieve.
The example abilities assessed describe the capabilities needed to identify relevant memories for these tasks.
A \textbf{Subset} denotes a dataset partition organized by topic, scenario, source, or difficulty, allowing the same task type to be evaluated in different contexts.
Each task is instantiated from a valid subset--task-type combination.
When all task types apply to every subset, the total number of tasks (\#Tasks) equals the number of subsets (\#Subsets) multiplied by the number of task types.
For example, EPBench contains nine subsets, each evaluated on the same six task types listed in Table~\ref{tab:episodic_tasks}.
Each subset--task-type combination constitutes a separate retrieval task, yielding $9 \times 6 = 54$ tasks.
When task types apply only to selected subsets, only valid combinations are counted.
For instance, ReMe contains six subsets: the three AppWorld subsets include only \texttt{generalized\_query}, whereas the three BFCL subsets include both \texttt{generalized\_query} and \texttt{task\_query}.
ReMe therefore contributes $3 \times 1 + 3 \times 2 = 9$ tasks, as detailed in Table~\ref{tab:procedural_tasks}.
\section{Instructions}
\label{app:instrcutions}
Task instructions specify the retrieval objective for each dataset or task type.
Let $q$ denote an evaluation query together with any contextual information added during benchmark construction.
When task instructions are used, we prepend the corresponding instruction to $q$.
The instructed query is formulated as follows:
\begin{equation}
    \begin{aligned}
        q_{\mathrm{inst}} & ={}  \texttt{Instruct:} ~~\texttt{\{task instruction\}} \\
                             & \quad \text{\texttt{\textbackslash n}}~\texttt{Query:}~q.
    \end{aligned}
\end{equation}
$q_{\mathrm{inst}}$ denotes the instructed query, and \texttt{\textbackslash n} represents a newline.
Without task instructions, the model receives only the evaluation query:
\begin{equation}
    q_{\mathrm{noinst}} = q.
\end{equation}
Contextual information already included in $q$, such as an explicit temporal reference, is retained in both settings.
For the main results, we use the instruction settings listed in Table~\ref{tab:model_links}.
The instruction analysis in \S\ref{sec:in_depth_analysis} compares selected models with and without task instructions.
Tables~\ref{tab:task_inst_1} and~\ref{tab:task_inst_2} list the task instructions for each dataset and task type in LMEB.

\section{Experimental Setup}
\label{app:setup}

\subsection{Benchmarked Embedding Models}
We evaluate 30 embedding models ranging from 33 million (M)
to 12 billion (B) parameters.
Table~\ref{tab:model_links} provides links to the public
model checkpoints and summarizes the task-instruction settings used in our evaluation.
The evaluated models are summarized by family below.

\smallsectionnodot{KaLM.}
KaLM-Embedding-Gemma3-12B-2511
(abbreviated as KaLM-Embedding-Gemma3),
KaLM-Embedding-V2.5, and KaLM-Embedding-V1~\citep{DBLP:journals/corr/abs-2506-20923,DBLP:journals/corr/abs-2501-01028}.

\smallsectionnodot{BGE.}
bge-multilingual-gemma2, bge-m3 (Dense),
bge-large-en-v1.5, bge-base-en-v1.5,
and bge-small-en-v1.5~\citep{DBLP:conf/sigir/XiaoLZMLN24,DBLP:journals/corr/abs-2402-03216}.
For bge-m3, we evaluate only its dense embeddings.

\smallsectionnodot{Qwen3-Embedding.}
Qwen3-Embedding-8B, Qwen3-Embedding-4B,
and Qwen3-Embedding-0.6B~\citep{DBLP:journals/corr/abs-2506-05176}.

\smallsectionnodot{F2LLM.}
F2LLM-v2-8B and F2LLM-v2-4B~\citep{f2llm-v2}.

\smallsectionnodot{SFR.}
SFR-Embedding-Mistral~\citep{SFRAIResearch2024}
and SFR-Embedding-2\_R~\citep{SFR-embedding-2}.

\smallsectionnodot{E5.}
e5-mistral-7b-instruct, multilingual-e5-large-instruct,
e5-large-v2, multilingual-e5-small,
and e5-small-v2~\citep{DBLP:journals/corr/abs-2402-05672,DBLP:journals/corr/abs-2212-03533}.

\smallsectionnodot{Jina.}
jina-v5-text-small and jina-v5-text-nano\footnote{We use
the retrieval-optimized checkpoints,
\texttt{jina-embeddings-v5-text-small-retrieval} and
\texttt{jina-embeddings-v5-text-nano-retrieval},
respectively.}~\citep{akram2026jinaembeddingsv5texttasktargetedembeddingdistillation}.

\smallsectionnodot{Snowflake Arctic.}
snowflake-arctic-embed-l-v2.0 and
snowflake-arctic-embed-m-v2.0~\citep{yu2024arctic}.

\smallsectionnodot{Other Models.}
NV-Embed-v2~\citep{DBLP:conf/iclr/Lee0XRSCP25},
Giga-Embeddings-instruct~\citep{kolodin2025gigaembeddings},
stella\_en\_400M\_v5~\citep{zhang2024jasper},
granite-embedding-311m-multilingual-r2
(abbreviated as granite-embedding-311m-ml-r2)~\citep{awasthy2026graniteembeddingmultilingualr2},
EmbeddingGemma-300M~\citep{vera2025embeddinggemmapowerfullightweighttext},
and nomic-embed-text-v1~\citep{nussbaum2024nomic}.

\subsection{Benchmarked Sparse Retrieval Models}
We further evaluate three sparse retrievers, covering BM25-based
lexical retrieval and learned sparse representations.

\smallsectionnodot{BM25S.}
We use bm25s~\citep{lu2024bm25s}\footnote{\url{https://github.com/xhluca/bm25s}}
as the BM25 lexical retrieval baseline.

\smallsectionnodot{SPLADE.}
We evaluate splade-v3-distilbert~\citep{lassance2024splade}\footnote{\url{https://huggingface.co/naver/splade-v3-distilbert}},
the DistilBERT-based checkpoint of the SPLADE-v3 family.

\smallsectionnodot{MILCO.}
We evaluate milco-650m~\citep{nguyen2026milco}\footnote{\url{https://huggingface.co/omai-research/milco-650m}},
a multilingual learned sparse retrieval model.
\subsection{Benchmarked Reranking Models}
We evaluate 13 reranking models ranging from 270M to 8B
parameters in a two-stage retrieval pipeline.
All models rerank the same set of up to 100 candidate
memories retrieved by KaLM-Embedding-V2.5 for each query.

\smallsectionnodot{Lychee-Reranker.}
Lychee-Reranker-Large, Lychee-Reranker-Small,
and Lychee-Reranker-Nano~\citep{zhao2026kalmrerankerv1fastlateinteraction}.
These names denote the Large, Small, and Nano R2 checkpoints
in the KaLM-Reranker-V1 family, respectively.
All three models are evaluated using Matryoshka embedding pooling
(MEP) with a compression ratio of $4\times$.
For these models, the reported parameter counts refer to
the active parameters in the decoder.

\smallsectionnodot{Qwen3 and GTE.}
Qwen3-Reranker-8B, Qwen3-Reranker-4B,
and Qwen3-Reranker-0.6B~\citep{DBLP:journals/corr/abs-2506-05176},
together with
gte-multilingual-reranker-base~\citep{DBLP:conf/emnlp/ZhangZLXDTLYXHZ24}.

\smallsectionnodot{BGE.}
bge-reranker-v2-gemma, bge-reranker-large,
and bge-reranker-v2-m3~\citep{DBLP:journals/corr/abs-2402-03216,li2024llama2vec,DBLP:conf/sigir/XiaoLZMLN24}.

\smallsectionnodot{Mxbai.}
mxbai-rerank-large-v2 and
mxbai-rerank-base-v2~\citep{li2025prorank}.

\smallsectionnodot{Jina.}
jina-reranker-v2-base-multilingual~\citep{jinaai2025rerankerv2base}.

Table~\ref{tab:reranker_links} provides links to the public reranking
model checkpoints and summarizes the task-instruction
settings used in our evaluation.
\subsection{Implementation Details}
We describe the evaluation metric, embedding and reranking protocols,
task-instruction settings, maximum input lengths, and computational resources below.

\noindent\textbf{Evaluation Metric.}
Following BEIR and MTEB~\citep{DBLP:journals/corr/abs-2104-08663,DBLP:conf/eacl/MuennighoffTMR23},
we evaluate retrieval performance using nDCG@10, which accounts
for both relevance and ranking positions.
Score aggregation follows the procedure described in
\S\ref{sec:evaluation_protocol}, with all scores reported on a 0--100 scale
and rounded only after aggregation.
\smallsection{Embedding Evaluation}
Models are evaluated without task-specific fine-tuning on LMEB.
Queries and memory items are encoded separately, and memories
are ranked by embedding similarity.
Retrieval is restricted to the specified candidate pool when
a memory scope constraint applies; otherwise, the full task
corpus is searched.
\smallsection{Reranking Evaluation}
KaLM-Embedding-V2.5 serves as the first-stage retriever.
For each query, it retrieves up to 100 memories from the
specified candidate pool, or from the full task corpus
when no scope constraint applies.
All 13 rerankers rescore and reorder the same candidates
and are evaluated using the same relevance judgments
as the first-stage retriever.

\smallsection{Task Instructions}
The main results use the task-instruction settings reported in
Tables~\ref{tab:model_links} and~\ref{tab:reranker_links}.
For the instruction analysis in \S\ref{sec:in_depth_analysis}, selected models are evaluated
under both \textbf{w/o inst.} and \textbf{w/ inst.} settings:
the former uses the query without a task instruction, whereas
the latter includes the corresponding task instruction.
Temporal and interaction context already included in the query
is retained in both settings.
Task instructions and query formats are provided
in~\S\ref{app:instrcutions}.

\input{tables/lmeb_sparse_ret}

\smallsection{Maximum Input Length}
Embedding models encode queries and memories separately,
with a default maximum input length of 1024 tokens for each.
Rerankers use a default maximum input length of 1024 tokens
for the query and document combined.
For models with shorter sequence limits, we use their maximum supported length.
Task instructions and formatting are included in these limits.
For embedding models using the 1024-token limit, we also
run the evaluation with a 512-token limit.
Their overall nDCG@10 scores differ by about 0.2 points
or less between the two settings on the 0--100 scale,
without affecting the main experimental conclusions.
\smallsection{Computational Resources}
Embedding and reranking experiments are conducted using four NVIDIA H100 GPUs.
\section{Dataset Licenses}
\label{app:license}
We summarize licensing information and source-content terms for the 22 datasets in LMEB based on their original releases and associated repositories.
\begin{itemize}[leftmargin=1em] 
    \item LongMemEval, MemBench, MLDR, and MemGovern: Provided under the MIT License.
    \item EPBench: The Figshare data release is distributed under CC0 1.0; the benchmark code is MIT-licensed.
    \item KnowMeBench, Covid-QA, Gorilla, ToolBench, ReMe, Proced\_mem\_bench, DeepPlanning, and NovelQA: Provided under Apache License 2.0.
    \item LoCoMo and ConvoMem: Provided under the CC BY-NC 4.0 license.
    \item QASPER: Provided under the CC BY 4.0 license.
    \item SciFact: Provided under the CC BY-NC 2.0 license.
    \item PeerQA: Provided under the CC BY-NC-SA 4.0 license.
    \item LooGLE: Provided under the CC BY-SA 4.0 license.
    \item ESG-Reports, REALTALK, TMD: Rights to the source material remain with the respective rights holders.
\end{itemize}
\section{Weighted Jaccard Similarity}
\label{app:jaccard}
We use weighted Jaccard similarity~\citep{DBLP:conf/icdm/Ioffe10} to quantify lexical overlap between dataset corpora based on their unigram word distributions.
For two corpora $S$ and $T$, let $\mathcal{V}$ denote the union of their vocabularies.
For each word $k \in \mathcal{V}$, $S_k$ and $T_k$ denote its relative frequencies in $S$ and $T$, respectively, calculated as its occurrence count divided by the total word count in the corresponding corpus.
If a word is absent from a corpus, its relative frequency in that corpus is set to zero.
The weighted Jaccard similarity is defined as follows:
\begin{equation}
    J(S,T)
    =
    \frac{\sum_{k \in \mathcal{V}} \min(S_k,T_k)}
         {\sum_{k \in \mathcal{V}} \max(S_k,T_k)} .
\end{equation}
The score ranges from $0$ to $1$, with higher values indicating greater lexical overlap.
A score of $0$ indicates no shared words, whereas a score of $1$ indicates identical normalized unigram distributions.
\section{Sparse Retrieval Results}
\label{app:sparse_results}
Table~\ref{tab:lmeb_sparse_results} presents the nDCG@10 results
of three sparse retrievers on LMEB, complementing the evaluation
of embedding models in \S\ref{sec:main_rets}.
The results reveal three main findings:

\smallsectionnodot{(1) Dense embeddings do not always outperform sparse retrieval.}
milco-650m achieves a Mean (Dataset) score of 56.03,
outperforming 18 of the 30 embedding models in
Table~\ref{table:main_best_inst}.
These include substantially larger models such as
Qwen3-Embedding-8B (55.94) and e5-mistral-7b-instruct (55.21).

\smallsectionnodot{(2) Sparse retrieval is competitive on semantic memory.}
Despite having only 650M parameters, milco-650m achieves a
semantic score of 60.36, only 1.82 points below NV-Embed-v2 (7B),
the strongest embedding model for this memory type
(Table~\ref{table:main_best_inst}).
However, its Mean (Dataset) score remains 5.38 points below that
of the best embedding model, indicating that the competitiveness
of sparse retrieval varies across memory types.

\smallsectionnodot{(3) Dialogue retrieval remains a common weakness.}
Among the four memory types, dialogue yields the lowest scores
for all three sparse retrievers, with the highest score reaching only 49.04.
This is consistent with the pattern observed for embedding models in
\S\ref{sec:main_rets}, indicating that dialogue retrieval remains
challenging for both sparse and dense retrieval models.
These results show that LMEB can reveal model strengths across memory types and guide model selection and optimization.
\input{tables/grep}
\section{Agentic GREP Retrieval}
\label{app:agentic_grep}
We evaluate agentic GREP on the LoCoMo single-hop and multi-hop subsets, containing 840 and 280 queries, respectively.
We use DeepSeek-V4-Flash~\citep{xu2026deepseek} with a lightweight terminal agent adapted from Pi\footnote{\url{https://github.com/earendil-works/pi}}.
The agent generates and executes GREP commands and returns a ranked list of memories, with up to 12 turns for single-hop queries and 16 turns for multi-hop queries.
For each query, search uses the same candidate pool, and results are evaluated using the same relevance labels.
\smallsectionnodot{(1) Agentic GREP leads with a much larger model.}
As shown in Table~\ref{tab:agentic_grep}, agentic GREP achieves nDCG@10 scores of 75.94 and 58.52 on single-hop and multi-hop queries, respectively, outperforming KaLM-Embedding-V2.5 with Lychee-Reranker-Nano by 4.99 and 7.74 points.
However, it uses DeepSeek-V4-Flash with 284B total parameters and 13B active parameters, compared with the 0.5B retriever and the reranker with 0.27B active parameters.
These results demonstrate the effectiveness of agentic GREP, with gains obtained using a substantially larger model.
\smallsectionnodot{(2) Agentic search requires substantial LLM inference.}
Agentic GREP uses an average of 80.3K tokens per single-hop query and 208.6K tokens per multi-hop query, costing approximately CNY~0.0198 and CNY~0.041 per query, respectively.
The total cost is CNY~16.62 for the 840 single-hop queries and CNY~11.48 for the 280 multi-hop queries, totaling CNY~28.10 across all 1,120 queries.
Agentic GREP uses iterative LLM reasoning and search to identify relevant memories, operating under a different
retrieval setting from embedding retrieval with or without reranking.
These results highlight the potential of agentic search for memory retrieval and the importance of considering inference cost in practical deployment.

\input{tables/dialogue_examples}
\input{tables/semantic_examples}
\input{tables/procedural_examples}

\input{tables/model_links}
\input{tables/episodic_tasks}
\input{tables/dialogue_tasks}
\clearpage
\begin{table*}[p]
    \centering
    \input{tables/semantic_tasks}
    \vspace{4.5em}
    \input{tables/procedural_tasks}
\end{table*}

\input{tables/instructions}

\input{tables/reranking_model_links}

\input{tables/detailed_embedding_results}

\input{tables/detailed_reranking_results}

\begin{figure*}[t]
    \centering
    \subfigure[LMEB (Episodic).]{%
        \includegraphics[width=0.485\linewidth]{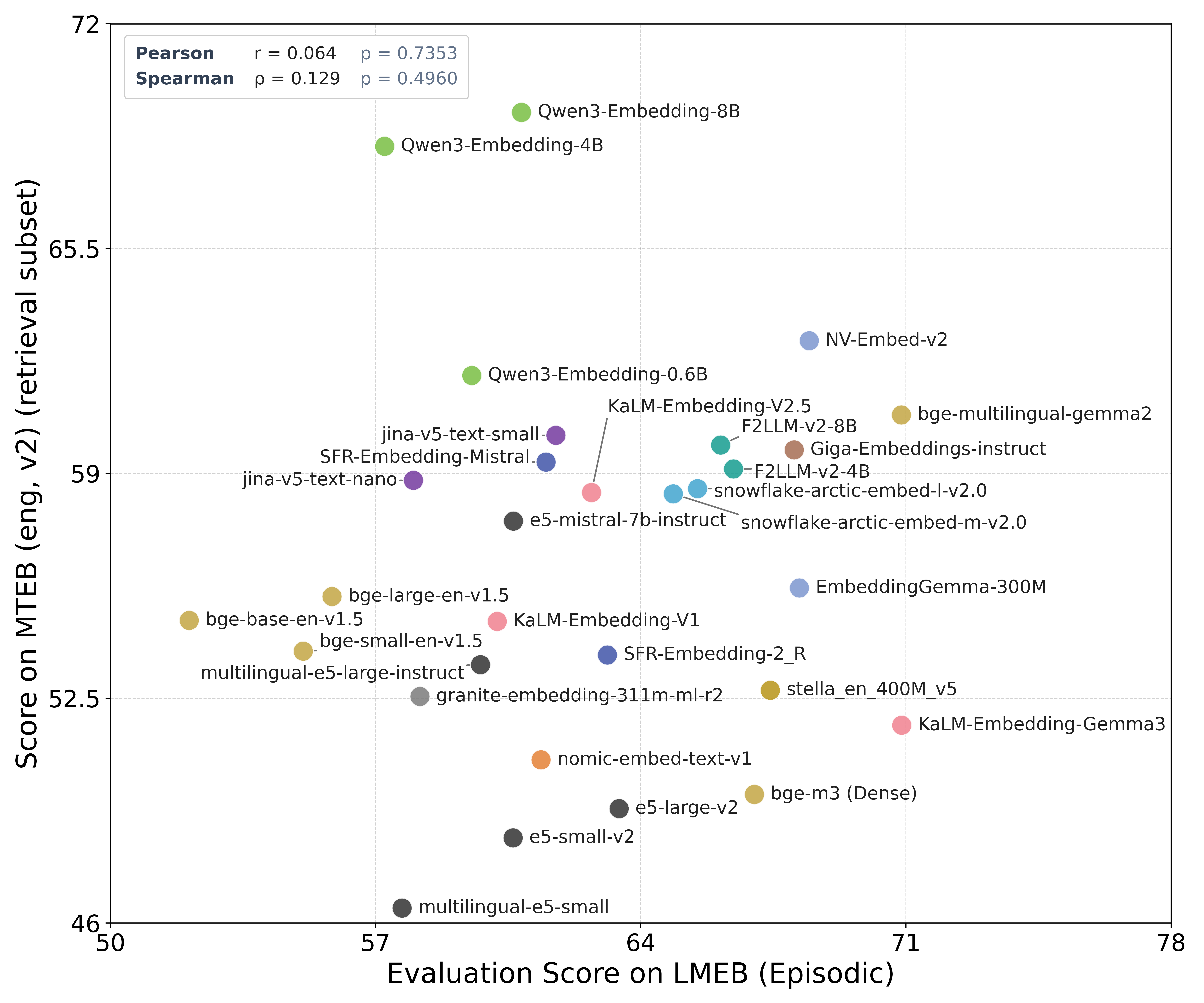}
        \label{fig:corr_episodic}}
    \hfill
    \subfigure[LMEB (Dialogue).]{%
        \includegraphics[width=0.485\linewidth]{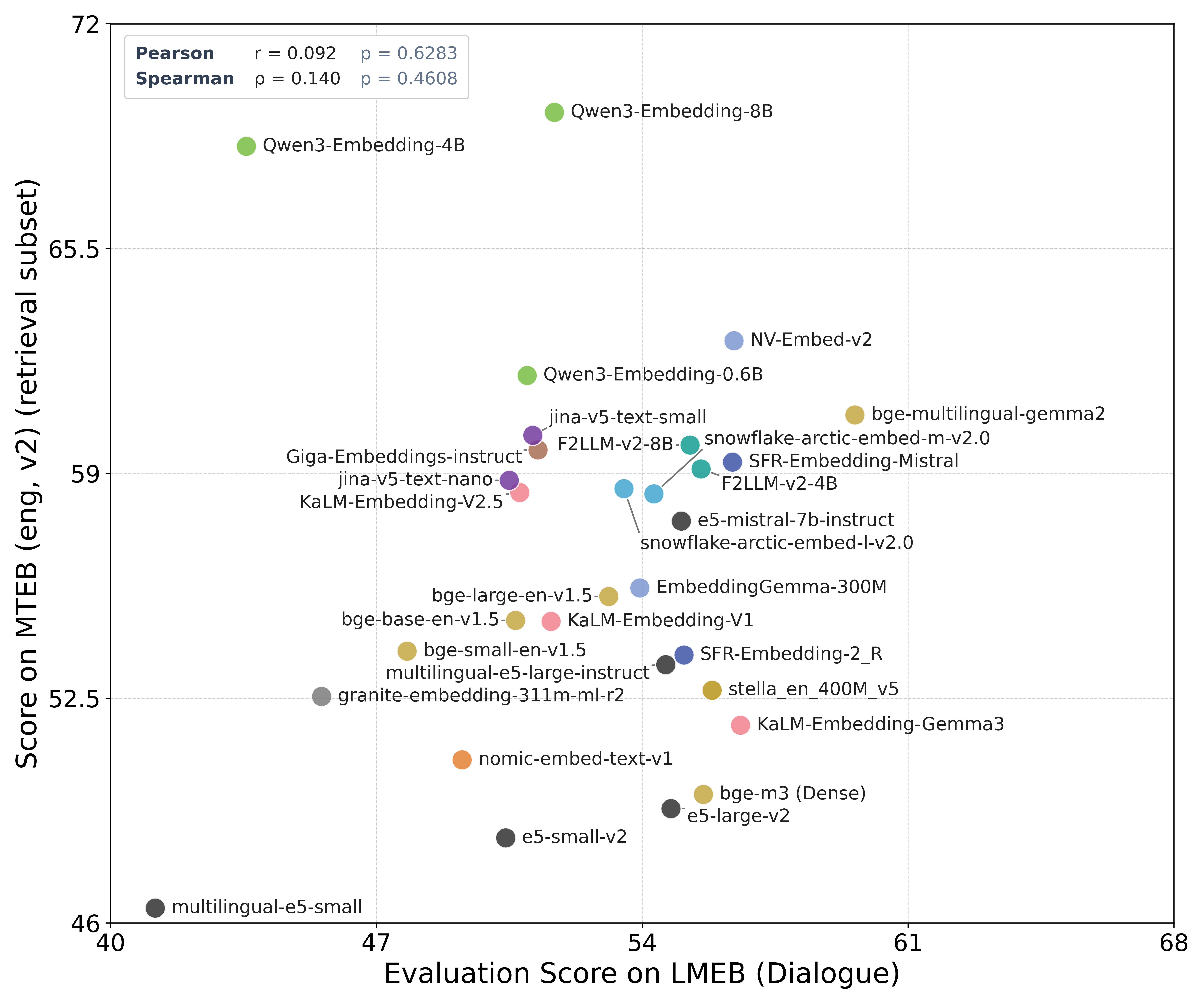}
        \label{fig:corr_dialogue}}
    \medskip
    \subfigure[LMEB (Semantic).]{%
        \includegraphics[width=0.485\linewidth]{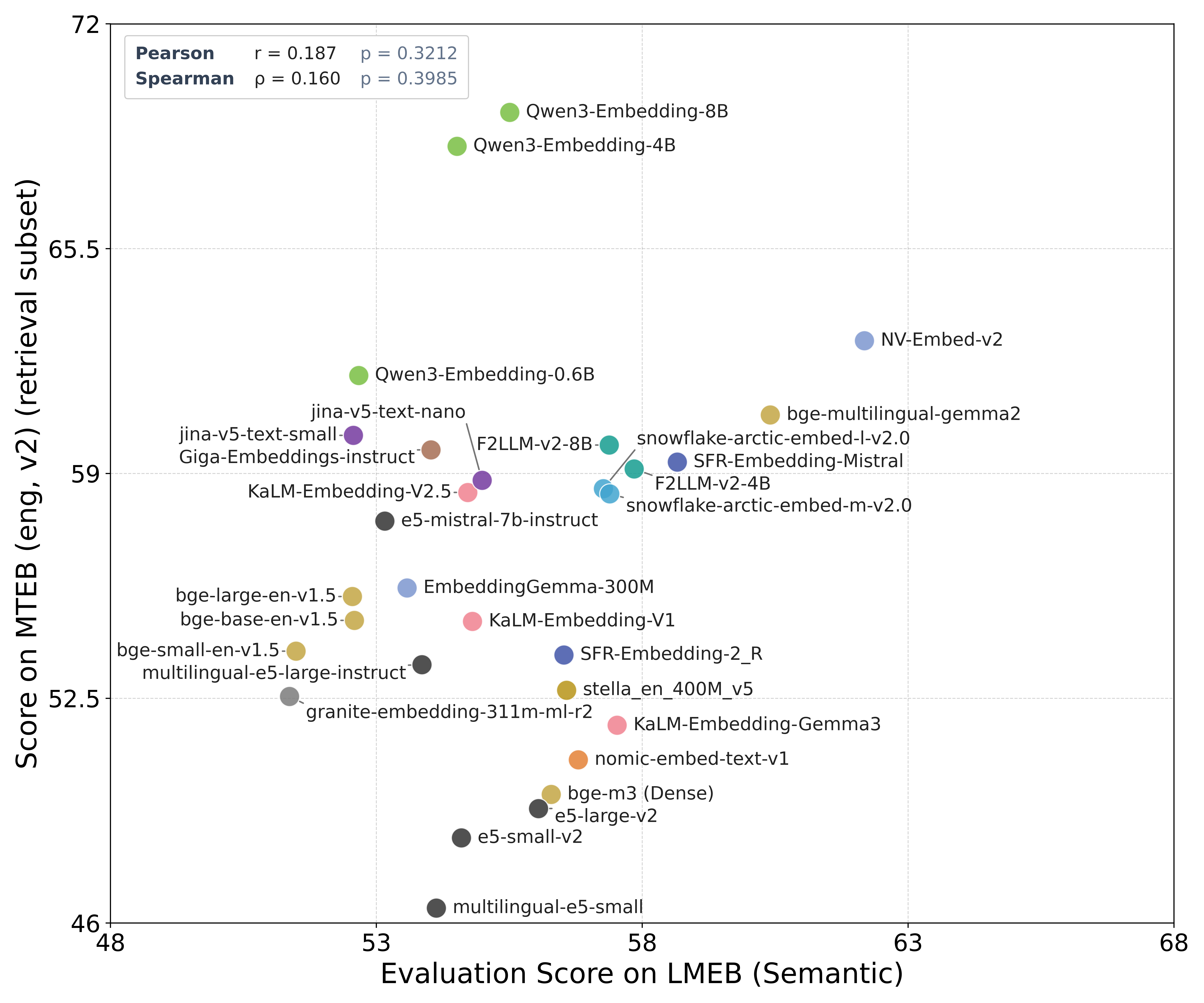}
        \label{fig:corr_semantic}}
    \hfill
    \subfigure[LMEB (Procedural).]{%
        \includegraphics[width=0.485\linewidth]{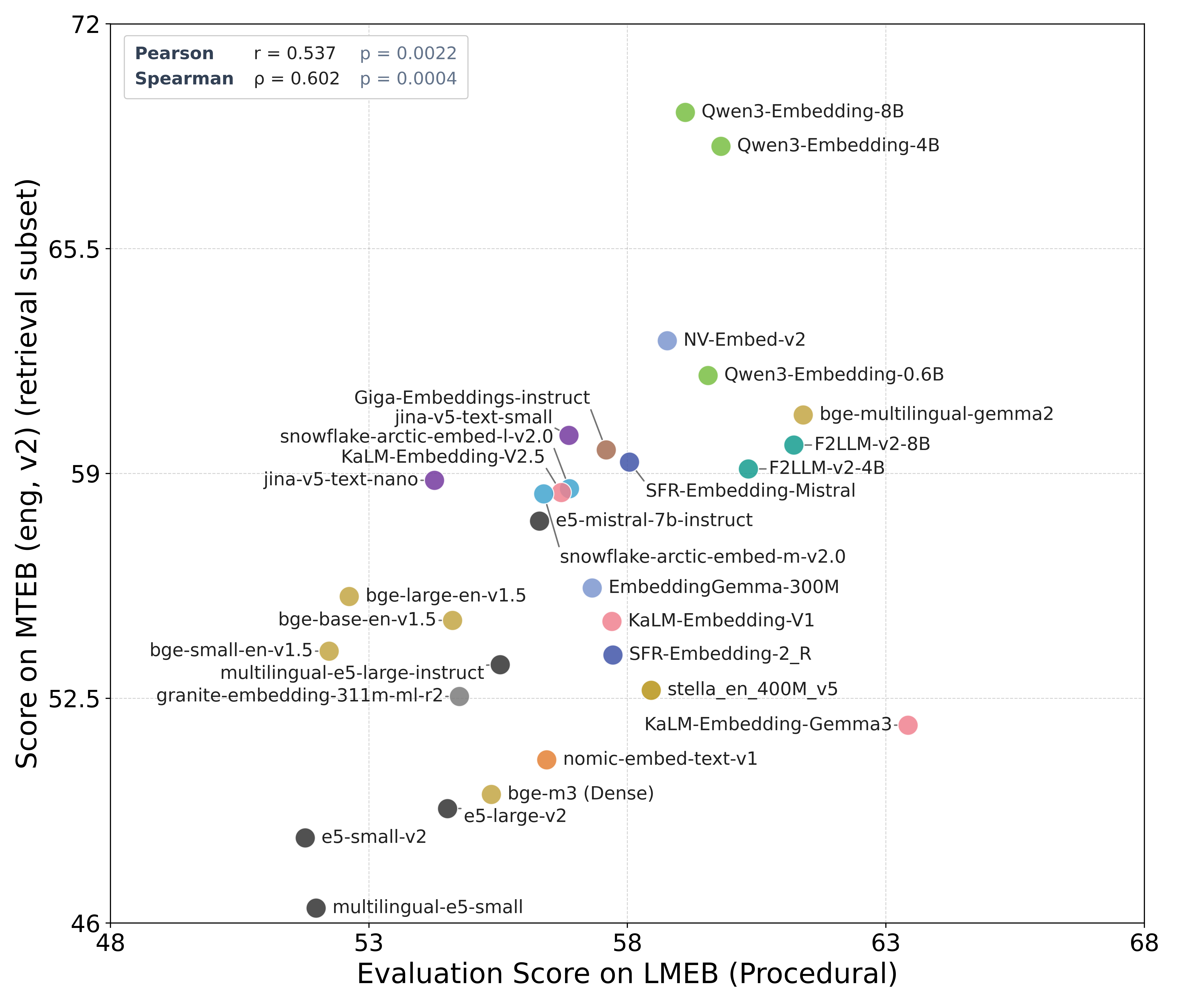}
        \label{fig:corr_procedural}}
    \caption{Pearson and Spearman correlations between the evaluation scores on each LMEB memory type and MTEB (eng, v2) (retrieval subset)~\cite{DBLP:conf/iclr/EnevoldsenCKKMS25}. Both use nDCG@10 as the evaluation metric. Marker colors indicate model families.}
    \label{fig:corr_by_memory_type}
\end{figure*}

%% file: tables/datasets_links.tex
\begin{table*}[ht]
    \centering
    \renewcommand\arraystretch{1.3}
    \resizebox{\textwidth}{!}{\begin{tabular}{ l | l }
        \toprule
        \multicolumn{1}{l|}{\textbf{Dataset}} &
        \multicolumn{1}{|c}{\textbf{Website (Link)}} \\
        \midrule
        \multicolumn{2}{c}{\textbf{Episodic Memory}} \\ 
        \hline
        EPBench~\citep{DBLP:conf/iclr/HuetB025} & \url{https://doi.org/10.6084/m9.figshare.28244480} \\
        KnowMeBench~\citep{wu2026knowme} & \url{https://github.com/QuantaAlpha/KnowMeBench/tree/main/KnowmeBench} \\
         \hline
        \multicolumn{2}{c}{\textbf{Dialogue Memory}} \\ 
        \hline
        LoCoMo~\citep{DBLP:conf/acl/MaharanaLTBBF24} & \url{https://github.com/snap-research/locomo/tree/main/data} \\
        LongMemEval~\citep{DBLP:conf/iclr/WuWYZCY25} & \url{https://huggingface.co/datasets/xiaowu0162/longmemeval-cleaned}\\
        REALTALK~\citep{DBLP:journals/corr/abs-2502-13270} & \url{https://github.com/danny911kr/REALTALK/tree/main/data} \\
        TMD~\citep{DBLP:journals/corr/abs-2406-00057}  & \url{https://github.com/Zyphra/TemporalMemoryDataset} \\
        MemBench~\citep{DBLP:conf/acl/Tan000DD25} & \url{https://github.com/import-myself/Membench/tree/main/MemData} \\
        ConvoMem~\citep{DBLP:journals/corr/abs-2511-10523} & \url{https://huggingface.co/datasets/Salesforce/ConvoMem} \\
         \hline
        \multicolumn{2}{c}{\textbf{Semantic Memory}} \\ 
        \hline
       QASPER~\citep{DBLP:conf/naacl/DasigiLBCSG21} & \url{https://huggingface.co/datasets/allenai/qasper} \\
       NovelQA~\citep{DBLP:conf/iclr/WangNPWGDBH0WZ25} & \url{https://huggingface.co/datasets/NovelQA/NovelQA} \\
       PeerQA~\citep{DBLP:conf/naacl/BaumgartnerBG25} & \url{https://huggingface.co/datasets/UKPLab/PeerQA} \\
       Covid-QA~\citep{moller-etal-2020-covid} & \url{https://huggingface.co/datasets/illuin-conteb/covid-qa} \\
       ESG-Reports~\citep{DBLP:journals/corr/abs-2505-17166} & \url{https://huggingface.co/datasets/illuin-conteb/esg-reports} \\
       MLDR~\citep{DBLP:journals/corr/abs-2402-03216} & \url{https://huggingface.co/datasets/illuin-conteb/mldr-conteb-eval} \\
       LooGLE~\citep{DBLP:conf/acl/LiWZZ24} & \url{https://huggingface.co/datasets/bigai-nlco/LooGLE} \\
       SciFact~\citep{DBLP:conf/emnlp/WaddenLLWZCH20} & \url{https://huggingface.co/datasets/allenai/scifact} \\
         \hline
        \multicolumn{2}{c}{\textbf{Procedural Memory}} \\ 
        \hline
       Gorilla~\citep{DBLP:conf/nips/PatilZ0G24} & \makecell[l]{\url{https://huggingface.co/datasets/mangopy/ToolRet-Queries} \\ \url{https://huggingface.co/datasets/mangopy/ToolRet-Tools}} \\
       ToolBench~\citep{DBLP:conf/iclr/QinLYZYLLCTQZHT24} & \makecell[l]{\url{https://huggingface.co/datasets/mangopy/ToolRet-Queries} \\ \url{https://huggingface.co/datasets/mangopy/ToolRet-Tools}} \\
       ReMe~\citep{cao2025remember} & \url{https://github.com/agentscope-ai/ReMe/tree/main/docs/library/paper_data/task} \\
       Proced\_mem\_bench~\citeyear{DBLP:journals/corr/abs-2511-21730} & \url{https://github.com/qpiai/Proced_mem_bench/tree/main/procedural_memory_benchmark} \\
       MemGovern~\citep{wang2026memgovern} & \url{https://github.com/QuantaAlpha/MemGovern/tree/main/data} \\ 
       DeepPlanning~\citep{zhang2026deepplanning}  & \url{https://huggingface.co/datasets/Qwen/DeepPlanning} \\
        \bottomrule
    \end{tabular}}
    \caption{Links to the dataset sources used in \lmeb, including original releases and processed versions.}
    \label{tab:dataset_links}
\end{table*}

%% file: tables/episodic_examples.tex
\begin{table*}[t]
    \centering
    \renewcommand\arraystretch{1.25}
    \resizebox{\textwidth}{!}{
    \begin{tabular}{ l | l | l | c }
        \toprule
        \multicolumn{1}{l|}{\textbf{Dataset}} &
        \multicolumn{1}{|c}{\textbf{Query}} &
        \multicolumn{1}{|c}{\textbf{Relevant-Document}} &
        \multicolumn{1}{|c}{\textbf{Granularity}} \\
        \midrule
        \midrule
         EPBench & \multicolumn{1}{p{6.5cm}|}{Think about Aurora Chavez's experiences. Describe all the key events they've been involved in, focusing on what happened rather than when or where it occurred.} & \multicolumn{1}{p{10.cm}|}{\dots\dots Aurora implemented blockchain solutions with a determination that bordered on obsession.  \dots\dots It wasn't until Samara Bayes tapped her on the shoulder that Aurora Chavez realized how much time had passed.  She blinked, momentarily disoriented as she emerged from her coding trance.  Samara Bayes gestured towards the refreshment table, a knowing smile on their face. \dots\dots} &  Event \\
         \hline
         KnowMeBench& \multicolumn{1}{p{6.5cm}|}{On August 15, 1969, which shipping company's bus did the narrator's family take?} &  \multicolumn{1}{p{10.cm}|}{\input{tables/knowmebench_ex}} & Event \\
        \bottomrule
    \end{tabular}
    }
    \caption{Examples of query-relevant document pairs from episodic memory retrieval datasets.}
    \label{tab:episodic_example}
\end{table*}

%% file: tables/knowmebench_ex.tex
\vspace{-0.8\baselineskip}
\begin{Verbatim}[breaklines=true]
{
    "id": 1,
    "timestamp": "1969-08-15 14:00:30",
    "location": "Rocky path",
    "action": "I watched the vehicle cross a bridge, driving down the narrow fjord with its right turn signal blinking before finally stopping.",
    "dialogue": null,
    "environment": "The bus belonged to Aucity Shipping Company, painted in two shades of brown.",
    "background": null,
    "inner_thought": null
}
\end{Verbatim}

%% file: tables/lmeb_sparse_ret.tex
\begin{table*}[!t]
    \centering
    \small
    \renewcommand{\arraystretch}{1.1}
    \setlength{\tabcolsep}{6pt}
    \resizebox{\textwidth}{!}{%
    \begin{tabular}{@{}lc|cccc|cc@{}}
        \toprule
        \textbf{Model} &
        \textbf{Size} &
        \textbf{Episodic} &
        \textbf{Dialogue} &
        \textbf{Semantic} &
        \textbf{Procedural} &
        \textbf{Mean (Dataset)} &
        \textbf{Mean (Type)} \\
        \midrule
        \multicolumn{2}{@{}l|}{Num. Datasets ($\rightarrow$)} &
        2 & 6 & 8 & 6 & 22 & 22 \\
        \midrule
        bm25s & -- &
        \textbf{66.11} & 43.91 & 50.59 & 51.28 & 50.37 & 52.97 \\
        splade-v3-distilbert & 67M &
        64.23 & 47.70 & 54.35 & \textbf{55.40} & 53.72 & 55.42 \\
        milco-650m & 650M &
        65.83 & \textbf{49.04} & \textbf{60.36} & 53.98 &
        \textbf{56.03} & \textbf{57.30} \\
        \bottomrule
    \end{tabular}%
    }
    \caption{Performance of sparse retrieval models on LMEB, measured by nDCG@10. Mean (Dataset) and Mean (Type) denote average scores across all 22 datasets and the four memory types, respectively. Best results are \textbf{boldfaced}.}
    \label{tab:lmeb_sparse_results}
\end{table*}

%% file: tables/grep.tex
\begin{table}[t]
    \centering
    \small
    \renewcommand{\arraystretch}{1.2}
    \setlength{\tabcolsep}{3pt}
    \begin{tabular*}{\columnwidth}{
        @{}l@{\extracolsep{\fill}}rr@{}
    }
        \toprule
        \textbf{Method} &
        \textbf{Single-hop} &
        \textbf{Multi-hop} \\
        \midrule\midrule
        KaLM-Embedding-V2.5
        & 50.92 & 36.31 \\
        \hspace{0.6em}+ Lychee-Reranker-Nano
        & 70.95 & 50.78 \\
        \midrule
        Agentic GREP
        & \textbf{75.94} & \textbf{58.52} \\
        \bottomrule
    \end{tabular*}
    \caption{Retrieval performance on the LoCoMo single-hop and multi-hop
    subsets, measured by nDCG@10. The second row adds reranking to
    KaLM-Embedding-V2.5, while agentic GREP uses
    DeepSeek-V4-Flash.}
    \label{tab:agentic_grep}
\end{table}

%% file: tables/dialogue_examples.tex
\begin{table*}[tp]
    \centering
    \renewcommand\arraystretch{1.18}
    \fvset{fontsize=\small,baselinestretch=1.1}
    \begin{adjustbox}{width=\textwidth,max totalheight=0.96\textheight,center}
    \begin{tabular}{ l | l | l | c }
        \toprule
        \multicolumn{1}{l|}{\textbf{Dataset}} &
        \multicolumn{1}{|c}{\textbf{Query}} &
        \multicolumn{1}{|c}{\textbf{Relevant-Document}} &
        \multicolumn{1}{|c}{\textbf{Granularity}} \\
        \midrule
        \midrule
         LoCoMo & \multicolumn{1}{p{4.5cm}|}{When is Melanie planning on going camping?} & \multicolumn{1}{p{13.5cm}|}{\input{tables/locomo_ex}} &  Turn \\
         \hline
         LongMemEval & \multicolumn{1}{p{4.5cm}|}{How many weeks ago did I meet up with my aunt and receive the crystal chandelier?} & \multicolumn{1}{p{13.5cm}|}{\input{tables/longmeneval_ex}} &  Session \\
         \hline
         REALTALK & \multicolumn{1}{p{4.5cm}|}{When is Elise's first yoga class?} & \multicolumn{1}{p{13.5cm}|}{\input{tables/realtalk_ex}} &  Turn \\
         \hline
         TMD & \multicolumn{1}{p{4.5cm}|}{What did we chat about on May 8th?} & \multicolumn{1}{p{13.5cm}|}{\input{tables/tmd_ex}} &  Turn \\
         \hline
         MemBench & \multicolumn{1}{p{4.5cm}|}{Who is older, Nolan Hayes or Sophie Turner?} & \multicolumn{1}{p{13.5cm}|}{\input{tables/membench_ex}} &  Round \\
         \hline
         ConvoMem & \multicolumn{1}{p{4.5cm}|}{Who is my favorite sports team?} & \multicolumn{1}{p{13.5cm}|}{\input{tables/convomem_ex}} &  Turn \\
        \bottomrule
    \end{tabular}
    \end{adjustbox}
    \caption{Examples of query-relevant document pairs from dialogue memory retrieval datasets.}
    \label{tab:dialogue_example}
\end{table*}

%% file: tables/locomo_ex.tex
\vspace{-1.5\baselineskip}
\begin{Verbatim}[breaklines=true]
{
  "speaker": "Melanie",
  "dia_id": "D2:7",
  "date_time": "1:14 pm on 25 May, 2023",
  "text": "... We're thinking about going camping next month. ..."
}
\end{Verbatim}

%% file: tables/longmeneval_ex.tex
\vspace{-1.2\baselineskip}
\begin{Verbatim}[breaklines=true]
[
    ......
    {
        "role": "user",
        "Data time": "05:25 AM on Saturday 04 March, 2023",
        "content": "... By the way, I also got a stunning crystal chandelier from my aunt today, which used to belong to my great-grandmother ..."
    }
    ......
]
\end{Verbatim}

%% file: tables/realtalk_ex.tex
\vspace{-0.8\baselineskip}
\begin{Verbatim}[breaklines=true]
{
    "clean_text": "Hey Kate! Yes, I did end up signing up for a Yoga class. My first class is today at 5:00pm! I’m super excited. I went and got myself a new yoga set and a yoga mat to be prepared. Any other tips for my first day?",
    "speaker": "elise",
    "date_time": "17.01.2024, 19:57:41",
    "dia_id": "D13:2"
}
{
    "clean_text": "I just had my first yoga class and it went amazing!",
    "speaker": "elise",
    "date_time": "18.01.2024, 01:18:57",
    "dia_id": "D13:4"
}
\end{Verbatim}

%% file: tables/tmd_ex.tex
\vspace{-0.8\baselineskip}\begin{Verbatim}[breaklines=true]
{
    "speaker": "Caroline",
    "dia_id": "D1:1",
    "text": "Hey Mel! Good to see you! How have you been?",
    "date_time": "01:56:04 AM on Monday 08 May, 2023",
    "response_number": "0"
}
......
{
    "speaker": "Melanie",
    "dia_id": "D1:18",
    "text": "Yep, Caroline. Taking care of ourselves is vital. I'm off to go swimming with the kids. Talk to you soon!",
    "date_time": "01:58:09 AM on Monday 08 May, 2023",
    "response_number": "17"
}
\end{Verbatim}

%% file: tables/membench_ex.tex
\vspace{-0.8\baselineskip}
\begin{Verbatim}[breaklines=true]
{
    "sid": 49,
    "user_message": "My boss, Nolan Hayes, is 39 years old.",
    "assistant_message": "Noted, Nolan Hayes is 39 years old.",
    "time": "'2024-10-01 08:49' Tuesday",
    "place": "Boston, MA",
    "rel": "boss",
    "attr": "age",
    "value": "39 years old"
}
{
    "sid": 76,
    "user_message": "Sophie Turner, my aunt, is 27 years old.",
    "assistant_message": "That's great to hear! It's always nice to have family connections, especially with someone as talented as Sophie.",
    "time": "'2024-10-01 09:16' Tuesday",
    "place": "Boston, MA",
    "rel": "Aunt",
    "attr": "age",
    "value": "27 years old"
}
\end{Verbatim}

%% file: tables/convomem_ex.tex
\vspace{-0.8\baselineskip}
\begin{Verbatim}[breaklines=true]
{
  "speaker": "User",
  "text": "I'm trying to schedule a trip back to Portland to see my family soon.   I should check the Trail Blazers schedule; taking a client to a game would be a great touch.   They've been my favorite team since I was a kid."
}
{
  "speaker": "User",
  "text": "I know I used to be all about basketball, but after living in Seattle for so long and getting season tickets, I'm completely hooked on football.  The Seattle Seahawks are hands down my favorite team now, the atmosphere at the games is just incredible."
}
\end{Verbatim}

%% file: tables/semantic_examples.tex
\begin{table*}[tp]
    \centering
    \small
    \renewcommand\arraystretch{1.2}
    \begin{adjustbox}{width=\textwidth,max totalheight=0.95\textheight,center}
    \linespread{1.15}\selectfont
    \begin{tabular}{ l | l | l | c }
        \toprule
        \multicolumn{1}{l|}{\textbf{Dataset}} &
        \multicolumn{1}{|c}{\textbf{Query}} &
        \multicolumn{1}{|c}{\textbf{Relevant-Document}} &
        \multicolumn{1}{|c}{\textbf{Granularity}} \\
        \midrule
        \midrule
        QASPER & \multicolumn{1}{p{5.cm}|}{How big is the ANTISCAM dataset?} &  \multicolumn{1}{p{11.5cm}|}{\texttt{<Title>}Datasets ::: AntiScam Dataset\texttt{</Title>} \texttt{<Text>} To enrich available non-collaborative task datasets, we created a corpus of human-human anti-scam dialogs in order to learn human elicitation strategies.  We chose a popular Amazon customer service scam scenario to collect dialogs between users and attackers who aim to collect users information.  We posted a role-playing task on the Amazon Mechanical Turk platform and collected a typing conversation dataset named AntiScam.  We collected 220 human-human dialogs.  The average conversation length is 12.45 turns and the average utterance length is 11.13 words.  Only 172 out of 220 users successfully identified their partner as an attacker, suggesting that the attackers are well trained and not too easily identifiable.  We recruited two expert annotators who have linguistic training to annotate 3,044 sentences in 100 dialogs, achieving a 0.874 averaged weighted kappa value.\texttt{</Text>}}  &  Paragraph \\
        \hline
        NovelQA & \multicolumn{1}{p{5.cm}|}{Down which river do Huck and Jim travel?} &  \multicolumn{1}{p{11.5cm}|}{\texttt{<Text>} ‘All right — that’s mighty good; they won’t find me, and they’ll think I’ve been killed, and floated down the river — there’s something up there that ‘ll help them think so — so don’t you lose no time, Jim, but just shove off for the big water as fast as ever you can.’ I never felt easy till the raft was two mile below there and out in the middle of the Mississippi. Then we hung up our signal lantern, and judged that we was free and safe once more. I hadn’t had a bite to eat since yesterday, so Jim he got out some corn-dodgers and buttermilk, and pork and cabbage and greens — there ain’t nothing in the world so good when it’s cooked right — and whilst I eat my supper we talked and had a good time. I was powerful glad to get away from the feuds, and so was Jim to get away from the swamp. We said there warn’t no home like raft, after all. Other places do seem so cramped up and smothery, but a raft don’t. You feel mighty free and easy and comfortable on a raft. \texttt{</Text>}} &  Paragraph \\
        \hline
        PeerQA & \multicolumn{1}{p{5.cm}|}{How many words are in the corpus in total?} &  \multicolumn{1}{p{11.5cm}|}{\texttt{<Title>} Descriptive Statistics \texttt{</Title>} \texttt{<Text>} The corpus includes 180 essays: 120 (Draft1 and Draft2) with an average of about 400 words and 60 (Draft3) with an average of around 500 words. \texttt{</Text>}} &  Sentence \\
        \hline
        Covid-QA & \multicolumn{1}{p{5.cm}|}{What does the NTCP protein mediate?} &  \multicolumn{1}{p{11.5cm}|}{\texttt{<Text>} NTCP is a transmembrane protein, usually located in the lateral surface (canalicular) of hepatocytes, which mediates bile acid transport [7].  Geographically, it is a good candidate for an HBV receptor.  Moreover, the authors could convert the cell lines previously non-permissible to HBV infection to permissible by over-expression of NTCP, again supporting its possible role in the HBV infection process.  This can be a critical and long-awaited discovery toward understanding HBV receptors and establishing an experimental HBV infection system. \texttt{</Text>}} &  Paragraph \\
        \hline
        
        ESG-Reports & \multicolumn{1}{p{5.cm}|}{What are McDonald's Franchisees?} &  \multicolumn{1}{p{11.5cm}|}{\texttt{<Text>} ¢ Franchisees: Collective group of independent individuals and entities owning and operating McDonald’s restaurants under one of the following structures — conventional franchise, developmental license or affiliate.  For more information on the Company’s structure and scope see our latest Annual Report. ¢ McDonald’s restaurants/Restaurants: Includes restaurants owned by the Company and its Franchisees.  Introduction Our Planet Forward-Looking Statements This report contains certain forward-looking statements or projections that reflect our expectations regarding future events, performance and goals and involve several risks and uncertainties.  Generally speaking, any statement in this report not based upon historical fact is a forward-looking statement.  Forward-looking statements can also be identified by the use of forward-looking or conditional words, such mas “could,” “should,” “can,” “enable,” “continue,” estimate,” “forecast,” “intend,” “look,” “may,” “will,”\texttt{</Text>}} &  Paragraph \\
        \hline
        MLDR & \multicolumn{1}{p{5.cm}|}{What is Steven Whyte's occupation?} &  \multicolumn{1}{p{11.5cm}|}{\texttt{<Text>} Steven Whyte (born 17 March 1969) is a sculptor classically trained in the traditional methodology of figurative bronze and portrait sculpture living in Carmel, California. He has produced many public memorials and installations in both England and throughout the United States with subjects ranging from miners, to soldiers and fire fighters.  He is credited with over fifty life size and larger bronze public figures and major monuments including The Silverdale Mining Memorial, The Lance Sergeant Jack Baskeyfield VC Tribute, The Spirit of 1948, and The Dr. John Roberts Monument.  Whyte's multimillion-dollar, sixteen-figure monument in San Diego, California entitled National Salute to Bob Hope and the Military is one of his most notable works. \texttt{</Text>}} &  Paragraph \\
        \hline
        LooGLE & \multicolumn{1}{p{5.cm}|}{What were some of the subjects that Picardo illustrated in his post-student days?} &  \multicolumn{1}{p{11.5cm}|}{\texttt{<Title>}José Luis Picardo\texttt{</Title>} \texttt{<Text>} While he was still a student, architects commissioned murals from him for the interiors of their buildings, and employed him within their practices for the graphic representations and perspectives of their plans. Picardo executed his first professional mural painting at the age of 20 in 1939 in the Cine Fígaro (Figaro Cinema) in Madrid, commissioned by his architecture mentor Luis Moya Blanco. The painting of murals was the main source of income for Picardo during his youth and early career.As a student Picardo also began to illustrate many articles and later several covers for the Spanish architectural magazines Revista Nacional de Arquitectura and the Boletín de la Dirección General de Arquitectura. His drawings in these publications have been described as showing "increasing sophistication" and being of "complexity and extraordinary quality". Particularly noted in his post-student days were illustrations portraying Madrid in the 1950s and 1960s, the Spanish protectorate in Morocco, and sketches of the Canarias (Canary Islands) in 1953. He also showed a growing interest in historic architecture, in particular its preservation and restoration. Picardo completed his training by making increasingly numerous travels to study buildings around Spain and abroad. \texttt{</Text>}} &  Paragraph \\
        \hline
        SciFact & \multicolumn{1}{p{5.cm}|}{All hematopoietic stem cells segregate their chromosomes randomly.} &  \multicolumn{1}{p{11.5cm}|}{\texttt{<Title>}Haematopoietic stem cells do not asymmetrically segregate chromosomes or retain BrdU \texttt{</Title>} \texttt{<Text>} Sequential administration of 5-chloro-2-deoxyuridine and 5-iodo-2-deoxyuridine indicated that all HSCs segregate their chromosomes randomly. \texttt{</Text>}} &  Sentence \\
        \bottomrule
    \end{tabular}
    \end{adjustbox}
    \caption{Examples of query-relevant document pairs from semantic memory retrieval datasets. The corpus title is enclosed within \texttt{<Title>} and \texttt{</Title>} tags, while the corpus text is enclosed within \texttt{<Text>} and \texttt{</Text>} tags.}
    \label{tab:semantic_example}
\end{table*}

%% file: tables/procedural_examples.tex
\begin{table*}[tp]
    \centering
    \small
    \renewcommand\arraystretch{1.12}
    \fvset{fontsize=\small,baselinestretch=1.05}
    \begin{adjustbox}{width=\textwidth,max totalheight=0.96\textheight,center}
    \begin{tabular}{ l | l | l | c }
        \toprule
        \multicolumn{1}{l|}{\textbf{Dataset}} &
        \multicolumn{1}{|c}{\textbf{Query}} &
        \multicolumn{1}{|c}{\textbf{Relevant-Document}} &
        \multicolumn{1}{|c}{\textbf{Granularity}} \\
        \midrule
        \midrule
         Gorilla & \multicolumn{1}{p{4.5cm}|}{You are building a virtual global tour guide that can identify languages from the audio of people speaking. Use a model to identify which language is being spoken.} & \multicolumn{1}{p{18.cm}|}{\input{tables/gorilla_ex}} &  Tool \\
         \hline
         ToolBench & \multicolumn{1}{p{4.5cm}|}{My family is considering moving to a new city and we want to compare the sales tax rates of different cities.  Can you help us by providing the sales tax breakdowns by ZIP code for the cities we're interested in?  We would like to know the state, county, city, and any additional special sales taxes.} & \multicolumn{1}{p{18.cm}|}{\input{tables/toolbench_ex}} &  Tool \\
         \hline
         ReMe & \multicolumn{1}{p{4.5cm}|}{For your investment portfolio, could you inform me of the current price of 'Quasar Ltd.'?} & \multicolumn{1}{p{18.cm}|}{\input{tables/reme_ex}} &  Experience \\
         \hline
         Proced\_mem\_bench & \multicolumn{1}{p{4.5cm}|}{Examine a phone with the desk lamp while holding it} & \multicolumn{1}{p{18.cm}|}{\input{tables/proced_mem_bench_ex}} &  Trajectory \\
         \hline
         MemGovern & \multicolumn{1}{p{4.5cm}|}{Statistical distribution moments incorrectly defined as infinite instead of undefined (NaN) for distributions with undefined moments, causing inconsistent test behavior.} & \multicolumn{1}{p{18.cm}|}{Strategic Approach: The fix addresses the fundamental mathematical issue that certain statistical distributions have undefined moments rather than infinite moments.  The strategy involves correctly representing undefined moments as NaN (Not a Number) instead of infinity to properly reflect the mathematical reality and ensure consistent test behavior. Step-by-Step Implementation: 1.  Update the distribution's moment calculation method to return NaN for all moments instead of infinite values 2.  Modify test validation logic to skip moment checks when the expected moment value is not finite (NaN or infinite) 3.  Replace infinite moment checks with finite moment checks in test assertions 4.  Ensure sample mean and variance tests only run when the theoretical moments are finite} &  Experience \\
         \hline
         DeepPlanning & \multicolumn{1}{p{4.5cm}|}{I'm putting together a complete outdoor adventure outfit and need to find three specific items online.  First, I'm looking for a Women's Rho Hybrid Zip Neck Top that's proven popular with customers - it needs to have more than 25 four-star reviews and total sales volume exceeding 4000 to ensure it's a reliable choice.  Next, I need an all-seasons item that has solid customer feedback, specifically one with exactly 50 three-star ratings and more than 3000 total reviews, which shows it's been thoroughly tested by many buyers.  Finally, I'm searching for Women's Omni-Shield Adventure Trek Pants targeted specifically for women, and I need the exact item that has 18 four-star ratings and 689 total sales volume - these specific numbers help me identify the right product variant I've been researching.} & \multicolumn{1}{p{18.cm}|}{\input{tables/deepplanning_ex}} & Item \\
        \bottomrule
    \end{tabular}
    \end{adjustbox}
    \caption{Examples of query-relevant document pairs from procedural memory retrieval datasets.}
    \label{tab:procedural_example}
\end{table*}

%% file: tables/gorilla_ex.tex
\vspace{-1.5\baselineskip}
\begin{Verbatim}[breaklines=true]
{
    "domain": "Audio Classification", "framework": "Hugging Face Transformers", "functionality": "Language Identification", "api_call": "AutoModelForSpeechClassification.from_pretrained('sanchit-gandhi
    /whisper-medium-fleurs-lang-id')", "api_arguments": ["model = AutoModelForSpeechClassification.from_pretrained(
    'sanchit-gandhi/whisper-medium-fleurs-lang-id')", "processor = Wav2Vec2Processor.from_pretrained(
    'sanchit-gandhi/whisper-medium-fleurs-lang-id')"], "python_environment_requirements": ["transformers==4.27.0.dev0", "pytorch==1.13.1", "datasets==2.9.0", "tokenizers==0.13.2"], "example_code": "", "performance": {"dataset": "google/xtreme_s", "accuracy": 0.8805}, "description": "This model is a fine-tuned version of openai/whisper-medium on the FLEURS subset of the google/xtreme_s dataset. It is used for language identification in audio classification tasks.", "name": "sanchit-gandhi/whisper-medium-fleurs-lang-id"
}
\end{Verbatim}

%% file: tables/toolbench_ex.tex
\vspace{-0.8\baselineskip}
\begin{Verbatim}[breaklines=true]
{
    "category_name": "Finance", "required_parameters": [], 
    "optional_parameters": [{"name": "state", "type": "STRING", "description": "State name.", "default": ""}, {"name": "city", "type": "STRING", "description": "City name.", "default": ""}, {"name": "zip_code", "type": "STRING", "description": "Valid US ZIP code.", "default": "90210"}], 
    "method": "GET", 
    "template_response": {"zip_code": "str", "total_rate": "str", "state_rate": "str", "city_rate": "str", "county_rate": "str", "additional_rate": "str"}, 
    "name": "Sales_Tax_by_API_Ninjas_v1_salestax", 
    "description": "API Ninjas Sales Tax API endpoint. Returns one or more sales tax breakdowns by ZIP code according to the specified parameters. Each breakdown includes the state sales tax (if any), county sales tax (if any), city sales tax (if any), and any additional special sales taxes. All tax values are presented in decimals (e.g. 0.1 means 10% tax). \n\n Exactly one of the following must be set: zip_code or (city + state)"
}
\end{Verbatim}

%% file: tables/reme_ex.tex
\vspace{-0.8\baselineskip}
\begin{Verbatim}[breaklines=true]
{
    "when_to_use": "When a user requests stock information and subsequent watchlist management\n",
    "content": "1.  Use get_symbol_by_name to map company name to stock symbol\n2.  Fetch stock details via get_stock_info for price data\n3.  Add symbol to watchlist using add_to_watchlist\n4.  Verify watchlist state with get_watchlist\n5.  Use send_message with proper receiver_id and formatted content for communication\n"
}
\end{Verbatim}

%% file: tables/proced_mem_bench_ex.tex
\vspace{-0.8\baselineskip}
\begin{Verbatim}[breaklines=true]
{
  "task_description": "examine the alarmclock with the desklamp.",
  "state_action_pairs": [
    {
      "step_id": 1,
      "state": "You are in the middle of a room. Looking quickly around you, you see a bed 1, a desk 1, a drawer 5, a drawer 4, a drawer 3, a drawer 2, a drawer 1, a dresser 1, and a garbagecan 1.",
      "action": "go to desk 1"
    },
    {
      "step_id": 2,
      "state": "On the desk 1, you see a alarmclock 4, a alarmclock 3, a alarmclock 2, a alarmclock 1, a book 1, a cd 2, a cd 1, a creditcard 3, a creditcard 2, a desklamp 1, a keychain 3, a pen 1, and a pencil 3.",
      "action": "use desklamp 1"
    },
    {
      "step_id": 3,
      "state": "You turn on the desklamp 1.",
      "action": "examine alarmclock 1"
    },
    {
      "step_id": 4,
      "state": "On the desk 1, you see a alarmclock 4, a alarmclock 3, a alarmclock 2, a alarmclock 1, a book 1, a cd 2, a cd 1, a creditcard 3, a creditcard 2, a desklamp 1, a keychain 3, a pen 1, and a pencil 3.",
      "action": "examine desk 1"
    },
    {
      "step_id": 5,
      "state": "On the desk 1, you see a alarmclock 4, a alarmclock 3, a alarmclock 2, a alarmclock 1, a book 1, a cd 2, a cd 1, a creditcard 3, a creditcard 2, a desklamp 1, a keychain 3, a pen 1, and a pencil 3.",
      "action": "use desklamp 1 on alarmclock 1"
    },
    {
      "step_id": 6,
      "state": "You turn on the desklamp 1.",
      "action": "examine alarmclock 1 with desklamp 1"
    }
  ]
}
\end{Verbatim}

%% file: tables/deepplanning_ex.tex
\vspace{-0.8\baselineskip}
\begin{Verbatim}[breaklines=true]
{
  "product_id": "8e4ca530", "name": "Women's Rho Hybrid Zip Neck Top",
  .....
  "sales_volume": {"monthly": 350, "total": 4200},
  "rating": {"average_score": 4.8, "total_reviews": 215, "distribution": {"5_star": 180, "4_star": 28, "3_star": 5, "2_star": 1, "1_star": 1}},
  ......
}
{
  "product_id": "a393e541", "name": "Women's AirSense Pleated Wide Pants",
  ......
  "suitable_season": "All Seasons",
  ......
  "rating": {"average_score": 4.8, "total_reviews": 3150, "distribution": { "5_star": 2700, "4_star": 350, "3_star": 50, "2_star": 30, "1_star": 20}},
  ......
}
{
  "product_id": "581cf5c8", "name": "Women's Omni-Shield Adventure Trek Pants",
  ......
  "target_demographic": "Women",
  ......
  "sales_volume": { "monthly": 74, "total": 689
  },
  "rating": { "average_score": 4.8, "total_reviews": 135, "distribution": { "5_star": 112, "4_star": 18, "3_star": 3, "2_star": 1, "1_star": 1}},
  ......
}
\end{Verbatim}

%% file: tables/model_links.tex
\begin{table*}[ht]
    \centering
    \renewcommand\arraystretch{1.35}
    \resizebox{\textwidth}{!}{
    \begin{tabular}{l | c | l}
        \toprule
        \textbf{Embedding Model} &
        \textbf{Instruction} &
        \multicolumn{1}{c}{\textbf{Public Model Checkpoints (Link)}} \\
        \midrule
        \midrule
        KaLM-Embedding-Gemma3-12B-2511~\citeyear{DBLP:journals/corr/abs-2506-20923}
        & $\checkmark$ & \url{https://huggingface.co/tencent/KaLM-Embedding-Gemma3-12B-2511} \\

        KaLM-Embedding-V2.5~\citep{DBLP:journals/corr/abs-2506-20923}
        & $\checkmark$ & \url{https://huggingface.co/KaLM-Embedding/KaLM-embedding-multilingual-mini-instruct-v2.5} \\

        KaLM-Embedding-V1~\citep{DBLP:journals/corr/abs-2501-01028}
        & $\checkmark$ & \url{https://huggingface.co/HIT-TMG/KaLM-embedding-multilingual-mini-instruct-v1.5} \\
        \midrule

        bge-multilingual-gemma2~\citeyear{DBLP:conf/sigir/XiaoLZMLN24,DBLP:journals/corr/abs-2402-03216}
        & $\checkmark$ & \url{https://huggingface.co/BAAI/bge-multilingual-gemma2} \\

        bge-m3 (Dense)~\citep{DBLP:journals/corr/abs-2402-03216}
        & $\times$ & \url{https://huggingface.co/BAAI/bge-m3} \\

        bge-large-en-v1.5~\citep{DBLP:conf/sigir/XiaoLZMLN24}
        & $\times$ & \url{https://huggingface.co/BAAI/bge-large-en-v1.5} \\

        bge-base-en-v1.5~\citep{DBLP:conf/sigir/XiaoLZMLN24}
        & $\times$ & \url{https://huggingface.co/BAAI/bge-base-en-v1.5} \\

        bge-small-en-v1.5~\citep{DBLP:conf/sigir/XiaoLZMLN24}
        & $\times$ & \url{https://huggingface.co/BAAI/bge-small-en-v1.5} \\
        \midrule

        Qwen3-Embedding-8B~\citep{DBLP:journals/corr/abs-2506-05176}
        & $\checkmark$ & \url{https://huggingface.co/Qwen/Qwen3-Embedding-8B} \\

        Qwen3-Embedding-4B~\citep{DBLP:journals/corr/abs-2506-05176}
        & $\checkmark$ & \url{https://huggingface.co/Qwen/Qwen3-Embedding-4B} \\

        Qwen3-Embedding-0.6B~\citep{DBLP:journals/corr/abs-2506-05176}
        & $\checkmark$ & \url{https://huggingface.co/Qwen/Qwen3-Embedding-0.6B} \\
        \midrule

        F2LLM-v2-8B~\citep{f2llm-v2}
        & $\checkmark$ & \url{https://huggingface.co/codefuse-ai/F2LLM-v2-8B} \\

        F2LLM-v2-4B~\citep{f2llm-v2}
        & $\checkmark$ & \url{https://huggingface.co/codefuse-ai/F2LLM-v2-4B} \\
        \midrule

        SFR-Embedding-Mistral~\citep{SFRAIResearch2024}
        & $\checkmark$ & \url{https://huggingface.co/Salesforce/SFR-Embedding-Mistral} \\

        SFR-Embedding-2\_R~\citep{SFR-embedding-2}
        & $\checkmark$ & \url{https://huggingface.co/Salesforce/SFR-Embedding-2_R} \\
        \midrule

        e5-mistral-7b-instruct~\citep{DBLP:journals/corr/abs-2402-05672}
        & $\checkmark$ & \url{https://huggingface.co/intfloat/e5-mistral-7b-instruct} \\

        multilingual-e5-large-instruct~\citep{DBLP:journals/corr/abs-2402-05672}
        & $\checkmark$ & \url{https://huggingface.co/intfloat/multilingual-e5-large-instruct} \\

        e5-large-v2~\citep{DBLP:journals/corr/abs-2212-03533}
        & $\times$ & \url{https://huggingface.co/intfloat/e5-large-v2} \\

        multilingual-e5-small~\citep{DBLP:journals/corr/abs-2402-05672}
        & $\times$ & \url{https://huggingface.co/intfloat/multilingual-e5-small} \\

        e5-small-v2~\citep{DBLP:journals/corr/abs-2212-03533}
        & $\times$ & \url{https://huggingface.co/intfloat/e5-small-v2} \\
        \midrule

        jina-v5-text-small~\citep{akram2026jinaembeddingsv5texttasktargetedembeddingdistillation}
        & $\checkmark$ & \url{https://huggingface.co/jinaai/jina-embeddings-v5-text-small-retrieval} \\

        jina-v5-text-nano~\citep{akram2026jinaembeddingsv5texttasktargetedembeddingdistillation}
        & $\checkmark$ & \url{https://huggingface.co/jinaai/jina-embeddings-v5-text-nano-retrieval} \\
        \midrule

        snowflake-arctic-embed-l-v2.0~\citep{yu2024arctic}
        & $\checkmark$ & \url{https://huggingface.co/Snowflake/snowflake-arctic-embed-l-v2.0} \\

        snowflake-arctic-embed-m-v2.0~\citep{yu2024arctic}
        & $\checkmark$ & \url{https://huggingface.co/Snowflake/snowflake-arctic-embed-m-v2.0} \\
        \midrule

        NV-Embed-v2~\citep{DBLP:conf/iclr/Lee0XRSCP25}
        & $\checkmark$ & \url{https://huggingface.co/nvidia/NV-Embed-v2} \\

        Giga-Embeddings-instruct~\citep{kolodin2025gigaembeddings}
        & $\checkmark$ & \url{https://huggingface.co/ai-sage/Giga-Embeddings-instruct} \\

        stella\_en\_400M\_v5~\citep{zhang2024jasper}
        & $\checkmark$ & \url{https://huggingface.co/NovaSearch/stella_en_400M_v5} \\

        granite-embedding-311m-ml-r2~\citep{awasthy2026graniteembeddingmultilingualr2}
        & $\times$ & \url{https://huggingface.co/ibm-granite/granite-embedding-311m-multilingual-r2} \\

        EmbeddingGemma-300M~\citep{vera2025embeddinggemmapowerfullightweighttext}
        & $\times$ & \url{https://huggingface.co/google/embeddinggemma-300m} \\

        nomic-embed-text-v1~\citep{nussbaum2024nomic}
        & $\checkmark$ & \url{https://huggingface.co/nomic-ai/nomic-embed-text-v1} \\

        \bottomrule
    \end{tabular}
    }
    \caption{Publicly available embedding model checkpoints used for evaluation, grouped by model family. Instruction indicates whether LMEB task instructions are used: $\checkmark$ and $\times$ denote evaluation with and without instructions, respectively.}
    \label{tab:model_links}
\end{table*}

%% file: tables/episodic_tasks.tex
\begin{table*}[t]
    \centering
    \renewcommand\arraystretch{1.2}
    \resizebox{\textwidth}{!}{
    \begin{tabular}{ l | c | l | c | c}
        \toprule
        \multicolumn{1}{l|}{\textbf{Dataset}} &
        \multicolumn{1}{|c}{\textbf{\#Subsets}} &
        \multicolumn{1}{|c}{\textbf{Task Type}} &
        \multicolumn{1}{|c}{\textbf{
Example abilities assessed}} &
        \multicolumn{1}{|c}{\textbf{\# Tasks}} \\
        \midrule
        \midrule
        \multirow[t]{6}{*}{\textbf{EPBench}}  & \multirow[t]{6}{*}{9} & \multicolumn{1}{p{3.cm}|}{Entities} & \multicolumn{1}{p{10.cm}|}{Retrieve entities associated with specific events, times, or locations.} & \multirow[t]{6}{*}{54} \\
         \cline{3-4}
           &  & \multicolumn{1}{p{3.cm}|}{Event\_contents} & \multicolumn{1}{p{10.cm}|}{Retrieve events or activities involving specific entities.} &  \\
        \cline{3-4}
           &  & \multicolumn{1}{p{3.cm}|}{Full\_event\_details} & \multicolumn{1}{p{10.cm}|}{ Retrieve events with specific time, location, entities, and content} &  \\
        \cline{3-4}
           &  & \multicolumn{1}{p{3.cm}|}{Other\_entities} & \multicolumn{1}{p{10.cm}|}{Retrieve all other entities involved in a given event, excluding the specified ones.} &  \\
        \cline{3-4}
           &  & \multicolumn{1}{p{3.cm}|}{Spaces} & \multicolumn{1}{p{10.cm}|}{Retrieve locations associated with the events or entities specified in the query.} &  \\
        \cline{3-4}
           &  & \multicolumn{1}{p{3.cm}|}{Times} & \multicolumn{1}{p{10.cm}|}{Retrieve all dates when the specific events occurred.} &  \\
         \hline
         \multirow[t]{3}{*}{\textbf{KnowMeBench}} & \multirow[t]{3}{*}{{3}} & \multicolumn{1}{p{3.5cm}|}{adversarial\_abstention} & \multicolumn{1}{p{10.cm}|}{Retrieve memories providing counter-evidence to invalid query premises.} & \multirow[t]{3}{*}{{15}} \\
        \cline{3-4}
         &  & \multicolumn{1}{p{3.5cm}|}{information\_extraction} & \multicolumn{1}{p{10.cm}|}{Retrieve all relevant entities under strict temporal and spatial constraints.} &  \\
        \cline{3-4}
         &  & \multicolumn{1}{p{3.5cm}|}{mind-body\_interaction} & \multicolumn{1}{p{10.cm}|}{Retrieve action–state pairs to support reasoning about ironic or contradictory behaviors.} &  \\
        \cline{3-4}
         &  & \multicolumn{1}{p{3.75cm}|}{mnestic\_trigger\_analysis} & \multicolumn{1}{p{10.cm}|}{Retrieve memory events triggered by specific sensory or associative cues.} &  \\
         \cline{3-4}
         &  & \multicolumn{1}{p{3.5cm}|}{temporal\_reasoning} & \multicolumn{1}{p{10.cm}|}{Retrieve temporally constrained memories for timeline-based reasoning.} &  \\
        \bottomrule
    \end{tabular}
    }
    \caption{Overview of episodic retrieval tasks, including task types and examples of the abilities assessed.}
    \label{tab:episodic_tasks}
    \vspace{-0.1cm}
\end{table*}

%% file: tables/dialogue_tasks.tex
\begin{table*}[tp]
    \centering
    \footnotesize
    \renewcommand\arraystretch{1.3}
    \begin{adjustbox}{width=\textwidth,max totalheight=0.99\textheight,center}
    \begin{tabular}{ l | c | l | c | c}
        \toprule
        \multicolumn{1}{l|}{\textbf{Dataset}} &
        \multicolumn{1}{|c}{\textbf{\#Subsets}} &
        \multicolumn{1}{|c}{\textbf{Task Type}} &
        \multicolumn{1}{|c}{\textbf{
Example abilities assessed}} &
        \multicolumn{1}{|c}{\textbf{\# Tasks}} \\
        \midrule
        \midrule
        \multirow[t]{5}{*}{\textbf{LoCoMo}}  & \multirow[t]{5}{*}{1} & \multicolumn{1}{p{3.cm}|}{adversarial} & \multicolumn{1}{p{10.cm}|}{Retrieve evidence to support detecting unanswerable or misleading questions.} & \multirow[t]{5}{*}{5} \\

          \cline{3-4}
          & & \multicolumn{1}{p{3.cm}|}{multi\_hop} & \multicolumn{1}{p{10.cm}|}{Retrieve and integrate relevant information across multiple sessions to support multi-step reasoning.} &  \\
          \cline{3-4}
          & & \multicolumn{1}{p{3.cm}|}{open\_domain} & \multicolumn{1}{p{10.cm}|}{Retrieve conversation evidence relevant to questions that may require commonsense or world knowledge.} &  \\
          \cline{3-4}
          & & \multicolumn{1}{p{3.cm}|}{single\_hop} & \multicolumn{1}{p{10.cm}|}{Retrieve relevant information from a single session.} &  \\
          \cline{3-4}
          & & \multicolumn{1}{p{3.cm}|}{temporal\_reasoning} & \multicolumn{1}{p{10.cm}|}{Retrieve time-stamped or chronologically organized dialogue to support temporal reasoning.} &  \\
        \hline
        \multirow[t]{6}{*}{\textbf{LongMemEval}}  & \multirow[t]{6}{*}{1} & \multicolumn{1}{p{3.cm}|}{knowledge\_update} & \multicolumn{1}{p{10.cm}|}{Retrieve the most recent user information to detect changes in the user’s states or preferences.} & \multirow[t]{6}{*}{6} \\
        \cline{3-4}
        & & \multicolumn{1}{p{3.cm}|}{multi\_session} & \multicolumn{1}{p{10.cm}|}{Retrieve and aggregate relevant user information across two or more sessions.} & \\
        \cline{3-4}
        & & \multicolumn{1}{p{3.2cm}|}{single\_session\_assistant} & \multicolumn{1}{p{10.cm}|}{Retrieve the assistant’s own previously stated content within a single session.} & \\
        \cline{3-4}
        & & \multicolumn{1}{p{3.5cm}|}{single\_session\_preference} & \multicolumn{1}{p{10.cm}|}{Retrieve the user's stated preferences and constraints within a single session that are relevant to the query.} & \\
        \cline{3-4}
        & & \multicolumn{1}{p{3.cm}|}{single\_session\_user} & \multicolumn{1}{p{10.cm}|}{Retrieve user-mentioned facts or attributes within a single session to answer follow-up questions that depend on that information accurately and consistently.} & \\
        \cline{3-4}
        & & \multicolumn{1}{p{3.cm}|}{temporal\_reasoning} & \multicolumn{1}{p{10.cm}|}{Retrieve timestamped metadata and explicit time references from dialogue to align events chronologically and support reasoning about order, recency, duration, and time-dependent constraints.} & \\
        \hline
        
        \multirow[t]{3}{*}{\textbf{REALTALK}}  & \multirow[t]{3}{*}{1} & \multicolumn{1}{p{3.cm}|}{commonsense} & \multicolumn{1}{p{10.cm}|}{Retrieve conversational context supporting questions that require commonsense or world knowledge.} & \multirow[t]{6}{*}{3} \\
        \cline{3-4}
        & & \multicolumn{1}{p{3.cm}|}{multi\_hop} & \multicolumn{1}{p{10.cm}|}{Retrieve evidence across multiple sessions to connect dispersed details and support multi-step reasoning over integrated context.} &  \\
        \cline{3-4}
        & & \multicolumn{1}{p{3.cm}|}{temporal\_reasoning} & \multicolumn{1}{p{10.cm}|}{Retrieve chronologically relevant dialogue and time cues to reason about event sequences and time-based dependencies.} &  \\
        \hline

        \multirow[t]{3}{*}{\textbf{TMD}}  & \multirow[t]{1}{*}{1} & \multicolumn{1}{p{3.cm}|}{content\_time\_qs} & \multicolumn{1}{p{10.cm}|}{Retrieve relevant content by conditioning on speaker identity, timestamp/session metadata, etc.} & \multirow[t]{6}{*}{12} \\

        \cline{3-4}
         &  & \multicolumn{1}{p{3.cm}|}{date\_span\_time\_qs} & \multicolumn{1}{p{10.cm}|}{Retrieve dialogue content within a specified absolute date range (DATE1-DATE2).} & \\

          \cline{3-4}
         &  & \multicolumn{1}{p{3.cm}|}{dates\_time\_qs} & \multicolumn{1}{p{10.cm}|}{Retrieve dialogue associated with a specific absolute date.} & \\

          \cline{3-4}
         &  & \multicolumn{1}{p{3.cm}|}{day\_span\_time\_qs} & \multicolumn{1}{p{10.cm}|}{Retrieve discussions over a rolling recent window (\textit{e.g.,} last three days/last week).} & \\

          \cline{3-4}
         &  & \multicolumn{1}{p{3.cm}|}{earlier\_today\_time\_qs} & \multicolumn{1}{p{10.cm}|}{Retrieve topics discussed earlier the same day by using intra-day time cues.} & \\

          \cline{3-4}
         &  & \multicolumn{1}{p{3.cm}|}{last\_named\_day\_time\_qs} & \multicolumn{1}{p{10.cm}|}{Retrieve dialogue from the most recent occurrence of a specified weekday (\emph{e.g.,} ``last Sunday'').} & \\

          \cline{3-4}
         &  & \multicolumn{1}{p{3.cm}|}{month\_time\_qs} & \multicolumn{1}{p{10.cm}|}{Retrieve dialogue occurring within a specified month.} & \\

          \cline{3-4}
         &  & \multicolumn{1}{p{3.cm}|}{rel\_day\_time\_qs} & \multicolumn{1}{p{10.cm}|}{Retrieve dialogue from N days ago by resolving relative day offsets and mapping them to the correct date(s).} & \\

          \cline{3-4}
         &  & \multicolumn{1}{p{3.cm}|}{rel\_month\_time\_qs} & \multicolumn{1}{p{10.cm}|}{Retrieve dialogue from N months ago by resolving relative month offsets and month boundaries.} & \\

          \cline{3-4}
         &  & \multicolumn{1}{p{3.cm}|}{rel\_session\_time\_qs} & \multicolumn{1}{p{10.cm}|}{Retrieve content from N sessions ago by identifying session order and selecting the correct prior session(s).} & \\

          \cline{3-4}
         &  & \multicolumn{1}{p{3.cm}|}{session\_span\_time\_qs} & \multicolumn{1}{p{10.cm}|}{Retrieve dialogue between two referenced sessions (S1-S2).} & \\

          \cline{3-4}
         &  & \multicolumn{1}{p{3.cm}|}{session\_time\_qs} & \multicolumn{1}{p{10.cm}|}{Retrieve dialogue from a specified session number.} & \\
         \hline

        \multirow[t]{10}{*}{\textbf{MemBench}${\dag}$}  & \multirow[t]{10}{*}{10} & \multicolumn{1}{p{3.cm}|}{aggregative} & \multicolumn{1}{p{10.cm}|}{Retrieve evidence about more than two entities on a common attribute.} & \multirow[t]{10}{*}{10} \\

         \cline{3-4}
           &  & \multicolumn{1}{p{3.cm}|}{comparative} & \multicolumn{1}{p{10.cm}|}{Retrieve evidence about two entities from multiple messages and compare them on a shared attribute.} & \\

        \cline{3-4}
           &  & \multicolumn{1}{p{3.cm}|}{emotion} & \multicolumn{1}{p{10.cm}|}{Retrieve consecutive messages within a time window to infer the user’s emotional state.} & \\

        \cline{3-4}
           &  & \multicolumn{1}{p{3.cm}|}{knowledge\_updating} & \multicolumn{1}{p{10.cm}|}{Retrieve the most recent mention among conflicting updates.} & \\

        \cline{3-4}
           &  & \multicolumn{1}{p{3.cm}|}{multi\_hop} & \multicolumn{1}{p{10.cm}|}{Retrieve multiple relevant messages.} & \\

        \cline{3-4}
           &  & \multicolumn{1}{p{3.cm}|}{multi\_session\_assistant} & \multicolumn{1}{p{10.cm}|}{Retrieve multiple relevant assistant messages.} & \\

        \cline{3-4}
           &  & \multicolumn{1}{p{3.cm}|}{post\_processing} & \multicolumn{1}{p{10.cm}|}{Retrieve multiple messages that collectively support the required additional reasoning steps.} & \\

        \cline{3-4}
           &  & \multicolumn{1}{p{3.cm}|}{preference} & \multicolumn{1}{p{10.cm}|}{Retrieve relevant preference signals across multiple messages.} & \\

        \cline{3-4}
           &  & \multicolumn{1}{p{3.cm}|}{single\_hop} & \multicolumn{1}{p{10.cm}|}{Retrieve the single supporting message.} & \\

        \cline{3-4}
           &  & \multicolumn{1}{p{3.cm}|}{single\_session\_assistant} & \multicolumn{1}{p{10.cm}|}{Retrieve a single supporting assistant message.} & \\

        \hline

        \multirow[t]{6}{*}{\textbf{ConvoMem}${\ddag}$}  & \multirow[t]{6}{*}{6} & \multicolumn{1}{p{3.cm}|}{abstention\_evidence} & \multicolumn{1}{p{10.cm}|}{Retrieve evidence indicating the information is missing/undecided and support a correct abstention or ``not specified yet'' response.} & \multirow[t]{6}{*}{6} \\

        \cline{3-4}
           &  & \multicolumn{1}{p{3.cm}|}{assistant\_facts\_evidence} & \multicolumn{1}{p{10.cm}|}{Retrieve multiple supporting assistant-provided statements.} & \\

        \cline{3-4}
           &  & \multicolumn{1}{p{3.cm}|}{changing\_evidence} & \multicolumn{1}{p{10.cm}|}{Retrieve the chain of updates needed to determine the latest valid fact.} & \\

        \cline{3-4}
           &  & \multicolumn{1}{p{4.0cm}|}{implicit\_connection\_evidence} & \multicolumn{1}{p{10.cm}|}{Retrieve multiple context cues and connect them to infer the implied answer, grounding the inference in retrieved evidence.} & \\

        \cline{3-4}
           &  & \multicolumn{1}{p{3.cm}|}{preference\_evidence} & \multicolumn{1}{p{10.cm}|}{Retrieve distributed preference constraints relevant to the query.} & \\

        \cline{3-4}
           &  & \multicolumn{1}{p{3.cm}|}{user\_evidence} & \multicolumn{1}{p{10.cm}|}{Retrieve and link multiple user-stated facts across turns to answer a query that requires jointly grounded evidence.} & \\
        \bottomrule
    \end{tabular}
    \end{adjustbox}
    \caption{Overview of dialogue retrieval tasks, including task types and examples of the abilities assessed. $\dag$ MemBench has 10 subsets, each corresponding to a single task type; therefore, MemBench has 10 tasks. $\ddag$ ConvoMem has 6 subsets, each corresponding to a single task type; therefore, ConvoMem has 6 tasks.}
    \label{tab:dialogue_tasks}
    \vspace{-0.1cm}
\end{table*}

%% file: tables/semantic_tasks.tex
    \footnotesize
    \renewcommand\arraystretch{1.5}
    \begin{adjustbox}{width=\textwidth,max totalheight=0.7\textheight,center}
    \begin{tabular}{ l | c | l | c | c}
        \toprule
        \multicolumn{1}{l|}{\textbf{Dataset}} &
        \multicolumn{1}{|c}{\textbf{\#Subsets}} &
        \multicolumn{1}{|c}{\textbf{Task Type}} &
        \multicolumn{1}{|c}{\textbf{
Example abilities assessed}} &
        \multicolumn{1}{|c}{\textbf{\# Tasks}} \\
        \midrule
        \midrule
        \multirow[t]{1}{*}{\textbf{QASPER}}  & \multirow[t]{1}{*}{1} & \multicolumn{1}{p{3.cm}|}{QASPER} & \multicolumn{1}{p{10.cm}|}{Retrieve evidence passages from full NLP research papers to answer information-seeking questions.} & \multirow[t]{1}{*}{1} \\
        \hline

        \multirow[t]{1}{*}{\textbf{NovelQA}}  & \multirow[t]{7}{*}{1} & \multicolumn{1}{p{3.cm}|}{Character} & \multicolumn{1}{p{10.cm}|}{Retrieve passages providing information about the queried character.} & \multirow[t]{7}{*}{7} \\

        \cline{3-4}
          &  & \multicolumn{1}{p{3.cm}|}{Meaning} & \multicolumn{1}{p{10.cm}|}{Retrieve symbol/metaphor contexts or semantically matched passages to support an implication or chapter-level match.} &  \\

        \cline{3-4}
          &  & \multicolumn{1}{p{3.cm}|}{Plot} & \multicolumn{1}{p{10.cm}|}{Retrieve narrative cues about plot developments.} &  \\

        \cline{3-4}
          &  & \multicolumn{1}{p{3.cm}|}{Relation} & \multicolumn{1}{p{10.cm}|}{Retrieve distributed factual evidence and connect it to resolve entity relations.} &  \\

        \cline{3-4}
          &  & \multicolumn{1}{p{3.cm}|}{Setting} & \multicolumn{1}{p{10.cm}|}{Retrieve distributed factual evidence and connect it to resolve settings.} &  \\

        \cline{3-4}
          &  & \multicolumn{1}{p{3.cm}|}{Span} & \multicolumn{1}{p{10.cm}|}{Retrieve evidence identifying the temporal or geographical range of the story.} &  \\

        \cline{3-4}
          &  & \multicolumn{1}{p{3.cm}|}{Times} & \multicolumn{1}{p{10.cm}|}{Retrieve passages documenting occurrences of the queried event or entity to support counting.} &  \\
        \hline
        
        \multirow[t]{1}{*}{\textbf{PeerQA}}  & \multirow[t]{1}{*}{1} & \multicolumn{1}{p{3.cm}|}{PeerQA} & \multicolumn{1}{p{10.cm}|}{Retrieve sentence-level evidence from full scientific papers to answer peer-review–style questions grounded in the paper text.} & \multirow[t]{1}{*}{1} \\
        \hline
        
        \multirow[t]{1}{*}{\textbf{Covid-QA}}  & \multirow[t]{1}{*}{1} & \multicolumn{1}{p{3.cm}|}{Covid-QA} & \multicolumn{1}{p{10.cm}|}{Retrieve answer-bearing passages from COVID-19 biomedical articles (CORD-19) to support expert-annotated scientific question answering.} & \multirow[t]{1}{*}{1} \\
        \hline
        
        \multirow[t]{1}{*}{\textbf{ESG-Reports}}  & \multirow[t]{1}{*}{1} & \multicolumn{1}{p{3.cm}|}{ESG-Reports} & \multicolumn{1}{p{10.cm}|}{Retrieve relevant passages from long, text-converted ESG reports.} & \multirow[t]{1}{*}{1} \\
        \hline
        
        \multirow[t]{1}{*}{\textbf{MLDR}}  & \multirow[t]{1}{*}{1} & \multicolumn{1}{p{3.cm}|}{MLDR} & \multicolumn{1}{p{10.cm}|}{Retrieve the gold passage from long English Wikipedia documents.} & \multirow[t]{1}{*}{1} \\
        \hline
        
        \multirow[t]{2}{*}{\textbf{LooGLE}$\dag$}  & \multirow[t]{2}{*}{2} & \multicolumn{1}{p{3.cm}|}{LongDepQA} & \multicolumn{1}{p{10.cm}|}{Retrieve multiple supporting passages distributed across the document.} & \multirow[t]{1}{*}{2} \\

        \cline{3-4}
          & & \multicolumn{1}{p{3.cm}|}{ShortDepQA} & \multicolumn{1}{p{10.cm}|}{Retrieve the single supporting sentence or paragraph.} &  \\

        \hline
        
        \multirow[t]{1}{*}{\textbf{SciFact}}  & \multirow[t]{1}{*}{1} & \multicolumn{1}{p{3.cm}|}{SciFact} & \multicolumn{1}{p{10.cm}|}{Retrieve evidence sentences from scientific abstracts to support or refute a claim.} & \multirow[t]{1}{*}{1} \\
        \bottomrule
    \end{tabular}
    \end{adjustbox}
    \caption{Overview of semantic retrieval tasks, including task types and examples of the abilities assessed. $\dag$ LooGLE has 2 subsets, each corresponding to a single task type; therefore, LooGLE has 2 tasks.}
    \label{tab:semantic_tasks}
    \vspace{-0.2cm}

%% file: tables/procedural_tasks.tex
    \footnotesize
    \renewcommand\arraystretch{1.5}
    \begin{adjustbox}{width=\textwidth,max totalheight=0.7\textheight,center}
    \begin{tabular}{ l | c | l | c | c}
        \toprule
        \multicolumn{1}{l|}{\textbf{Dataset}} &
        \multicolumn{1}{|c}{\textbf{\#Subsets}} &
        \multicolumn{1}{|c}{\textbf{Task Type}} &
        \multicolumn{1}{|c}{\textbf{
Example abilities assessed}} &
        \multicolumn{1}{|c}{\textbf{\# Tasks}} \\
        \midrule
        \midrule
        \multirow[t]{3}{*}{\textbf{Gorilla}$\dag$}  & \multirow[t]{3}{*}{3} & \multicolumn{1}{p{3.cm}|}{gorilla\_huggingface} & \multicolumn{1}{p{10.cm}|}{Retrieve the appropriate HuggingFace model for the specified task query.} & \multirow[t]{3}{*}{3} \\

        \cline{3-4}
          &  & \multicolumn{1}{p{3.cm}|}{gorilla\_pytorch} & \multicolumn{1}{p{10.cm}|}{Retrieve the most task-aligned TorchHub API entry by matching the query intent.} &  \\

        \cline{3-4}
          &  & \multicolumn{1}{p{3.cm}|}{gorilla\_tensor} & \multicolumn{1}{p{10.cm}|}{Retrieve the most task-aligned TensorHub API entry by matching the query intent.} &  \\
        \hline

        \multirow[t]{1}{*}{\textbf{ToolBench}}  & \multirow[t]{1}{*}{1} & \multicolumn{1}{p{3.cm}|}{ToolBench} & \multicolumn{1}{p{10.cm}|}{Retrieve the relevant API documentation to support tool-use instructions, including both single-tool and multi-tool scenarios.} & \multirow[t]{1}{*}{1} \\

        \hline

        \multirow[t]{2}{*}{\textbf{ReMe}$\ddag$}  & \multirow[t]{2}{*}{6} & \multicolumn{1}{p{3.cm}|}{generalized\_query} & \multicolumn{1}{p{10.cm}|}{Retrieve experiences that remain applicable under a generalized query.} & \multirow[t]{1}{*}{9} \\

        \cline{3-4}
          & & \multicolumn{1}{p{3.cm}|}{task\_query} & \multicolumn{1}{p{10.cm}|}{Retrieve the most relevant experience that matches a specific task query.} & \\
          
        \hline

        \multirow[t]{3}{*}{\textbf{Proced\_mem\_bench}}  & \multirow[t]{3}{*}{1} & \multicolumn{1}{p{3.cm}|}{easy} & \multicolumn{1}{p{10.cm}|}{Retrieve the correct single-object placement procedure for a specified object-location query.} & \multirow[t]{3}{*}{3} \\

        \cline{3-4}

        &  & \multicolumn{1}{p{3.cm}|}{medium} & \multicolumn{1}{p{10.cm}|}{Retrieve multi-step procedures that include required state changes (\emph{e.g.,} heat/clean) before placement.} &  \\

        \cline{3-4}

        &  & \multicolumn{1}{p{3.cm}|}{hard} & \multicolumn{1}{p{10.cm}|}{Retrieve composite procedures requiring multi-object coordination (\emph{e.g.,} handling counts or multiple instances) and ensure coverage of all required objects.} &  \\

        \hline

        \multirow[t]{1}{*}{\textbf{MemGovern}}  & \multirow[t]{1}{*}{48} & \multicolumn{1}{p{3.cm}|}{MemGovern} & \multicolumn{1}{p{10.cm}|}{Retrieve fix experiences distilled from historical GitHub issue--PR--patch records to identify relevant solution patterns and steps for a bug-fixing task.} & \multirow[t]{1}{*}{48} \\

        \hline

        \multirow[t]{3}{*}{\textbf{DeepPlanning}$\delta$}  & \multirow[t]{3}{*}{3} & \multicolumn{1}{p{3.cm}|}{shopping\_level1} & \multicolumn{1}{p{10.cm}|}{Retrieve items that satisfy step-level/local constraints.} & \multirow[t]{3}{*}{3} \\

        \cline{3-4}
        
         & & \multicolumn{1}{p{3.cm}|}{shopping\_level2} & \multicolumn{1}{p{10.cm}|}{Retrieve a set of items relevant to shopping requests involving global constraints.} & \\

         \cline{3-4}
        
         & & \multicolumn{1}{p{3.cm}|}{shopping\_level3} & \multicolumn{1}{p{10.cm}|}{Retrieve items that match the specified conditions.} & \\
        \bottomrule
    \end{tabular}
    \end{adjustbox}
    \caption{Overview of procedural retrieval tasks, including task types and examples of the abilities assessed. $\dag$ Gorilla has 3 subsets, each corresponding to a single task type; therefore, Gorilla has 3 tasks. $\ddag$ ReMe has 6 subsets: appworld\_qwen3\_8b, appworld\_qwen3\_14b, appworld\_qwen3\_32b, bfcl\_qwen3\_8b, bfcl\_qwen3\_14b, and bfcl\_qwen3\_32b. The first three subsets contain only the generalized\_query task type, while the latter three subsets include both generalized\_query and task\_query. Therefore, ReMe comprises 9 tasks in total. $\delta$ DeepPlanning has 3 subsets, each corresponding to a single task type; therefore, DeepPlanning has 3 tasks.}
    \label{tab:procedural_tasks}
    \vspace{-0.1cm}

%% file: tables/instructions.tex
\begin{table*}[p]
    \centering
    \footnotesize
    \renewcommand\arraystretch{1.6}
    \begin{adjustbox}{width=\textwidth,max totalheight=0.96\textheight,center}
    \begin{tabular}{lll}
        \toprule
        \multicolumn{1}{l}{\textbf{Dataset}} &
        \multicolumn{1}{l}{\textbf{Task Type}} &
        \multicolumn{1}{l}{\textbf{Instruction}} \\
        \midrule
        \midrule
        \multicolumn{3}{c}{\textbf{Episodic Memory}} \\ 
        \hline
        \multirow[t]{6}{*}{\textbf{EPBench}}  &  \multicolumn{1}{p{3.cm}}{Entities} & \multicolumn{1}{p{15.cm}}{Given a specific event with time or location constraints, retrieve documents identifying the protagonists involved} \\
        \cline{2-3}
        &  \multicolumn{1}{p{3.cm}}{Event\_contents} & \multicolumn{1}{p{15.cm}}{Given a query, retrieve documents that answer the query}  \\
        \cline{2-3}
        &  \multicolumn{1}{p{3.cm}}{Full\_event\_details} & \multicolumn{1}{p{15.cm}}{Given time, location, entities, and event content, retrieve relevant documents about the event(s), including what occurred and any other pertinent information}  \\
        \cline{2-3}
        &  \multicolumn{1}{p{3.cm}}{Other\_entities} & \multicolumn{1}{p{15.cm}}{Given an event description with time/location and known entities, retrieve passages describing the event and naming any additional involved entities}  \\
        \cline{2-3}
        &  \multicolumn{1}{p{3.cm}}{Spaces} & \multicolumn{1}{p{15.cm}}{Given a query with constraints, retrieve passages that explicitly mention associated locations that satisfy those constraints}  \\
        \cline{2-3}
         &  \multicolumn{1}{p{3.cm}}{Times} & \multicolumn{1}{p{15.cm}}{Given a specific event description, retrieve passages that contain when the event occurred}  \\
         \hline

         \multirow[t]{5}{*}{\textbf{KnowMeBench}}  &  \multicolumn{1}{p{3.cm}}{adversarial\_abstention} & \multicolumn{1}{p{15.cm}}{Given a query, retrieve documents that answer the query} \\
         \cline{2-3}
         &  \multicolumn{1}{p{3.cm}}{information\_extraction} & \multicolumn{1}{p{15.cm}}{Given a query, retrieve documents that answer the query} \\
         \cline{2-3}
         &  \multicolumn{1}{p{4.cm}}{mind-body\_interaction} & \multicolumn{1}{p{15.cm}}{Given a mind-body interaction query, retrieve passages linking physical actions with internal states to answer the query}  \\
         \cline{2-3}
         &  \multicolumn{1}{p{3.cm}}{mnestic\_trigger\_analysis} & \multicolumn{1}{p{15.cm}}{Given a mnestic trigger query, retrieve passages where sensory or associative cues trigger memories, highlighting the cue, recalled event, and context to answer the query}  \\
         \cline{2-3}
         &  \multicolumn{1}{p{3.cm}}{temporal\_reasoning} & \multicolumn{1}{p{15.cm}}{Given a temporally constrained query, retrieve passages that answer the question}  \\
         \hline
         \multicolumn{3}{c}{\textbf{Dialogue Memory}} \\ 
         \hline

         \multirow[t]{5}{*}{\textbf{LoCoMo}}  &  \multicolumn{1}{p{3.cm}}{adversarial} & \multicolumn{1}{p{15.cm}}{Given a query, retrieve documents that answer the query} \\
         \cline{2-3}
         &  \multicolumn{1}{p{3.cm}}{multi\_hop} & \multicolumn{1}{p{15.cm}}{Given a multi-hop question, retrieve documents from multiple sessions to answer the question} \\
          \cline{2-3}
         &  \multicolumn{1}{p{3.cm}}{open\_domain} & \multicolumn{1}{p{15.cm}}{Given a question, retrieve user-provided context or external world-knowledge passages that answer the question} \\
          \cline{2-3}
         &  \multicolumn{1}{p{3.cm}}{single\_hop} & \multicolumn{1}{p{15.cm}}{Given a query, retrieve documents that answer the query} \\
         \cline{2-3}
         &  \multicolumn{1}{p{3.cm}}{temporal\_reasoning} & \multicolumn{1}{p{15.cm}}{Given a temporally grounded query, retrieve relevant passages that answer the question} \\
         \hline

         \multirow[t]{6}{*}{\textbf{LongMemEval}}  &  \multicolumn{1}{p{3.cm}}{knowledge\_update} & \multicolumn{1}{p{15.cm}}{Given a question, retrieve the latest information to answer the question} \\
         \cline{2-3}
          &  \multicolumn{1}{p{3.cm}}{multi\_session} & \multicolumn{1}{p{15.cm}}{Given a query, retrieve documents that answer the query} \\
          \cline{2-3}
          &  \multicolumn{1}{p{3.cm}}{single\_session\_assistant} & \multicolumn{1}{p{15.cm}}{Given a query, retrieve documents that answer the query} \\
          \cline{2-3}
          &  \multicolumn{1}{p{3.cm}}{single\_session\_preference} & \multicolumn{1}{p{15.cm}}{Given a query, retrieve the user's stated preferences that can help answer the query} \\
          \cline{2-3}
          &  \multicolumn{1}{p{3.cm}}{single\_session\_user} & \multicolumn{1}{p{15.cm}}{Given a query, retrieve documents that answer the query} \\
          \cline{2-3}
          &  \multicolumn{1}{p{3.cm}}{temporal\_reasoning} & \multicolumn{1}{p{15.cm}}{Given a temporal query, retrieve relevant passages that answer the query} \\
          \hline

         \multirow[t]{3}{*}{\textbf{REALTALK}}  &  \multicolumn{1}{p{3.cm}}{commonsense} & \multicolumn{1}{p{15.cm}}{Given a query, retrieve documents that answer the query} \\
         \cline{2-3}
         &  \multicolumn{1}{p{3.cm}}{multi\_hop} & \multicolumn{1}{p{15.cm}}{Given a multi-hop question, retrieve relevant passages that answer the query} \\
         \cline{2-3}
         &  \multicolumn{1}{p{3.cm}}{temporal\_reasoning} & \multicolumn{1}{p{15.cm}}{Given a temporal query, retrieve relevant passages that answer the query} \\
        \hline

         \multirow[t]{12}{*}{\textbf{TMD}}  &  \multicolumn{1}{p{3.cm}}{content\_time\_qs} & \multicolumn{1}{p{15.cm}}{Given a temporal query, retrieve relevant passages that answer the query} \\
         \cline{2-3}
         &  \multicolumn{1}{p{3.cm}}{date\_span\_time\_qs} & \multicolumn{1}{p{15.cm}}{Given an absolute date span (DATE1–DATE2), retrieve dialogue occurring within this period} \\
         \cline{2-3}
         &  \multicolumn{1}{p{3.cm}}{dates\_time\_qs} & \multicolumn{1}{p{15.cm}}{Given a specific absolute date, retrieve dialogue from that date} \\
         \cline{2-3}
         &  \multicolumn{1}{p{3.cm}}{day\_span\_time\_qs} & \multicolumn{1}{p{15.cm}}{Given a rolling recent window and the current date, retrieve dialogue that occurred within that period} \\
         \cline{2-3}
         &  \multicolumn{1}{p{3.cm}}{earlier\_today\_time\_qs} & \multicolumn{1}{p{15.cm}}{Given an intra-day time cue and the current date, retrieve dialogue that occurred earlier today} \\
         \cline{2-3}
         &  \multicolumn{1}{p{3.cm}}{last\_named\_day\_time\_qs} & \multicolumn{1}{p{15.cm}}{Given a query for a specific weekday with the current date, retrieve dialogue from the most recent matching weekday} \\
         \cline{2-3}
         &  \multicolumn{1}{p{3.cm}}{month\_time\_qs} & \multicolumn{1}{p{15.cm}}{Given a specific month and the current date, retrieve dialogue occurring within that month} \\
         \cline{2-3}
         &  \multicolumn{1}{p{3.cm}}{rel\_day\_time\_qs} & \multicolumn{1}{p{15.cm}}{Given a relative-day query and the current date, retrieve dialogue from the resolved date corresponding to that number of days ago} \\
         \cline{2-3}
         &  \multicolumn{1}{p{3.cm}}{rel\_month\_time\_qs} & \multicolumn{1}{p{15.cm}}{Given a relative-month query with the current date, retrieve dialogue from the month corresponding to that number of months ago} \\
         \cline{2-3}
         &  \multicolumn{1}{p{3.cm}}{rel\_session\_time\_qs} & \multicolumn{1}{p{15.cm}}{Given a relative-session query with current session index, retrieve dialogue from the session corresponding to that number of sessions ago} \\
         \cline{2-3}
         &  \multicolumn{1}{p{3.cm}}{session\_span\_time\_qs} & \multicolumn{1}{p{15.cm}}{Given a session-span query (S1–S2), retrieve dialogue from sessions that fall within the specified range} \\
         \cline{2-3}
         &  \multicolumn{1}{p{3.cm}}{session\_time\_qs} & \multicolumn{1}{p{15.cm}}{Given a query for a specific session, retrieve dialogue from that session} \\
         \hline

        \multirow[t]{10}{*}{\textbf{MemBench}}  &  \multicolumn{1}{p{3.cm}}{aggregative} & \multicolumn{1}{p{15.cm}}{Given a query, retrieve documents that answer the query} \\
        \cline{2-3}
        &  \multicolumn{1}{p{3.cm}}{comparative} & \multicolumn{1}{p{15.cm}}{Given a query, retrieve documents that answer the query} \\
        \cline{2-3}
        &  \multicolumn{1}{p{3.cm}}{emotion} & \multicolumn{1}{p{15.cm}}{Given a query with a timestamp, retrieve documents that capture emotional cues to infer most likely user sentiment} \\
        \cline{2-3}
        &  \multicolumn{1}{p{3.cm}}{knowledge\_updating} & \multicolumn{1}{p{15.cm}}{Given a question, retrieve the latest information to answer the question} \\
        \cline{2-3}
        &  \multicolumn{1}{p{3.cm}}{multi\_hop} & \multicolumn{1}{p{15.cm}}{Given a multi-hop query, retrieve documents that answer the query} \\
        \cline{2-3}
        &  \multicolumn{1}{p{3.cm}}{multi\_session\_assistant} & \multicolumn{1}{p{15.cm}}{Given a query, retrieve assistant messages containing past recommendations that help answer the query} \\
        \cline{2-3}
        &  \multicolumn{1}{p{3.cm}}{post\_processing} & \multicolumn{1}{p{15.cm}}{Given a query, retrieve documents that answer the query} \\
        \cline{2-3}
        &  \multicolumn{1}{p{3.cm}}{preference} & \multicolumn{1}{p{15.cm}}{Given a query, retrieve the user's stated preferences that can help answer the query} \\
        \cline{2-3}
        &  \multicolumn{1}{p{3.cm}}{single\_hop} & \multicolumn{1}{p{15.cm}}{Given a query, retrieve documents that answer the query} \\
        \cline{2-3}
        &  \multicolumn{1}{p{3.cm}}{single\_session\_assistant} & \multicolumn{1}{p{15.cm}}{Given a query, retrieve assistant messages containing past recommendations that help answer the query} \\
        \hline

        \multirow[t]{6}{*}{\textbf{ConvoMem}}  &  \multicolumn{1}{p{3.cm}}{abstention\_evidence} & \multicolumn{1}{p{15.cm}}{Given a query, retrieve documents that answer the query} \\
        \cline{2-3}
        &  \multicolumn{1}{p{3.cm}}{assistant\_facts\_evidence} & \multicolumn{1}{p{15.cm}}{Given a query, retrieve assistant messages that answer the query} \\
        \cline{2-3}
        &  \multicolumn{1}{p{3.cm}}{changing\_evidence} & \multicolumn{1}{p{15.cm}}{Given a question, retrieve the latest information to answer the question} \\
        \cline{2-3}
        &  \multicolumn{1}{p{3.cm}}{implicit\_connection\_evidence} & \multicolumn{1}{p{15.cm}}{Given a query, retrieve documents that answer the query} \\
        \cline{2-3}
        &  \multicolumn{1}{p{3.cm}}{preference\_evidence} & \multicolumn{1}{p{15.cm}}{Given a query, retrieve the user's stated preferences that can help answer the query} \\
        \cline{2-3}
        &  \multicolumn{1}{p{3.cm}}{user\_evidence} & \multicolumn{1}{p{15.cm}}{Given a query, retrieve documents that answer the query} \\
        
        \bottomrule
    \end{tabular}
    \end{adjustbox}
    \caption{Task instructions (datasets 1--8 of 22) for LMEB evaluation.}
    \label{tab:task_inst_1}
    \vspace{-0.1cm}
\end{table*}
\clearpage

\begin{table*}[p]
    \centering
    \footnotesize
    \renewcommand\arraystretch{1.69}
    \begin{adjustbox}{width=\textwidth,max totalheight=1.0\textheight,center}
    \begin{tabular}{lll}
        \toprule
        \multicolumn{1}{l}{\textbf{Dataset}} &
        \multicolumn{1}{l}{\textbf{Task Type}} &
        \multicolumn{1}{l}{\textbf{Instruction}} \\
        \midrule
        \midrule
        \multicolumn{3}{c}{\textbf{Semantic Memory}} \\ 
        \hline
        \multirow[t]{1}{*}{\textbf{QASPER}}  &  \multicolumn{1}{p{3.cm}}{QASPER} & \multicolumn{1}{p{15.cm}}{Given a query, retrieve documents that answer the query} \\
        \hline

        \multirow[t]{1}{*}{\textbf{NovelQA}}  &  \multicolumn{1}{p{3.cm}}{Character} & \multicolumn{1}{p{15.cm}}{Given a query, retrieve documents that answer the query} \\
        \cline{2-3}
         &  \multicolumn{1}{p{3.cm}}{Meaning} & \multicolumn{1}{p{15.cm}}{Given a query, retrieve documents that answer the query} \\
         \cline{2-3}
         &  \multicolumn{1}{p{3.cm}}{Plot} & \multicolumn{1}{p{15.cm}}{Given a query, retrieve documents that answer the query} \\
         \cline{2-3}
         &  \multicolumn{1}{p{3.cm}}{Relation} & \multicolumn{1}{p{15.cm}}{Given a query, retrieve documents that answer the query} \\
         \cline{2-3}
         &  \multicolumn{1}{p{3.cm}}{Setting} & \multicolumn{1}{p{15.cm}}{Given a query, retrieve documents that answer the query} \\
         \cline{2-3}
         &  \multicolumn{1}{p{3.cm}}{Span} & \multicolumn{1}{p{15.cm}}{Given a query, retrieve documents that answer the query} \\
         \cline{2-3}
         &  \multicolumn{1}{p{3.cm}}{Times} & \multicolumn{1}{p{15.cm}}{Given a query, retrieve all passages mentioning the event to answer the query} \\
         \hline

        \multirow[t]{1}{*}{\textbf{PeerQA}}  &  \multicolumn{1}{p{3.cm}}{PeerQA} & \multicolumn{1}{p{15.cm}}{Given a query, retrieve documents that answer the query} \\
        \hline

        \multirow[t]{1}{*}{\textbf{Covid-QA}}  &  \multicolumn{1}{p{3.cm}}{Covid-QA} & \multicolumn{1}{p{15.cm}}{Given a query on COVID-19, retrieve documents that answer the query} \\
        \hline

        \multirow[t]{1}{*}{\textbf{ESG-Reports}}  &  \multicolumn{1}{p{3.cm}}{ESG-Reports} & \multicolumn{1}{p{15.cm}}{Given a query, retrieve documents that answer the query} \\
        \hline

        \multirow[t]{1}{*}{\textbf{MLDR}}  &  \multicolumn{1}{p{3.cm}}{MLDR} & \multicolumn{1}{p{15.cm}}{Given a query, retrieve documents that answer the query} \\
        \hline

        \multirow[t]{2}{*}{\textbf{LooGLE}}  &  \multicolumn{1}{p{3.cm}}{LongDepQA} & \multicolumn{1}{p{15.cm}}{Given a long-dependency question, retrieve documents that answer the query} \\
        \cline{2-3}
         &  \multicolumn{1}{p{3.cm}}{ShortDepQA} & \multicolumn{1}{p{15.cm}}{Given a query, retrieve documents that answer the query} \\
        \hline

        \multirow[t]{1}{*}{\textbf{SciFact}}  &  \multicolumn{1}{p{3.cm}}{SciFact} & \multicolumn{1}{p{15.cm}}{Given a scientific claim, retrieve documents that support or refute the claim} \\
        \hline
        \multicolumn{3}{c}{\textbf{Procedural Memory}} \\ 
        \hline

        \multirow[t]{3}{*}{\textbf{Gorilla}}  &  \multicolumn{1}{p{3.cm}}{gorilla\_huggingface} & \multicolumn{1}{p{15.cm}}{Given a task query, retrieve HuggingFace models to complete it} \\
        \cline{2-3}
        &  \multicolumn{1}{p{3.cm}}{gorilla\_pytorch} & \multicolumn{1}{p{15.cm}}{Given a task query, retrieve TorchHub APIs to fulfill the query's needs} \\
        \cline{2-3}
        &  \multicolumn{1}{p{3.cm}}{gorilla\_tensor} & \multicolumn{1}{p{15.cm}}{Given a task query, retrieve TensorHub APIs to meet the query's requirements} \\
        \hline

        \multirow[t]{1}{*}{\textbf{ToolBench}}  &  \multicolumn{1}{p{3.cm}}{ToolBench} & \multicolumn{1}{p{15.cm}}{Given a tool-use query, retrieve API documentations aligned with the needs of the query} \\
        \hline
        
        \multirow[t]{2}{*}{\textbf{ReMe}}  &  \multicolumn{1}{p{3.cm}}{generalized\_query} & \multicolumn{1}{p{15.cm}}{Given a generalized query, retrieve experience that is suitable for this query} \\
        \cline{2-3}
         &  \multicolumn{1}{p{3.cm}}{task\_query} & \multicolumn{1}{p{15.cm}}{Given a query, retrieve the most relevant experience that aligns with the specified task requirements} \\
        \hline
        
         \multirow[t]{3}{*}{\textbf{Proced\_mem\_bench}}  &  \multicolumn{1}{p{3.cm}}{easy} & \multicolumn{1}{p{15.cm}}{Given a query, retrieve useful procedures applicable to this query} \\
         \cline{2-3}
         &  \multicolumn{1}{p{3.cm}}{medium} & \multicolumn{1}{p{15.cm}}{Given a query, retrieve useful procedures applicable to this query} \\
         \cline{2-3}
         &  \multicolumn{1}{p{3.cm}}{hard} & \multicolumn{1}{p{15.cm}}{Given a query, retrieve useful procedures applicable to this query} \\
         \hline

         \multirow[t]{1}{*}{\textbf{MemGovern}}  &  \multicolumn{1}{p{3.cm}}{MemGovern} & \multicolumn{1}{p{15.cm}}{Given a query, retrieve experiences or solutions that help to solve the query} \\
         \hline

         \multirow[t]{3}{*}{\textbf{DeepPlanning}}  &  \multicolumn{1}{p{3.cm}}{shopping\_level1} & \multicolumn{1}{p{15.cm}}{Given a query, retrieve items that satisfy the requirements} \\
         \cline{2-3}
         &  \multicolumn{1}{p{3.cm}}{shopping\_level2} & \multicolumn{1}{p{15.cm}}{Given a query, retrieve items that satisfy the requirements} \\
         \cline{2-3}
         &  \multicolumn{1}{p{3.cm}}{shopping\_level3} & \multicolumn{1}{p{15.cm}}{Given a query, retrieve items that satisfy the requirements} \\
         
        \bottomrule
    \end{tabular}
    \end{adjustbox}
    \caption{Task instructions (datasets 9--22 of 22) for LMEB evaluation.}
    \label{tab:task_inst_2}
    \vspace{-0.1cm}
\end{table*}

%% file: tables/reranking_model_links.tex
\begin{table*}[ht]
    \centering
    \renewcommand\arraystretch{1.25}
    \setlength{\tabcolsep}{4pt}
    \resizebox{\textwidth}{!}{%
    \begin{tabular}{l | c | l}
        \toprule
        \textbf{Reranking Model} &
        \textbf{Instruction} &
        \multicolumn{1}{c}{\textbf{Public Model Checkpoints (Link)}} \\
        \midrule
        \midrule
        Lychee-Reranker-Large~\citep{zhao2026kalmrerankerv1fastlateinteraction}
        & $\checkmark$
        & \url{https://huggingface.co/KaLM-Embedding/KaLM-Reranker-V1-Large-R2} \\

        Lychee-Reranker-Small~\citep{zhao2026kalmrerankerv1fastlateinteraction}
        & $\checkmark$
        & \url{https://huggingface.co/KaLM-Embedding/KaLM-Reranker-V1-Small-R2} \\

        Lychee-Reranker-Nano~\citep{zhao2026kalmrerankerv1fastlateinteraction}
        & $\checkmark$
        & \url{https://huggingface.co/KaLM-Embedding/KaLM-Reranker-V1-Nano-R2} \\
        \midrule

        Qwen3-Reranker-8B~\citep{DBLP:journals/corr/abs-2506-05176}
        & $\checkmark$
        & \url{https://huggingface.co/Qwen/Qwen3-Reranker-8B} \\

        Qwen3-Reranker-4B~\citep{DBLP:journals/corr/abs-2506-05176}
        & $\checkmark$
        & \url{https://huggingface.co/Qwen/Qwen3-Reranker-4B} \\

        Qwen3-Reranker-0.6B~\citep{DBLP:journals/corr/abs-2506-05176}
        & $\checkmark$
        & \url{https://huggingface.co/Qwen/Qwen3-Reranker-0.6B} \\
        \midrule

        bge-reranker-v2-gemma~\citeyear{DBLP:journals/corr/abs-2402-03216,li2024llama2vec}
        & $\checkmark$
        & \url{https://huggingface.co/BAAI/bge-reranker-v2-gemma} \\

        bge-reranker-large~\citep{DBLP:conf/sigir/XiaoLZMLN24}
        & $\times$
        & \url{https://huggingface.co/BAAI/bge-reranker-large} \\

        bge-reranker-v2-m3~\citeyear{DBLP:journals/corr/abs-2402-03216,li2024llama2vec}
        & $\times$
        & \url{https://huggingface.co/BAAI/bge-reranker-v2-m3} \\
        \midrule

        mxbai-rerank-large-v2~\citep{li2025prorank}
        & $\checkmark$
        & \url{https://huggingface.co/mixedbread-ai/mxbai-rerank-large-v2} \\

        mxbai-rerank-base-v2~\citep{li2025prorank}
        & $\checkmark$
        & \url{https://huggingface.co/mixedbread-ai/mxbai-rerank-base-v2} \\
        \midrule

        gte-multilingual-reranker-base~\citep{DBLP:conf/emnlp/ZhangZLXDTLYXHZ24}
        & $\times$
        & \url{https://huggingface.co/Alibaba-NLP/gte-multilingual-reranker-base} \\
        \midrule

        jina-reranker-v2-base-multilingual~\citep{jinaai2025rerankerv2base}
        & $\times$
        & \url{https://huggingface.co/jinaai/jina-reranker-v2-base-multilingual} \\

        \bottomrule
    \end{tabular}%
    }
    \caption{Publicly available reranking model checkpoints used for evaluation, grouped by model family.
    Instruction indicates whether LMEB task instructions are used:
    $\checkmark$ and $\times$ denote evaluation with and without
    instructions, respectively.}
    \label{tab:reranker_links}
\end{table*}

%% file: tables/detailed_embedding_results.tex
\clearpage
\onecolumn
\begin{landscape}
    \vspace*{\fill}
    \centering
    \renewcommand{\arraystretch}{0.95}
    \setlength{\tabcolsep}{3pt}
    \resizebox{\linewidth}{!}{%
    \begin{tabular}{l*{22}{c}}
        \toprule
        \multirow{2}{*}{\textbf{Model}} &
        \multicolumn{2}{c}{\textbf{Episodic}} &
        \multicolumn{6}{c}{\textbf{Dialogue}} &
        \multicolumn{8}{c}{\textbf{Semantic}} &
        \multicolumn{6}{c}{\textbf{Procedural}} \\
        \cmidrule(lr){2-3}
        \cmidrule(lr){4-9}
        \cmidrule(lr){10-17}
        \cmidrule(lr){18-23}
        & \rotatebox{60}{EPBench}
        & \rotatebox{60}{KnowMeBench}
        & \rotatebox{60}{LoCoMo}
        & \rotatebox{60}{LongMemEval}
        & \rotatebox{60}{REALTALK}
        & \rotatebox{60}{TMD}
        & \rotatebox{60}{MemBench}
        & \rotatebox{60}{ConvoMem}
        & \rotatebox{60}{QASPER}
        & \rotatebox{60}{NovelQA}
        & \rotatebox{60}{PeerQA}
        & \rotatebox{60}{Covid-QA}
        & \rotatebox{60}{ESG-Reports}
        & \rotatebox{60}{MLDR}
        & \rotatebox{60}{LooGLE}
        & \rotatebox{60}{SciFact}
        & \rotatebox{60}{Gorilla}
        & \rotatebox{60}{ToolBench}
        & \rotatebox{60}{ReMe}
        & \rotatebox{60}{Proced\_mem\_bench}
        & \rotatebox{60}{MemGovern}
        & \rotatebox{60}{DeepPlanning} \\
        \midrule
        \multicolumn{23}{c}{\textbf{Embedding Models > 1B parameters}} \\
        \midrule
        KaLM-Embedding-Gemma3 & 89.53 & 52.26 & 42.77 & 81.44 & 44.38 & \textbf{33.10} & \textbf{73.79} & 64.08 & 51.54 & 33.14 & 32.53 & 78.05 & 49.49 & 76.37 & 58.60 & 80.53 & \textbf{50.23} & \textbf{64.72} & 66.39 & 52.36 & \underline{91.27} & \textbf{55.63} \\
        bge-multilingual-gemma2 & 88.27 & 53.50 & \textbf{56.11} & \textbf{81.69} & \textbf{48.64} & \underline{31.10} & \underline{72.95} & 67.11 & 52.79 & \underline{37.70} & 32.79 & \textbf{79.81} & 46.20 & \underline{85.27} & \underline{64.53} & \textbf{84.14} & \underline{45.90} & \underline{62.40} & 64.97 & 52.63 & 88.94 & 53.56 \\
        Qwen3-Embedding-8B & 79.32 & 42.38 & 36.35 & 77.59 & 41.22 & 28.22 & 71.31 & 55.43 & 47.39 & 28.06 & 30.14 & \underline{79.38} & 43.16 & 77.54 & 59.53 & 78.84 & 45.21 & 48.60 & \textbf{68.09} & 51.11 & 89.35 & 52.35 \\
        F2LLM-v2-8B & \underline{91.40} & 40.81 & 42.29 & \underline{81.46} & 40.56 & 29.79 & 70.97 & 66.48 & \underline{57.66} & 32.05 & \textbf{36.45} & 75.67 & 46.35 & 76.44 & 56.42 & 77.99 & 41.73 & 62.21 & \underline{67.94} & \textbf{55.08} & 87.25 & 53.11 \\
        NV-Embed-v2 & 80.43 & \textbf{56.48} & 51.06 & 77.34 & \underline{48.01} & 19.61 & 71.45 & \textbf{71.04} & 57.45 & \textbf{40.67} & 34.92 & 78.58 & \textbf{52.39} & \textbf{85.39} & \textbf{66.05} & 82.01 & 43.10 & 54.38 & 61.57 & 52.08 & \textbf{92.50} & 49.00 \\
        e5-mistral-7b-instruct & 74.08 & 47.20 & 46.22 & 77.47 & 42.77 & 25.63 & 69.48 & 68.64 & 51.04 & 26.75 & 32.01 & 61.56 & 49.18 & 70.73 & 51.99 & 81.99 & 42.84 & 55.93 & 61.42 & 50.71 & 86.45 & 40.42 \\
        SFR-Embedding-Mistral & 75.91 & 47.08 & 50.05 & 79.75 & 45.27 & 25.17 & 70.19 & 67.84 & 54.00 & 32.22 & \underline{36.33} & 77.72 & 48.48 & 80.08 & 57.30 & 83.16 & 43.35 & 56.31 & 63.32 & 51.62 & 88.26 & 45.38 \\
        SFR-Embedding-2\_R & 79.79 & 46.45 & \underline{52.99} & 67.96 & 46.12 & 24.10 & 69.99 & \underline{69.42} & 51.48 & 30.46 & 34.80 & 68.72 & 44.48 & 79.25 & 59.57 & \underline{83.49} & 42.53 & 56.52 & 63.04 & 51.56 & 88.12 & 44.52 \\
        Qwen3-Embedding-4B & 73.91 & 40.57 & 29.87 & 47.06 & 34.48 & 29.21 & 66.28 & 54.59 & 50.19 & 25.80 & 27.15 & 76.69 & 40.77 & 77.78 & 58.80 & 79.00 & 42.41 & 51.85 & 67.00 & 54.34 & 89.53 & \underline{53.75} \\
        F2LLM-v2-4B & \textbf{92.15} & 40.74 & 43.32 & 81.42 & 42.00 & 29.62 & 70.55 & 66.40 & \textbf{57.97} & 29.73 & 35.48 & 75.11 & \underline{51.44} & 80.78 & 56.69 & 75.57 & 42.97 & 61.52 & 66.89 & \underline{54.37} & 86.51 & 49.79 \\
        Giga-Embeddings-instruct & 82.40 & \underline{53.69} & 48.96 & 67.47 & 43.46 & 21.53 & 61.04 & 65.11 & 42.33 & 30.85 & 27.37 & 75.33 & 36.63 & 82.90 & 56.08 & 80.76 & 43.92 & 56.88 & 66.51 & 51.66 & 85.83 & 40.71 \\
        \midrule
        \multicolumn{23}{c}{\textbf{Embedding Models < 1B parameters}} \\
        \midrule
        jina-v5-text-small & 82.79 & 40.73 & 39.41 & 74.63 & 36.93 & 22.57 & \textbf{72.67} & 60.51 & 50.70 & 23.33 & 30.30 & 72.96 & 36.82 & 76.47 & 51.74 & 78.27 & 33.94 & \textbf{58.26} & \underline{64.84} & 49.13 & 86.60 & 48.43 \\
        Qwen3-Embedding-0.6B & 82.11 & 36.97 & 34.20 & 72.77 & 37.19 & 27.99 & 70.52 & 63.17 & 47.29 & 22.64 & 27.34 & 74.20 & 37.41 & 76.61 & 55.55 & \underline{80.31} & \textbf{42.93} & 53.71 & 64.34 & \textbf{53.80} & 87.82 & \underline{54.78} \\
        snowflake-arctic-embed-l-v2.0 & 85.72 & 45.28 & 48.46 & 71.21 & 41.07 & 24.14 & 70.17 & 66.07 & \underline{52.83} & \textbf{32.35} & 33.31 & 77.41 & \underline{44.32} & 79.85 & \underline{59.64} & 78.45 & 37.08 & 50.92 & 63.85 & 48.88 & \underline{88.68} & 51.89 \\
        multilingual-e5-large-instruct & 75.58 & 43.95 & \textbf{50.07} & 69.86 & 44.92 & 27.07 & 67.37 & \textbf{68.42} & 51.25 & 28.42 & 32.40 & 69.28 & 35.19 & 80.93 & 53.61 & 79.81 & 36.10 & 49.22 & 62.92 & 51.77 & 84.60 & 48.66 \\
        bge-m3 (Dense) & \underline{87.46} & 46.55 & 46.58 & \textbf{78.37} & 44.97 & 28.09 & 71.23 & 64.42 & 51.51 & \underline{32.22} & 29.73 & \textbf{79.42} & 40.74 & 81.33 & 58.93 & 76.39 & 34.30 & 46.31 & 61.40 & 51.40 & 85.99 & 52.81 \\
        KaLM-Embedding-V2.5 & 85.17 & 40.23 & 42.20 & 74.43 & 38.96 & 17.06 & 69.80 & 62.24 & 46.11 & 27.42 & 30.46 & 73.38 & 42.98 & 79.07 & 57.75 & \textbf{80.56} & 33.59 & 54.17 & 64.15 & 52.83 & 84.66 & 50.96 \\
        KaLM-Embedding-V1 & 79.81 & 40.62 & 40.88 & 75.77 & 39.82 & 19.81 & 67.52 & 65.78 & 51.70 & 27.43 & 31.95 & 74.86 & 44.03 & 75.41 & 57.10 & 76.02 & 34.65 & \underline{57.81} & 63.27 & 47.91 & 86.90 & \textbf{55.66} \\
        stella\_en\_400M\_v5 & 84.18 & \textbf{50.66} & \underline{49.03} & \underline{77.70} & \underline{46.35} & 23.79 & \underline{72.31} & 65.86 & 46.89 & 31.89 & 31.11 & 75.34 & \textbf{45.85} & 81.05 & \textbf{60.86} & 79.66 & \underline{39.00} & 55.55 & \textbf{66.19} & 51.68 & 86.40 & 51.92 \\
        bge-large-en-v1.5 & 70.75 & 40.96 & 43.76 & 74.24 & \textbf{47.21} & 24.01 & 71.10 & 58.38 & 45.46 & 24.80 & 27.57 & 72.68 & 41.38 & 80.57 & 54.06 & 73.89 & 33.68 & 40.14 & 62.82 & 49.78 & 84.01 & 45.28 \\
        e5-large-v2 & 81.09 & 45.77 & 46.44 & 77.27 & 43.55 & \underline{33.91} & 64.48 & 62.90 & 50.43 & 29.59 & 31.07 & 77.93 & 41.21 & \underline{83.30} & 58.73 & 76.13 & 38.30 & 42.39 & 62.96 & 45.53 & 86.45 & 51.49 \\
        granite-embedding-311m-ml-r2 & 81.30 & 35.04 & 32.95 & 69.10 & 34.26 & 17.62 & 59.11 & 60.32 & 45.03 & 23.36 & 29.17 & 68.85 & 43.91 & 75.72 & 50.65 & 74.24 & 34.43 & 46.46 & 63.69 & 49.95 & 87.93 & 46.04 \\
        EmbeddingGemma-300M & \textbf{89.23} & \underline{47.14} & 42.15 & 77.03 & 41.52 & \textbf{33.94} & 70.54 & 58.43 & 45.16 & 26.23 & 29.83 & 70.96 & 42.71 & 76.75 & 58.96 & 78.06 & 37.01 & 56.43 & 63.23 & 51.66 & 86.46 & 49.12 \\
        snowflake-arctic-embed-m-v2.0 & 86.37 & 43.34 & 44.86 & 70.99 & 42.99 & 29.45 & 70.36 & \underline{67.19} & \textbf{53.27} & 30.91 & \underline{34.37} & \underline{77.95} & 42.22 & 81.73 & 59.05 & 79.64 & 36.77 & 44.47 & 63.26 & 52.64 & \textbf{89.19} & 51.94 \\
        jina-v5-text-nano & 78.53 & 37.48 & 41.21 & 69.87 & 35.51 & 25.51 & 68.20 & 62.68 & 48.34 & 28.59 & 32.14 & 76.67 & 43.38 & 77.12 & 58.17 & 75.49 & 32.93 & 54.32 & 56.07 & 46.47 & 87.26 & 48.55 \\
        nomic-embed-text-v1 & 81.75 & 40.99 & 45.80 & 73.74 & 42.02 & 26.91 & 40.96 & 66.15 & 51.56 & 30.47 & \textbf{35.34} & 76.87 & 42.27 & 82.29 & 57.91 & 77.71 & 37.80 & 45.72 & 60.19 & \underline{53.34} & 88.00 & 53.57 \\
        multilingual-e5-small & 78.81 & 36.58 & 36.42 & 47.30 & 33.55 & 20.67 & 63.28 & 45.84 & 48.95 & 26.12 & 31.06 & 77.14 & 36.22 & \textbf{83.68} & 56.40 & 73.43 & 31.66 & 43.80 & 59.58 & 44.17 & 82.03 & 50.64 \\
        bge-base-en-v1.5 & 65.92 & 38.23 & 36.73 & 74.60 & 40.00 & 29.08 & 60.29 & 63.31 & 48.15 & 22.76 & 30.58 & 72.62 & 38.78 & 80.84 & 49.27 & 77.75 & 36.33 & 43.93 & 63.01 & 47.87 & 87.10 & 49.45 \\
        bge-small-en-v1.5 & 73.38 & 36.80 & 36.37 & 60.83 & 38.98 & 21.52 & 66.70 & 62.44 & 46.71 & 21.42 & 28.42 & 72.28 & 40.59 & 76.41 & 50.67 & 75.40 & 31.97 & 40.96 & 61.54 & 45.35 & 85.39 & 48.16 \\
        e5-small-v2 & 79.33 & 41.92 & 40.46 & 71.54 & 41.48 & 28.07 & 62.19 & 58.74 & 49.33 & 27.25 & 30.83 & 76.24 & 40.55 & 81.07 & 56.55 & 74.94 & 28.59 & 38.85 & 61.98 & 48.06 & 83.96 & 49.17 \\
        \bottomrule
    \end{tabular}%
    }
    \captionsetup{hypcap=false}
    \captionof{table}{Performance of embedding models across all 22 LMEB datasets, measured by nDCG@10. Within each parameter group, the best results are \textbf{boldfaced}, and the second-best results are \underline{underlined}. }
    \label{tab:lmeb_embedding_dataset_results}
    \vspace*{\fill}
\end{landscape}
\twocolumn

%% file: tables/detailed_reranking_results.tex
\clearpage
\onecolumn
\begin{landscape}
    \vspace*{\fill}
    \centering
    \renewcommand{\arraystretch}{1.1}
    \setlength{\tabcolsep}{3pt}
    \resizebox{\linewidth}{!}{%
    \begin{tabular}{l*{22}{c}}
        \toprule
        \multirow{2}{*}{\textbf{Model}} &
        \multicolumn{2}{c}{\textbf{Episodic}} &
        \multicolumn{6}{c}{\textbf{Dialogue}} &
        \multicolumn{8}{c}{\textbf{Semantic}} &
        \multicolumn{6}{c}{\textbf{Procedural}} \\
        \cmidrule(lr){2-3}
        \cmidrule(lr){4-9}
        \cmidrule(lr){10-17}
        \cmidrule(lr){18-23}
        & \rotatebox{60}{EPBench}
        & \rotatebox{60}{KnowMeBench}
        & \rotatebox{60}{LoCoMo}
        & \rotatebox{60}{LongMemEval}
        & \rotatebox{60}{REALTALK}
        & \rotatebox{60}{TMD}
        & \rotatebox{60}{MemBench}
        & \rotatebox{60}{ConvoMem}
        & \rotatebox{60}{QASPER}
        & \rotatebox{60}{NovelQA}
        & \rotatebox{60}{PeerQA}
        & \rotatebox{60}{Covid-QA}
        & \rotatebox{60}{ESG-Reports}
        & \rotatebox{60}{MLDR}
        & \rotatebox{60}{LooGLE}
        & \rotatebox{60}{SciFact}
        & \rotatebox{60}{Gorilla}
        & \rotatebox{60}{ToolBench}
        & \rotatebox{60}{ReMe}
        & \rotatebox{60}{Proced\_mem\_bench}
        & \rotatebox{60}{MemGovern}
        & \rotatebox{60}{DeepPlanning} \\
        \midrule
        \multicolumn{23}{c}{\textbf{First-stage Retriever}} \\
        \midrule
        KaLM-Embedding-V2.5 & 83.28 & 40.22 & 41.88 & 75.18 & 38.59 & 16.82 & 69.59 & 62.74 & 46.21 & 27.25 & 30.34 & 73.66 & 42.34 & 79.12 & 57.92 & 81.73 & 33.06 & 54.15 & 64.35 & 51.56 & 85.17 & 51.24 \\
        \midrule
        \multicolumn{23}{c}{\textbf{Second-stage Reranker}} \\
        \midrule
        \rowcolor{gray!10}
        \multicolumn{23}{l}{\emph{Models with more than 1B parameters}} \\
        Qwen3-Reranker-8B & \underline{96.26} & \underline{61.18} & \textbf{65.91} & 78.93 & \textbf{57.13} & \underline{55.58} & 76.30 & 64.64 & 55.72 & \underline{48.21} & \textbf{40.64} & \underline{90.11} & \textbf{65.00} & \textbf{90.94} & \textbf{75.72} & \underline{83.93} & \textbf{50.58} & \underline{72.27} & \textbf{69.64} & 51.27 & 92.72 & \underline{64.56} \\
        Qwen3-Reranker-4B & 92.70 & 60.88 & 63.43 & 53.72 & 55.91 & \textbf{58.79} & \underline{78.65} & 66.68 & \textbf{59.96} & 47.89 & \underline{38.84} & 89.95 & 62.85 & \underline{90.43} & 75.58 & 80.53 & 48.81 & \textbf{73.12} & \underline{69.05} & \underline{53.20} & 90.69 & 63.67 \\
        Lychee-Reranker-Large & \textbf{97.20} & \textbf{61.50} & \underline{64.80} & \textbf{89.54} & \underline{56.68} & 34.93 & 77.98 & \underline{70.34} & \underline{59.93} & \textbf{49.40} & 38.07 & 89.16 & \underline{64.03} & 90.33 & \underline{75.59} & \textbf{86.20} & \underline{50.32} & 69.77 & 68.54 & \textbf{56.66} & \textbf{94.19} & \textbf{65.10} \\
        bge-reranker-v2-gemma & 91.76 & 55.67 & 61.30 & \underline{87.23} & 54.74 & 34.13 & 76.50 & 67.12 & 52.16 & 43.63 & 34.43 & \textbf{90.17} & 57.60 & 88.34 & 68.89 & 83.14 & 40.24 & 64.32 & 63.90 & 46.02 & 85.90 & 53.71 \\
        mxbai-rerank-large-v2 & 95.50 & 57.22 & 61.33 & 87.16 & 54.43 & 41.96 & \textbf{81.18} & \textbf{70.41} & 58.82 & 44.08 & 34.78 & 89.14 & 60.52 & 86.39 & 71.61 & 81.86 & 43.52 & 65.24 & 67.09 & 45.20 & \underline{93.57} & 62.27 \\
        \midrule
        \rowcolor{gray!10}
        \multicolumn{23}{l}{\emph{Models with 0.5B--1B parameters}} \\
        Lychee-Reranker-Small & \underline{95.70} & \textbf{57.90} & \textbf{62.51} & \textbf{86.66} & \textbf{54.05} & 34.87 & 75.41 & 68.41 & \textbf{57.72} & \textbf{44.97} & 35.40 & 86.85 & \textbf{60.65} & \underline{86.90} & \textbf{73.43} & \textbf{85.56} & \textbf{46.48} & \underline{67.40} & 66.67 & \textbf{55.12} & \underline{93.13} & \underline{63.90} \\
        bge-reranker-large & 88.81 & 54.24 & 58.94 & 84.10 & 51.50 & \underline{35.30} & 71.48 & 62.55 & 49.73 & 36.44 & 31.59 & 65.38 & 50.94 & 85.32 & 68.53 & 79.38 & 36.31 & 53.41 & 64.31 & 51.92 & 90.96 & 29.90 \\
        bge-reranker-v2-m3 & 94.90 & 55.65 & 60.20 & 82.39 & \underline{53.89} & 31.84 & \underline{76.82} & \underline{69.15} & 52.62 & 43.90 & \underline{35.73} & \textbf{91.08} & \underline{57.03} & 86.68 & 71.50 & 80.99 & 37.82 & 56.43 & 65.47 & 54.44 & 90.87 & 53.28 \\
        Qwen3-Reranker-0.6B & 92.10 & \underline{56.68} & \underline{60.32} & 84.90 & 53.26 & \textbf{40.63} & 75.66 & 67.36 & \underline{56.39} & \underline{44.41} & \textbf{36.26} & 84.57 & 56.28 & \textbf{89.43} & \underline{72.61} & \underline{82.21} & \underline{44.04} & \textbf{68.22} & \underline{67.20} & \underline{54.63} & 91.84 & \textbf{64.04} \\
        mxbai-rerank-base-v2 & \textbf{95.86} & 53.80 & 55.76 & \underline{86.29} & 49.75 & 33.61 & \textbf{79.97} & \textbf{69.41} & 54.32 & 41.51 & 34.71 & \underline{88.33} & 50.63 & 86.28 & 71.06 & 81.10 & 43.29 & 64.42 & \textbf{68.74} & 30.82 & \textbf{93.56} & 61.18 \\
        \midrule
        \rowcolor{gray!10}
        \multicolumn{23}{l}{\emph{Models with fewer than 0.5B parameters}} \\
        gte-multilingual-reranker-base & \textbf{94.76} & 49.03 & 51.01 & 78.94 & 47.97 & \underline{35.02} & 68.76 & \underline{54.78} & \textbf{54.71} & \underline{41.20} & 30.90 & \underline{88.74} & \underline{53.11} & 86.79 & 70.43 & 74.90 & \underline{36.86} & 55.38 & \textbf{64.36} & \underline{49.64} & \underline{89.22} & \underline{58.61} \\
        jina-reranker-v2-base-multilingual & 87.35 & \underline{49.56} & \underline{56.22} & \underline{86.55} & \textbf{51.53} & \textbf{35.97} & \underline{72.66} & 54.70 & \underline{54.42} & 39.01 & \textbf{34.52} & \textbf{89.46} & 50.64 & \textbf{87.73} & \underline{71.00} & \underline{79.52} & \textbf{38.57} & \underline{60.92} & \underline{63.51} & 45.04 & 77.23 & 57.81 \\
        Lychee-Reranker-Nano & \underline{91.87} & \textbf{53.96} & \textbf{58.44} & \textbf{86.66} & \underline{50.89} & 31.44 & \textbf{74.11} & \textbf{64.89} & 53.36 & \textbf{41.34} & \underline{34.42} & 85.01 & \textbf{55.45} & \underline{87.23} & \textbf{71.43} & \textbf{83.76} & 33.25 & \textbf{65.04} & 62.47 & \textbf{51.95} & \textbf{91.93} & \textbf{62.07} \\
        \bottomrule
    \end{tabular}%
    }
    \captionsetup{hypcap=false}
    \captionof{table}{Reranking performance across all 22 LMEB datasets, measured by nDCG@10. KaLM-Embedding-V2.5 serves as the first-stage retriever, retrieving up to 100 candidates for each query, which are then reranked by the second-stage rerankers. Within each parameter group, the best results are \textbf{boldfaced}, and the second-best results are \underline{underlined}. Minor discrepancies between the KaLM-Embedding-V2.5 scores reported here and those in Table~\ref{tab:lmeb_embedding_dataset_results} arise from differences in hardware and experimental environments; these discrepancies are negligible. }
    \label{tab:lmeb_reranking_dataset_results}
    \vspace*{\fill}
\end{landscape}
\twocolumn